%% file: main.tex
\documentclass[twocolumn]{svjour3}

\smartqed  

\usepackage[]{hyperref}

\usepackage{graphicx}
\usepackage{setspace}
\usepackage{epstopdf}
\usepackage{multirow,tabularx,makecell}
\usepackage{soul}

\usepackage[numbers]{natbib}

\usepackage[nolist,nohyperlinks]{acronym}
\acrodef{GPS}[GPS]{Global Positioning System}
\acrodef{SLAM}[SLAM]{Simultaneous Localization And Mapping}
\acrodef{SLAMs}[SLAMs]{Simultaneous Localization And Mapping systems}
\acrodef{GPS}[GPS]{Global Positioning System}
\acrodef{RTK}[RTK]{Real-time Kinematics}
\acrodef{GNSS}[GNSS]{Global Navigation Satellite System}
\acrodef{ROS}[ROS]{Robot Operating System}
\acrodef{API}[API]{Application Programming Interface}
\acrodef{UAV}[UAV]{Unmanned Aerial Vehicle}
\acrodef{UGV}[UGV]{Unmanned Ground Vehicle}
\acrodef{UV}[UV]{Ultra-Violet}
\acrodef{LED}[LED]{Light-emitting Diode}
\acrodef{MBZIRC}[MBZIRC]{Mohamed Bin Zayed International Robotics Challenge}
\acrodef{DARPA}[DARPA]{Defense Advanced Research Projects Agency}
\acrodef{IMU}[IMU]{Inertial Measurement Unit}
\acrodef{LTI}[LTI]{Linear time-invariant}
\acrodef{MPC}[MPC]{Model Predictive Control}
\acrodef{UVDAR}[UVDAR]{Ultra-Violet Direction And Ranging}
\acrodef{DOF}[DOF]{degree-of-freedom}
\acrodef{DOFs}[DOFs]{degrees-of-freedom}
\acrodef{LiDAR}[LiDAR]{Light Detection and Ranging}
\acrodef{ESC}[ESC]{Electronic Speed Controller}
\acrodef{LKF}[LKF]{Linear Kalman Filter}
\acrodef{UKF}[UKF]{Unscented Kalman Filter}
\acrodef{EKF}[EKF]{Extended Kalman Filter}
\acrodef{RAS}[RAS]{Robotics and Automation Society}
\acrodef{IEEE}[IEEE]{Institute of Electrical and Electronics Engineers}
\acrodef{MRS}[MRS]{Multi-robot Systems Group}

\usepackage{fancyvrb} 

\usepackage{cellspace}
\newcolumntype{D}{>{\hfill}N{3}{2}<{\hfill}}

\usepackage{wrapfig}
\usepackage[font={small}]{caption}

\usepackage{siunitx}

\usepackage{subfig}
\usepackage[export]{adjustbox}

\usepackage{color}
\usepackage{url}

\usepackage{amsmath,amsfonts,amssymb,bm}
\usepackage{nicefrac}

\newcommand{\m}[1]{\ensuremath{\mathbf{#1}}}
\newcommand\numberthis{\addtocounter{equation}{1}\tag{\theequation}}

\newcommand{\reffig}[1]{Fig.~\ref{#1}}

\newcommand{\refsec}[1]{Sec.~\ref{#1}}

\newcommand{\real}{\mathbb{R}}
\newcommand{\minus}{\scalebox{0.75}[1.0]{$-$}}
\newcommand{\plus}{\scalebox{0.8}[0.8]{$+$}}

\newcommand{\updated}[1]{#1}
\newcommand{\added}[1]{#1}

\usepackage{tikz}
\usepackage{pgfplots}
\pgfplotsset{compat=1.14}
\usetikzlibrary{backgrounds,arrows,automata,shapes,positioning,calc,through,spy}
\pgfdeclarelayer{background}
\pgfdeclarelayer{foreground}
\pgfsetlayers{background,main,foreground}

\input{src/tikz}

\tikzset{new spy style/.style={spy scope={%
  magnification=5,
  size=1.25cm,
  connect spies,
  every spy on node/.style={
    rectangle,
    draw,
  },
  every spy in node/.style={
    draw,
    rectangle,
    fill=white
  }
  }
  }
  }

  \usepackage{scalerel}
  \usetikzlibrary{svg.path}
  \definecolor{orcidlogocol}{HTML}{A6CE39}
  \tikzset{
    orcidlogo/.pic={
      \fill[orcidlogocol] svg{M256,128c0,70.7-57.3,128-128,128C57.3,256,0,198.7,0,128C0,57.3,57.3,0,128,0C198.7,0,256,57.3,256,128z};
      \fill[white] svg{M86.3,186.2H70.9V79.1h15.4v48.4V186.2z}
      svg{M108.9,79.1h41.6c39.6,0,57,28.3,57,53.6c0,27.5-21.5,53.6-56.8,53.6h-41.8V79.1z M124.3,172.4h24.5c34.9,0,42.9-26.5,42.9-39.7c0-21.5-13.7-39.7-43.7-39.7h-23.7V172.4z}
      svg{M88.7,56.8c0,5.5-4.5,10.1-10.1,10.1c-5.6,0-10.1-4.6-10.1-10.1c0-5.6,4.5-10.1,10.1-10.1C84.2,46.7,88.7,51.3,88.7,56.8z};
    }
  }
  \newcommand\orcidicon[1]{\href{https://orcid.org/#1}{\mbox{\scalerel*{
    \begin{tikzpicture}[yscale=-1,transform shape]
      \pic{orcidlogo};
    \end{tikzpicture}
  }{|}}}}

\begin{document}\sloppy

  \title{The MRS UAV System: Pushing the Frontiers of Reproducible Research, Real-world Deployment, and Education with Autonomous Unmanned Aerial Vehicles}
  \titlerunning{The MRS UAV System}

  \author{Tomas Baca$^{*\orcidicon{0000-0001-9649-8277}}$ \and Matej Petrlik$^{\orcidicon{0000-0002-5337-9558}}$ \and Matous Vrba$^{\orcidicon{0000-0002-4823-8291}}$ \and Vojtech Spurny$^{\orcidicon{0000-0002-9019-1634}}$ \and Robert Penicka$^{\orcidicon{0000-0001-8549-4932}}$ \and Daniel Hert$^{\orcidicon{0000-0003-1637-6806}}$ \and Martin Saska$^{\orcidicon{0000-0001-7106-3816}}$}
  \authorrunning{Tomas Baca at al.}

  \institute{All authors are with \at
  Multi-Robot Systems group,
  Faculty of Electrical Engineering,
  Czech Technical University in Prague,
  Technicka~2,
  Prague,
  Czech Republic
  \and
  $^*$ corresponding author, \email{tomas.baca@fel.cvut.cz}
  }

  \date{Received: date / Accepted: date}

  \maketitle

  \begin{abstract}
    We present a multirotor \ac{UAV} control and estimation system for supporting replicable research through realistic simulations and real-world experiments.
    We propose a unique multi-frame localization paradigm for estimating the states of a \ac{UAV} in various frames of reference using multiple sensors simultaneously.
    The system enables complex missions in GNSS and GNSS-denied environments, including outdoor-indoor transitions and the execution of redundant estimators for backing up unreliable localization sources.
    Two feedback control designs are presented: one for precise and aggressive maneuvers, and the other for stable and smooth flight with a noisy state estimate.
    The proposed control and estimation pipeline are constructed without using the Euler/Tait-Bryan angle representation of orientation in 3D.
    Instead, we rely on rotation matrices and a novel heading-based convention to represent the one free rotational degree-of-freedom in 3D of a standard multirotor helicopter.
    We provide an actively maintained and well-documented open-source implementation, including realistic simulation of \acp{UAV}, sensors, and localization systems.
    The proposed system is the product of years of applied research on multi-robot systems, aerial swarms, aerial manipulation, motion planning, and remote sensing.
    All our results have been supported by real-world system deployment that subsequently shaped the system into the form presented here.
    In addition, the system was utilized during the participation of our team from the Czech Technical University in Prague in the prestigious MBZIRC 2017 and 2020 robotics competitions, and also in the DARPA Subterranean challenge.
    Each time, our team was able to secure top places among the best competitors from all over the world.
    \keywords{Unmanned Aerial Systems \and Multirotor Helicopters \and Control Systems Engineering \and Educational Robotics}
  \end{abstract}



  \section{INTRODUCTION}


  \begin{figure}[t]
    \centering
    \includegraphics[width=0.48\textwidth]{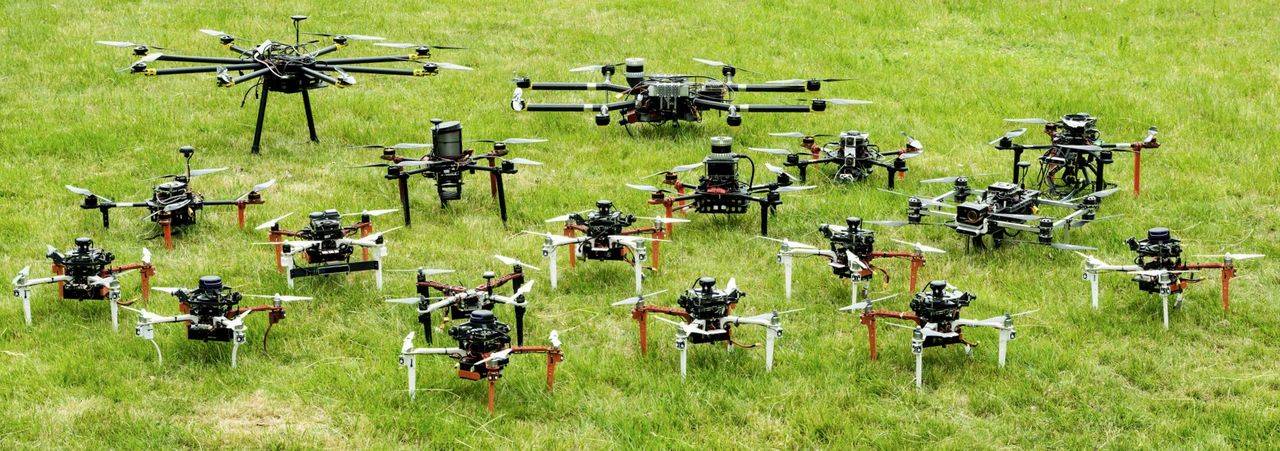}
    \caption{Multirotor UAV platforms equipped for various scenarios carried out by the system presented here.}
    \label{fig:platforms}
  \end{figure}

  The field of mobile robotics is steadily advancing towards smart, small and intelligent mobile agents, capable of autonomously solving complex tasks.
  Existing \ac{UGV} platforms already offer researchers complex functions (Clearpath Robotics\footnote{\url{http://clearpathrobotics.com}}, ETH ANYmal\footnote{\url{http://rsl.ethz.ch/robots-media/anymal}}).
  Ground robotics research tends to focus on high-level systems such as mission autonomy, robot localization, environment mapping, and remote sensing.
  Platforms such as the Boston Dynamics Spot\footnote{\url{http://www.bostondynamics.com/spot}} are out of the box equipped with automatic localization, mapping, path tracking, and navigation in an environment.
  However, \acp{UAV}, specifically multirotor helicopters, are still under intensive investigation in a wide range of research on all levels of their technological tree.
  Research in \acp{UAV} is still being carried in underlying fields of dynamic system modeling \cite{rodriguez2020explicit}, automatic feedback control \cite{jasim2020robust}, and trajectory optimization \cite{sanchez2020trajectory}.
  These fields are vital for understanding and for realizing autonomous flying machines capable of supporting research in higher-level sub-systems for autonomous navigation through an environment, for remote sensing, and for multi-agent systems.
  Only a handful of \ac{UAV} platforms are suited for research out of the box, and the researchers are most often tasked with developing full-stack \ac{UAV} control and guidance to support the needs of these platforms.

  Multirotor \acp{UAV} are capable of traversing 3D space and are often chosen for exploration and remote sensing in cluttered environments \cite{delmerico2019current}, especially when ground robots might fail.
  Their interesting dynamics makes them still a common choice for demonstrating novel techniques in control theory \cite{ramirez2020fuzzy}.
  In addition, their flight properties, most of all the ability to hover, make them excellent for carrying sensors and for aiding research in distributed and remote sensing.
  Due to the inherent instability of multirotor dynamics, a continuously-updated feedback control loop is necessary to maintain stable flight.
  This emphasizes the importance of the onboard localization, state estimation, and control software.
  Failure of this software threatens the vehicle itself and its surroundings.

  Experimental verification of novel methods for \ac{UAV} systems is nowadays becoming a standard in application-oriented research.
  However, this comes at the cost of obtaining and maintaining an experimental platform, preferably with a realistic simulation environment.
  This task is especially demanding if the platform is intended for use outside laboratory conditions.
  Replicating and validating existing research and comparing it to novel proposed approaches is a necessary part of the research process.
  We argue that novel methods in applied robotics should be published together with materials necessary for replicating the results.
  Although the amount of cutting-edge research published with enclosed sources is increasing, the situation is not yet ideal.
  A positive trend is most prominent in machine learning, in computer vision, and in \ac{SLAMs}\acused{SLAM}.
  Sadly, research groups rarely release all parts of their experimental and testing systems, making their results difficult to validate and replicate.
  Furthermore, research in the field of \ac{UAV} control is typically limited to non-realistic simulations and, in many cases no implementation in a real \ac{UAV} exists.
  To solve this, the robotics community has been collaborating on developing application frameworks \cite{hentout2016survey, elkady2012robotics, tsardoulias2017robotic} that unify the way for algorithms of different origin to interact and form complex robotics systems.
  The emergence of such frameworks helps to create systems that are reusable across research groups.

  The \ac{ROS} \cite{quigley2009ros} is one of many middleware robotics frameworks \cite{tsardoulias2017robotic, elkady2012robotics, hentout2016survey, inigo2012robotics, ceballos2011genom, metta2006yarp}.
  \ac{ROS} has several features that have raised it to the most prominent framework among \ac{UAV} researchers.
  Renowned sensor manufactures such as \emph{Velodyne}, \emph{Ouster}, \emph{Terabee}, \emph{Garmin}, \emph{MatrixVision} and \emph{Intel} are making a significant effort to provide \ac{ROS} drivers for their products.
  State-of-the-art research in computer vision and in \ac{SLAM} algorithms is often accompanied by functioning \ac{ROS} implementations of the published methods \cite{qin2018vins, zhang2014loam, kohlbrecher2011flexible}.
  Finally, open-source robotic simulators such Gazebo\footnote{Gazebo simulator, \url{http://gazebosim.org}} and CoppeliaSim\footnote{\url{http://www.coppeliarobotics.com}} (previously V-REP) provide integration with \ac{ROS}.
  These features allow researchers to focus their research more narrowly rather than on implementation aspects of a whole robotics pipeline.
  However, even with such advances, many tasks remain unsolved on the way to a real-world \ac{UAV} platform for research, especially to platforms that can perform outside laboratory conditions.

  Through this publication, we intend to share our full-stack \ac{UAV} platform with all essential capabilities for research, development, and testing of novel methods.
  Our system is a product of many years of development in various robotic projects.
  The proposed platform has provided support for state-of-the-art research and has resulted in dozens of high-quality publications in cooperation with several research groups.
  These works have focused on particular applications and on relevant research, but the underlying system will be thoroughly described and published for the first time in this manuscript.
  We offer a modular and extensible open-source platform, together with a complex simulation environment.
  The platform is suited for both indoor and outdoor use, with an emphasis on onboard multi-sensor fusion to allow safe execution of experiments outside laboratory conditions (see \reffig{fig:platforms} for a showcase of our hardware platforms).
  We propose a pair of feedback controllers that satisfy the needs of a wide range of applications, ranging from fast and aggressive flight to stable flight using unreliable sensors producing noisy data.



  \subsection{\updated{State of the art}}

  Research-focused \acp{UAV} are most commonly equipped with a low-level embedded flight controller.
  Available flight controllers \cite{ebeid2018survey} range from feature-packed open-source systems, such as Pixhawk, to proprietary commercial units manufactured by DJI.
  Table~\ref{tab:embedded_flight_controllers} shows a comparison of often used solutions.
  Pixhawk is often used in research projects (including our project), typically running either of the two open-source firmwares: PX4 \cite{meier2015px4} and ArduPilot\footnote{\url{http://ardupilot.org}}.
  \updated{Although all of these flight stacks provide sophisticated features up to waypoint tracking and mission execution, the features are rarely used within real-world applications and for real-world verification of research}.
  \updated{As supported by the existence of many high-level control systems \cite{sanchez2016aerostack, xiao2020xtdrone, furrer2016rotors, schmittle2018openuav, abeywardena2015design, mellado2013mavwork}, researchers instead use other onboard companion computers to execute a custom localization system, state estimators, and flight controllers, and only low-level control commands are provided for the embedded flight controller \cite{kamel2017robust, qin2018vins, gao2018online, mohta2018fast, sanchez2020trajectory}.}

  Several comparable \ac{UAV} systems have been published and released.
  Table~\ref{tab:uav_systems} compares existing solution with the system proposed in this publication.

  The RotorS \cite{furrer2016rotors} simulator is an initial release for the Aeroworks EU project\footnote{Aeroworks EU project, \url{http://www.aeroworks2020.eu}.}.
  It provides Gazebo-based simulation of the now discontinued \emph{Ascending Technologies} \ac{UAV} system.
  The control pipeline features are basic, with little potential for transfer to real-world conditions.
  The system does not appear to be kept up-to-date, which gradually diminishes its usability and applicability.
  Moreover, the latest supported version of ROS is \emph{ROS Kinetic}, which potentially provides lower compatibility with newer hardware and software.

  OpenUAV\footnote{OpenUAV, \url{http://github.com/Open-UAV}} \cite{schmittle2018openuav} is a \ac{UAV} swarm simulation testbed.
  The system does not appear to allow transfer to a real-world setting, and is designed only to support prototyping of basic research in swarming.
  The \acp{UAV} are assumed to be controlled and localized solely using an embedded flight controller with PX4 firmware.
  This is comparable hardly with the numerous sensors and localization systems that our system allows to simulate and to be used in a real-world scenario.

  ReCOPTER\footnote{ReCOPTER, \url{http://github.com/thedinuka/ReCOPTER}} \cite{abeywardena2015design} proposes an open-source multirotor system for research.
  The available materials were released as supporting material for the published paper.
  However, no software was attached, and the materials have not been updated since.
  Similarly, a framework for drone control using the Vicon localization system named MAVwork\footnote{MAVwork, \url{http://github.com/uavster/mavwork}} \cite{mellado2013mavwork} was published in 2011, but has not been updated since.
  Although sources were made available, they offered only basic features that would be difficult to transfer into a real-world scenario.

  The XTDrone\footnote{XTDrone, \url{http://github.com/robin-shaun/XTDrone}} \cite{xiao2020xtdrone} simulation testbed offers many complex functionalities that are comparable with our proposed system, including simulation of onboard sensors and complex localization systems.
  However, the control pipeline relies entirely on the PX4 embedded control software.
  This significantly limits any transfer to a custom hardware platform, or even the ability to simulate realistic conditions using onboard localization systems.
  Thus, the use of XTDrone outside laboratory conditions is mostly limited to \ac{GPS}-localized flight in a non-cluttered outdoor environment.

  The full-stack Aerostack system\footnote{Aerostack, \url{http://github.com/Vision4UAV/Aerostack}} \cite{sanchez2016aerostack, sanchez2016reliable} was designed for deployment of multirotor \acp{UAV}.
  The system is continuously being updated, and it offers an option to transfer to a real-world platform.
  According to the preprint\cite{suarez2020skyeye} where the authors used Aerostack during the MBZIRC 2020 competition, the system's real-world deployment is possible.
  However, with the used DJI-based flight controller, the control command supplied to the underlying embedded control layer are limited to desired orientation and thrust.
  This level of control limits the potential precision and control authority comparing to our system.
  Furthermore, the system lacks the feature of switching between multiple frames of reference, which is one of our system's contributions.
  As it happens, the team of authors of Aerostack did not compete in the wall-building challenge of MBZIRC 2020, in which we found the feature to be crucial to precisely collect bricks by a group of \acp{UAV}.

  Besides the Aerostack system, no other existing platform provides a full-stack system for a multirotor \ac{UAV} that is actively being supported and updated.
  Many publications provide accompanying software sources that are released without being further updated.
  By contrast, we have decided to publish and release our working system with all its components to allow members of the research community, research teams, and students to engage in \ac{UAV} research as effortlessly as possible.
  We aim to provide a thoroughly-documented open-source system to allow researchers and students to shorten their initial learning curve and to focus on their research instead of developing yet another control pipeline.
  In our case, the future continuity of our system is supported for use in the next 5+ years through our numerous activities in projects supported by European grants\footnote{\url{https://aerial-core.eu}, \url{http://rci.cvut.cz}} and by national grants\footnote{\url{http://mrs.felk.cvut.cz}}.

  The proposed platform is provided with two control designs --- extended \emph{SE(3) geometric tracking} \cite{lee2010geometric} for agile and aggressive flight, and the novel \emph{MPC controller} for stable flight using a potentially unreliable state estimate.
  However, we highlight the modularity of our platform, which can easily be extended with new control approaches as needed.
  The survey of UAV controllers provides a rich list of potentially useful control techniques \cite{nascimento2019position}.
  For example, a novel adaptive backstepping controller \cite{zhang2019robust, labbadi2019robust} may provide better performance during aggressive maneuvers, thanks to the included rotor drag compensation.
  The proposed extension to \emph{geometric tracking on SE(3)} \cite{lee2010geometric} can be further improved with remarks from \cite{lee2013nonlinear} to provide robust control to bounded uncertainties.
  Furthermore, nonlinear \ac{MPC} controllers are becoming popular \cite{nascimento2019nmpc, pereira2019nonlinear, kamel2017robust}, thanks to their inherent ability to deal with obstacle avoidance.
  However, when dealing not just with theoretical work but also with the deployment of \acp{UAV} in real-world conditions, we favor practicality over complexity.
  We therefore propose the use of relatively simple controllers (described further in \refsec{sec:reference_controller}), with well tractable performance.

  \added{
  Although the vast majority of existing control systems (all those mentioned above) and control designs for \acp{UAV} \cite{nascimento2019position} rely on the separation of the attitude and translational dynamics, methods for full-dynamics control also exist.
  Those end-to-end control designs \cite{poultney2018robust, lafflitto2018introduction} treat the full dynamics of the multirotor \ac{UAV} as one model.
  Despite the design being cutting-edge, we prefer splitting the dynamics for many practical reasons.
  We prefer the split design aspects for the ability to introspect and limit the inner states if needed.
  For example, limit the maximum acceleration or tilt produced by the controller, regardless of the outer translational control loop.
  Also, the large variance on the input (estimator noise) of an end-to-end controller can produce unbounded control outputs (moments) and states (acceleration, tilt, tilt rate) unless the control design natively supports constraining them.
  Finally, as will be discusses in this manuscript, we employ on attitude controller (\emph{geometric force tracking on SO(3))} and several \emph{outer loop} controllers (\emph{MPC}, \emph{SE(3)}, \emph{Speed tracker}) depending on the particular experimental scenario requirements.
  However, the proposed platform allows for the use of an end-to-end design, if needed.
  }

  \begin{table}
    \centering
    \def\arraystretch{0.8}
    \setlength\tabcolsep{1.5pt} 
    \begin{tabular}{c||c|c|c|c|c}
      \hline\noalign{\smallskip}
      \scriptsize platform & \begin{tabular}{@{}c@{}} \scriptsize open\\[-0.3em] \scriptsize source\end{tabular} & \scriptsize modular & \scriptsize SITL/HITL & \begin{tabular}{@{}c@{}} \scriptsize outside\\[-0.3em] \scriptsize lab\end{tabular} & \begin{tabular}{@{}c@{}} \scriptsize rate\\[-0.3em] \scriptsize input\end{tabular}\\
        \noalign{\smallskip}\hline\hline\noalign{\smallskip}
      \scriptsize \textbf{Pixhawk} & \scriptsize \textbf{SW \& HW} & \scriptsize \textbf{+} & \scriptsize \textbf{+} & \scriptsize \textbf{+} & \scriptsize \textbf{+}\\
      \scriptsize DJI & \scriptsize - & \scriptsize - & \scriptsize - \scriptsize (proprietary) & \scriptsize + & \scriptsize \textbf{-}\\
      \scriptsize Ardupilot & \scriptsize SW & \scriptsize + & \scriptsize + & \scriptsize + & \scriptsize \textbf{-}\\
      \scriptsize Parrot & \scriptsize SW & \scriptsize - & \scriptsize + & \scriptsize - & \scriptsize \textbf{-}\\
      \noalign{\smallskip}\hline
    \end{tabular}
    \caption{Comparison of commonly-used embedded flight controllers and low-level control systems. The Pixhawk flight controller was chosen due to several factors: both the hardware and software is open-source, the controller is modular enough to be used on a variety of custom multirotor platforms, Pixhawk supports both hardware- and software-in-the-loop simulation, can be used outside of laboratory conditions, and supports attitude rate input.\label{tab:embedded_flight_controllers}}
  \end{table}

  \begin{table*}
    \centering
    \def\arraystretch{0.8}
    \setlength\tabcolsep{4.0pt} 
    \begin{tabular}{c||c|c|c|c|c|c|c|c}
\noalign{\smallskip}\hline
      \small platform & \begin{tabular}{@{}c@{}} \small open\\[-0.1em] \small source\end{tabular} & \small modular & \small simulation & \begin{tabular}{@{}c@{}} \small outside\\[-0.1em] \small laboratory\end{tabular} & \begin{tabular}{@{}c@{}} \small multi-frame\\[-0.1em] \small localization\end{tabular}& \begin{tabular}{@{}c@{}} \small rate\\[-0.1em] \small output\end{tabular} & \begin{tabular}{@{}c@{}} \small software last\\[-0.1em] \small updated\end{tabular} & \small reference\\
        \noalign{\smallskip}\hline\hline\noalign{\smallskip}
      \small \small \textbf{MRS UAV system} & \small \textbf{+} & \small \textbf{+} & \small \textbf{+} & \small \textbf{+} & \small \textbf{+} & \small \textbf{+} & \small \textbf{2020} & ---\\
      \small Aerostack & \small + & \small + & \small + & \small + & \small - & \small - & \small 2020 & \small \cite{sanchez2016aerostack}\\
      \small XTDrone & \small + & \small + & \small + & \small - & \small - & \small - & \small 2020 & \small \cite{xiao2020xtdrone}\\
      \small RotorS & \small + & \small + & \small + & \small - & \small - & \small + & \small 2020 & \small \cite{furrer2016rotors}\\
      \small OpenUAV & \small + & \small - & \small + & \small - & \small - & \small - & \small 2020 & \small \cite{schmittle2018openuav}\\
      \small ReCOPTER & \small + & \small - & \small - & \small - & \small - & \small - & \small 2015 & \small \cite{abeywardena2015design}\\
      \small MAVwork & \small + & \small + & \small - & \small - & \small - & \small - & \small 2013 & \small \cite{mellado2013mavwork}\\
\noalign{\smallskip}\hline
    \end{tabular}
    \caption{Comparison of high-level open-source \ac{UAV} systems. The proposed system is extensible and modular, comes with an extensive simulation environment, is designed to be used outside of laboratory conditions, provides the novel multi-frame localization estimator, and supplies the attitude rate command to the underlying embedded flight controller.\label{tab:uav_systems}}
  \end{table*}



  \subsection{Contributions}

  The proposed system goes beyond existing systems with
  \begin{itemize}
    \item a novel bank-of-filters estimator design that overcomes challenges with diverse sensory equipment,
    \item a heading-oriented control design, devoid of ambiguous use of Euler/Tait-Bryan angles,
    \item a body/world disturbance estimation approach that does not rely on a specific state estimator design,
    \item a reliable MPC-based controller with the benefits of the nonlinear \updated{\emph{SO(3)}} force feedback,
    \item a system that can be employed with a variety of onboard localization systems and sensors,
    \item an ability to supply references in coordinate frames, which differ from the feedback loop reference frame,
    \item \added{extended trajectory generation mechanism \cite{mueller2015computationally} for generating fast trajectories through waypoints,}
    \item \added{an open-source implementation \footnote{\url{http://github.com/ctu-mrs/mrs_uav_system}} suitable for large variety of applications and scenarios,}
    \item \added{a modular design suitable for testing new control methods allowing them to be hot-swapped in mid-flight with the provided ones.}
  \end{itemize}

  The system is not only innovative, but also provides practical contributions to the community.
  The open-source implementation of the proposed platform has been tested extensively in real-world settings and in conditions of outdoor fields, in a forest, indoors, in a factory, in mines, caves and tunnels, during object manipulation, during fast and aggressive flights, and in autonomous landing on a moving platform.
  The system includes a simulation environment based on the Gazebo 3D simulator with realistic sensors and models that can be run in real time.
  The released platform is fully compatible with multiple releases of ROS (Melodic, Noetic), and is being actively used and maintained.
  The system is scalable for multiple \acp{UAV} and is well suited for research in swarming.



  \begin{figure}
    \centering
    \includegraphics[width=0.40\textwidth]{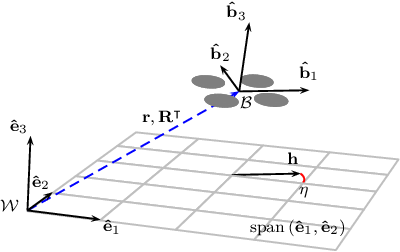}
    \caption{
      The image depicts the world frame $\mathcal{W}$ = $\{\mathbf{\hat{e}}_1$, $\mathbf{\hat{e}}_2$, $\mathbf{\hat{e}}_3\}$ in which the 3D position and the orientation of the \ac{UAV} body is expressed.
      The body frame $\mathcal{B}$ = $\{\mathbf{\hat{b}}_1$, $\mathbf{\hat{b}}_2$, $\mathbf{\hat{b}}_3\}$ relates to $\mathcal{W}$ by the translation $\mathbf{r} = \left[x, y, z\right]^{\intercal}$ and by rotation $\mathbf{R}^{\intercal}$.
      The \ac{UAV} heading vector $\mathbf{h}$, which is a projection of $\hat{\mathbf{b}}_1$ to the plane $span\left(\mathbf{\hat{e}}_1, \mathbf{\hat{e}}_2\right)$, forms the heading angle $\eta = \mathrm{atan2}\left(\mathbf{\hat{b}}_1^\intercal\mathbf{\hat{e}}_2, \mathbf{\hat{b}}_1^\intercal\mathbf{\hat{e}}_1\right) = \mathrm{atan2}\left(\mathbf{h}_{(2)}, \mathbf{h}_{(1)}\right)$.
      }
      \label{fig:coordinate_frames}
  \end{figure}



  \begin{figure*}
    \centering
    \resizebox{1.0\textwidth}{!}{
    \input{fig/tikz/pipeline_diagram.tex}
    }
    \caption{A diagram of the system architecture: \emph{Mission \& navigation} software supplies the position and heading reference ($\mathbf{r}_d$, $\eta_d$) to a reference tracker. \emph{Reference tracker} creates a smooth and feasible reference $\bm{\chi}$ for the reference feedback controller. The feedback \emph{Reference controller} produces the desired thrust and angular velocities ($T_d$, $\bm{\omega}_d$) for the Pixhawk embedded flight controller. \updated {The \emph{State estimator} fuses data from \emph{Onboard sensors} and \emph{Odometry \& localization} methods to create an estimate of the \ac{UAV} translation and rotation ($\mathbf{x}$, $\mathbf{R}$}).
    }
    \label{fig:system_architecture}
  \end{figure*}
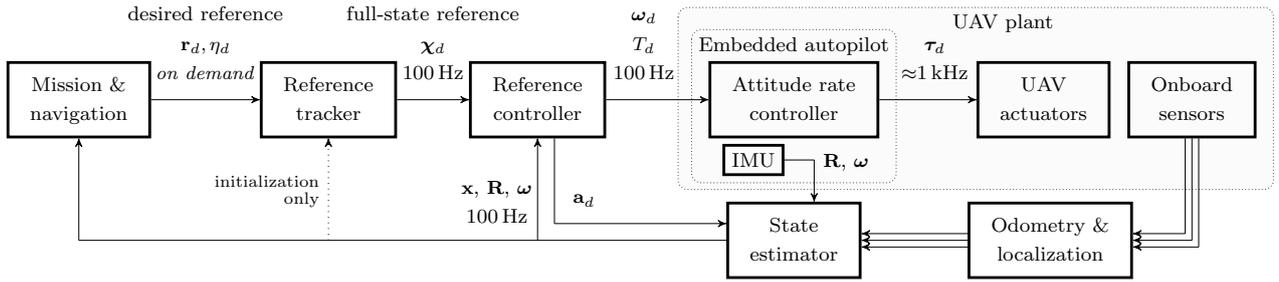



  \subsection{System Architecture \& Outline}

  We start with a description of the multirotor \ac{UAV} dynamics model (\refsec{sec:multirotor_model}), which is the foundation for further control design.
  The proposed platform consists of several interconnected subsystems, as depicted in~\reffig{fig:system_architecture}.
  \emph{Mission \& navigation} software supplies the desired trajectory, i.e., a time-parametrized sequence of the desired position and heading.
  The module is specific to any particular application of the platform (autonomous exploration, swarming, remote sensing) and is conveyed via the publications presented in \refsec{sec:pushing_the_frontiers}.
  We will therefore not focus on the module here, as numerous examples in other papers have shown where the proposed system may be applied.
  Onboard sensor data (e.g., position measurements from \ac{GPS}, velocity measurements from visual odometry) are processed by the \emph{State estimator}, which provides the \emph{Reference tracker} and the \emph{Reference controller} with hypotheses of \ac{UAV} states in all available frames of reference (\refsec{sec:state_estimation}).
  The block generates estimated states of the translational dynamics with the \ac{UAV} orientation, for all considered world frames of reference.
  One of the world frames is always selected as the \emph{main} frame, in which a feedback loop is closed by the \emph{Feedback controller} block.
  The desired trajectory is processed by a \emph{Reference tracker} (see \refsec{sec:reference_tracker}), and is then converted into a feasible, smooth, and evenly-sampled full-state control reference.
  The reference contains the desired position, its derivatives up to the jerk, the heading\footnote{The heading, as defined later in our paper, removes ambiguities caused by numerous conventions of the widely-used Tait-Bryan and Euler angles \cite{diebel2006representing}, and provides a user-friendly representation of the 4$^{\text{th}}$ controllable degree of freedom of a multirotor UAV.}, and the heading rate, supplied at \SI{100}{\hertz}.
  The reference is used by a \emph{Reference controller} (see \refsec{sec:reference_controller}) to provide feedback control of the translational dynamics and the orientation of the \ac{UAV}.
  This block creates an attitude rate $\bm{\omega}_d$ and a thrust command $T_d$, which are sent to an embedded flight controller\footnote{The proposed system is compatible with the Pixhawk flight controller, installed with PX4 firmware.}.
  We consider the underlying hardware platform already pre-configured with motors, motor speed controllers, and a basic embedded flight controller providing \emph{Attitude rate control}.
  We rely on the embedded flight controller for a backup control using a remote controller in case of a malfunction of the high-level computer.
  The flight controller encapsulates the underlying physical UAV system with motors and motor \acp{ESC} and creates 4 new controllable \ac{DOFs}: the desired angular speed around $\mathbf{\hat{b}}_1, \mathbf{\hat{b}}_2, \mathbf{\hat{b}}_3$ and the desired thrust $T_d \in \left[0, 1\right]$ of all propellers.
  This encapsulation provides an abstraction that allows us to control any standard multirotor helicopter regardless of the number of propellers and the geometry of its fuselage.
  \added{Section~\ref{sec:trajectory_generation} briefly describes a trajectory generation approach used within our system.}
  Section~\ref{sec:implementation} contains remarks on the implementation aspects of our system.
  Section~\ref{sec:experimental_evaluation} provides the results of an experimental evaluation of the control system, with emphasis on a comparison between the simulation environment and a real-world counterpart.
  Finally, examples of real-world use and application of the system for validating research, for education, and for competing in robotics competitions are presented in \refsec{sec:pushing_the_frontiers}.



  \begin{table*}
      \scriptsize
      \centering
      \noindent\rule{\textwidth}{0.5pt}
      \begin{tabular}{lll}
        $\mathbf{x}$, $\bm{\alpha}$ & vector, pseudo-vector, or tuple\\
        $\mathbf{\hat{x}}$, $\bm{\hat{\omega}}$& unit vector or unit pseudo-vector\\
        $\mathbf{\hat{e}}_1, \mathbf{\hat{e}}_2, \mathbf{\hat{e}}_3$ & elements of the \emph{standard basis} \\
        $\mathbf{X}, \bm{\Omega}$ & matrix \\
        $\mathbf{I}$ & identity matrix \\
        $x = \mathbf{a}^\intercal\mathbf{b}$ & inner product of $\mathbf{a}$, $\mathbf{b}$ $\in \mathbb{R}^3$\\
        $\mathbf{x} = \mathbf{a}\times\mathbf{b}$ & cross product of $\mathbf{a}$, $\mathbf{b}$ $\in \mathbb{R}^3$\\
        $\mathbf{x} = \mathbf{a}\circ\mathbf{b}$ & element-wise product of $\mathbf{a}$, $\mathbf{b}$ $\in \mathbb{R}^3$ \\
        $\mathbf{x}_{(n)}$ = $\mathbf{x}^\intercal\mathbf{\hat{e}}_n$ & $\mathrm{n}^{\mathrm{th}}$ vector element (row), $\mathbf{x}, \mathbf{e} \in \mathbb{R}^3$\\
        $\mathbf{X}_{(a,b)}$ & matrix element, (row, column)\\
        $x_{d}$ & $x_d$ is \emph{desired}, a reference\\
        $\dot{x}, \ddot{x}, \dot{\ddot{x}}$, $\ddot{\ddot{x}}$ & ${1^{\mathrm{st}}}$, ${2^{\mathrm{nd}}}$, ${3^{\mathrm{rd}}}$, and ${4^{\mathrm{th}}}$ time derivative of $x$\\
        $x_{[n]}$ & $x$ at the sample $n$ \\
        $\mathbf{A}, \mathbf{B}, \mathbf{x}$ & LTI system matrix, input matrix and input vector\\
        $\mathbf{A}_r, \mathbf{B}_r$ & translational LTI system and input matrices\\
        $\mathbf{A}_{\eta}, \mathbf{B}_{\eta}$ & heading LTI system and input matrices\\
        $\mathbf{A}_m, \mathbf{B}_m, \mathbf{x}_m$ & MPC system matrix, input matrix, state vector\\
        $\mathbf{Q}, \mathbf{S}$ & state MPC penalization matrices \\
        $\mathbf{x}_{max}, \dot{\mathbf{u}}_{max}$ & MPC state and slew rate constraints \\
        $p_1, p_2, p_3$ & parameters of the estimated system\\
        $a_t, b_t$ & parameters of a quadratic thrust curve\\
        \updated{\emph{SO(3)}} & 3D special orthogonal group of rotations\\
        \updated{\emph{SE(3)}} & \updated{\emph{SO(3)}}~$\times~\mathbb{R}^3$, special Euclidean group\\
        $\mathcal{L}_d$ & desired UAV plane subspace spanned by $\mathbf{\hat{b}}_{1d}$, $\mathbf{\hat{b}}_{2d}$\\
        $t$ & time, [\si{\second}]\\
        $\Delta t$ & time difference interval, [\si{\second}] \\
        $\mathcal{B}, \mathcal{W}$ & body-fixed and world frames of reference \\
        $span(\bullet)$ & the smallest vector space containing the vectors $\bullet$ \\
        $\mathbf{A} \succcurlyeq 0$ & a positive semi-definite matrix\\
        $\bullet^{\perp}$ & orthogonal complement to the vector space $\bullet$\\
      \end{tabular}%
      \begin{tabular}{lll}
        $m$, $m_e$, $m_a$ & nominal, estimated and apparent UAV mass, [\si{\kilogram}] \\
        $g$ & gravitational acceleration, [\si{\meter\per\square\second}] \\
        $f$ & total thrust force produced by the propellers, [\si{\newton}] \\
        $f_d$ & desired thrust force produced by a controller [\si{\newton}] \\
        $T$ & thrust, a collective motor speed $\in \left[0, 1\right]$\\
        $\bm{\tau}_d$ & desired individual motor speed of all $n$ motors \\
        $\eta$ & UAV heading angle, [\si{\radian}] \\
        $\mathbf{h}$ & UAV heading vector \\
        $\mathbf{H}$ & UAV heading rotation matrix \\
        $\mathbf{x}$ & estimated state vector \\
        $\bm{\chi}$ & feedback controller reference \\
        $\mathbf{r}$ & UAV position, [\si{\meter}] \\
        $\mathbf{a}_d$ & unbiased desired acceleration, [\si{\meter\per\square\second}]\\
        $\mathbf{R}_o$ & UAV orientation estimated by Pixhawk\\
        $\mathbf{R}$ & estimated UAV orientation with heading \\
        $\mathbf{R}_d^o$ & desired UAV orientation (as in \cite{lee2010geometric}) \\
        $\mathbf{R}_d$ & desired UAV orientation, heading-compliant \\
        $\bm{\omega}$ & angular velocity in $\mathcal{B}$, [\si{\radian\per\second}] \\
        $\bm{\omega}_d$ & desired UAV angular velocity in $\mathcal{B}$, [\si{\radian\per\second}]\\
        $\bm{\Omega}$ & tensor of angular velocity \\
        $\dot{\mathbf{R}}$ & UAV rotation matrix derivative  \\
        $\mathbf{d}_w$ & estimated world-frame disturbance, [\si{\newton}] \\
        $\mathbf{d}_b$ & estimated body-frame disturbance, [\si{\newton}] \\
        $\mathbf{c}_d$ & desired acceleration generated by MPC, [\si{\meter\per\square\second}] \\
        $\mathbf{k}_p,\mathbf{k}_v$ & position and velocity control gains \\
        $\mathbf{k}_{ib},\mathbf{k}_{iw}$ & body- and world-disturbance control gains \\
        $\mathbf{k}_R$ & orientation control gains \\
        $\mathbf{e}_p, \mathbf{e}_v, \mathbf{e}_R$ & position, velocity and orientation control error \\
        $\mathbf{e}$ & MPC control error \\
        $\mathcal{N}(\mu, \sigma^2)$ & normal distribution, mean $\mu$, variance $\sigma^2$ \\
      \end{tabular}
      \noindent\rule{\textwidth}{0.5pt}
      \caption{Mathematical notation, nomenclature and notable symbols.}
  \end{table*}




  \section{Multirotor aerial vehicle dynamics model}
  \label{sec:multirotor_model}

  The design of a high-performance attitude and position controller often requires an accurate model of the system.
  Here, we recall the widely-used dynamical model of a multirotor aerial vehicle~\cite{lee2010geometric}.
  Figure~\ref{fig:coordinate_frames} illustrates the coordinate frames used in this manuscript.
  For the sake of brevity, we do not explicitly annotate variables with their respective coordinate frames, since all variables with the exception of the angular velocities $\bm{\omega}$ are expressed in the world coordinate frame.
  The \ac{UAV} feedback control relies on state variables defined as:

  \vspace{0.5em}
  \noindent\begin{center}
    \begin{tabular}{l p{5cm}}
      $\mathbf{r}=\left[x, y, z\right]^\intercal$ & the position of the center of the mass of a \ac{UAV} in the world frame,\\
      $\dot{\mathbf{r}} \in \real^3$ & the velocity of the center of the mass of a \ac{UAV} in the world frame,\\
      $\ddot{\mathbf{r}} \in \real^3$ & the acceleration of the center of a mass of a \ac{UAV} in the world frame,\\
      ${\mathbf{R}} \in \updated{\emph{\text{SO(3)}}} \subseteq \real^{3\times3}$ & the rotation matrix from the body frame of a \ac{UAV} to the world frame, \newline $\mathrm{det}\,\mathbf{R} = 1$, $\mathbf{R}^\intercal = \mathbf{R}^{-1}$,\\
      $\bm{\omega} = [\omega_1, \omega_2, \omega_3]^\intercal$ & the angular velocity in the body frame of a \ac{UAV}.\\
    \end{tabular}
  \end{center}
  These states are linked by a nonlinear model, which has a translation part:
  \begin{equation}
    \updated{m\ddot{\mathbf{r}} = f\mathbf{R}\mathbf{\hat{e}}_3 - m g \mathbf{\hat{e}}_3,}
    \label{eq:model_uav_translation}
  \end{equation}
  and a rotational part
  \begin{equation}
    \dot{\mathbf{R}} =\mathbf{R}\mathbf{\Omega} ,
    \label{eq:model_uav_rotation}
  \end{equation}
  where $\mathbf{\Omega}$ is the tensor of angular velocity, under the condition $\mathbf{\Omega}\,\mathbf{v}= \bm{\omega}\times\mathbf{v}, \forall \mathbf{v} \in \mathbb{R}^3$.
  The vehicle experiences downwards gravitational acceleration with magnitude $g \in \mathbb{R}$ together with the thrust force $f$ created collectively by the propellers in the direction of $\mathbf{\hat{b}}_3$.
  However, as we are focused on non-aerobatic flight, we separately consider and estimate the azimuth of the $\mathbf{\hat{b}}_1$ axis in the world as the \ac{UAV} \emph{heading}.
  Under the condition of $|\mathbf{\hat{e}}_3^\intercal\mathbf{\hat{b}}_1| < 1$, we define the heading as
  \begin{equation}
    \eta = \mathrm{atan2}\left(\mathbf{\hat{b}}_1^\intercal\mathbf{\hat{e}}_2, \mathbf{\hat{b}}_1^\intercal\mathbf{\hat{e}}_1\right).
  \end{equation}
  The heading is a more intuitive alternative to the widely-used \emph{yaw} angle as one of the 4 controllable \ac{DOFs}.
  It is possible to use the yaw, but with the assumption that the tilt of the \ac{UAV} ($\mathrm{cos}^{\minus 1}\ \mathbf{\hat{b}}_3^{\intercal}\mathbf{\hat{e}}_3$) is low, near horizontal.
  We advice against the use of the \emph{Tait-Bryan angles} (commonly mistaken for \emph{Euler angles}), due to the overwhelming number of conventions, which often lead to misunderstanding.
  Generally, the widely-used \emph{yaw} angle (as in \emph{Euler angles}, \emph{Tait-Bryan angles} \cite{diebel2006representing}) has no direct meaning with respect to the particular orientation of any of the body axes in any of the conventions, since the final orientation also depends on the remaining two rotations (pitch, roll).
  A user would need to take the remaining part of the desired orientation (produced by the controllers) into account in the \emph{Mission \& navigation} software to properly design the desired \emph{yaw}, which leads to a \emph{chicken or egg} problem.
  We therefore, we define the heading vector by the $\hat{\mathbf{b}}_1$ axis as
  \begin{equation}
    \mathbf{h} = \left[\mathbf{R}_{(1,1)}, \mathbf{R}_{(2,1)}, 0\right]^\intercal = \left[\mathbf{b}_{1}^\intercal\mathbf{\hat{e}}_1, \mathbf{b}_{1}^\intercal\mathbf{\hat{e}}_2, 0\right]^\intercal
  \end{equation}
  and its normalized form
  \begin{equation}
    \mathbf{\hat{h}} = \frac{\mathbf{h}}{\|\mathbf{h}\|} = \left[\mathrm{cos}\,\eta, \mathrm{sin}\,\eta, 0\right]^\intercal.
  \end{equation}
  Figure~\ref{fig:coordinate_frames} illustrates the heading vector and the heading with respect to the UAV body frame.



  \section{State estimation}
  \label{sec:state_estimation}

  While the focus of this section is on estimating $\mathbf{r}$, $\dot{\mathbf{r}}$, and $\ddot{\mathbf{r}}$, the estimation of $\mathbf{R}$ and $\bm{\omega}$ can be solved individually thanks to the separation of (\ref{eq:model_uav_translation}) and (\ref{eq:model_uav_rotation}).
  From a practical standpoint, the estimation of the sub-model (\ref{eq:model_uav_translation}) can be executed on a high-level onboard computer, which has access to position/velocity measurements from onboard/external sensors.
  In contrast, the estimation of (\ref{eq:model_uav_rotation}) is better suited for an embedded flight controller with an integrated \ac{IMU}, which also handles motor mixing and the attitude rate feedback loop.
  Depending on the particular hardware, the high-level computer may not have access to the \ac{IMU} measurements at full rate without delay, and this could negatively impact the performance of the state estimator.
  We therefore, we consider the estimates of $\mathbf{R}$ (specifically, the estimate of $\hat{\mathbf{b}}_3$) and $\bm{\omega}$ as provided by an off-the-shelf embedded flight controller\footnote{We rely on the Pixhawk flight controller for attitude estimation and attitude rate control, \url{http://pixhawk.com}, \url{http://px4.io}.}.
  We rely on an attitude control loop, closed by the embedded flight controller.


  \subsection{Translational estimator model}
  \label{sec:translational_estimator_model}

  Our experience of working with \ac{UAV} estimators (both linear and nonlinear, and capable of estimating disturbances) has led us to simplify the estimator as much as possible.
  The reasons are pragmatic: tuning complex models and the respective estimators becomes impractical with increasing model dimensionality, increasing numbers of possible sensor configurations and \ac{UAV} types, and due to the range of experimental conditions.
  We aimed to simplify the estimation process as much as possible by leveraging the specific decoupled structure of the multirotor \ac{UAV} model and utilizing the ability of the proposed controllers to estimate disturbances \added{in the sense of external force acting on the vehicle}.
  We therefore, we model the translation dynamics of the \ac{UAV} as a point mass in 3D with an additional degree of freedom in the heading angle, $\eta$.
  The considered state vector $\mathbf{x}$ for the \emph{high-level} estimation of (\ref{eq:model_uav_translation}) consists of the components of position $\mathbf{r}$, its first two derivatives, and the heading $\eta$ with its first derivative as
  \begin{equation}
    \mathbf{x} = \left[x, \dot{x}, \ddot{x}, y, \dot{y}, \ddot{y}, z, \dot{z}, \ddot{z}, \eta, \dot{\eta}\right]^\intercal.
    \label{eq:state_vector_estimation}
  \end{equation}

  We model the high-level dynamics as a discrete and decoupled \ac{LTI} system
  \begin{equation}
    \mathbf{x}_{[t]} = \m{A}\mathbf{x}_{[t-1]} + \m{B}\mathbf{u}_{[t]},
    \label{eq:linear_model}
  \end{equation}
  with 4 independently estimated subsystems expressed together by matrices $\mathbf{A}$ and $\mathbf{B}$ as
  \begin{align*}
    &\begin{smallmatrix}
      \mathbf{A}\left(\Delta t, p_1, p_2, p_3\right)
    \end{smallmatrix} =\\
    &= \left[\begin{smallmatrix}
      \mathbf{A}_r\left(\Delta t, p_1\right) & \mathbf{0}                              & \mathbf{0}                              & \mathbf{0} \\
      \mathbf{0}                              & \mathbf{A}_r\left(\Delta t, p_1\right) & \mathbf{0}                              & \mathbf{0} \\
      \mathbf{0}                              & \mathbf{0}                              & \mathbf{A}_r\left(\Delta t, p_2\right) & \mathbf{0} \\
      \mathbf{0}                              & \mathbf{0}                              & \mathbf{0}                              & \mathbf{A}_{\eta}\left(\Delta t, p_3\right)
    \end{smallmatrix}\right], \numberthis
    \label{eq:model_A}
  \end{align*}
  \begin{align*}
    &\begin{smallmatrix}
      \mathbf{B}\left(\Delta t, p_1, p_2, p_3\right)
    \end{smallmatrix} = \left[\begin{smallmatrix}
      \mathbf{B}_r\left(\Delta t, 1-p_1\right)\\
      \mathbf{B}_r\left(\Delta t, 1-p_1\right)\\
      \mathbf{B}_r\left(\Delta t, 1-p_2\right)\\
      \mathbf{B}_{\eta}\left(\Delta t, 1-p_3\right)\\
    \end{smallmatrix}\right]. \numberthis
    \label{eq:model_B}
  \end{align*}
  Matrices $\mathbf{A}_r$ and $\mathbf{B}_r$ are the sub-system matrices for the translation part of the model:
  \begin{equation}
    \begin{split}
      \begin{smallmatrix}
        \mathbf{A}_r\left(\Delta t, a\right)
      \end{smallmatrix} = \left[\begin{smallmatrix}
        1 & \Delta t & \frac{\Delta t^2}{2}\\
        0 & 1 & \Delta t\\
        0 & 0 & a
      \end{smallmatrix}\right], \begin{smallmatrix}
        \mathbf{B}_r\left(\Delta t, b\right)
      \end{smallmatrix} = \left[\begin{smallmatrix}
        0\\
        0\\
        b\\
      \end{smallmatrix}\right],
    \end{split}
    \label{eq:model_translation}
  \end{equation}
  and $\mathbf{A}_{\eta}$ and $\mathbf{B}_{\eta}$ are the sub-system matrices for the heading part of the model:
  \begin{equation}
    \begin{split}
      \begin{smallmatrix}
        \mathbf{A}_{\eta}\left(\Delta t, a\right)
      \end{smallmatrix} = \left[\begin{smallmatrix}
        1 & \Delta t \\
        0 & a
      \end{smallmatrix}\right], \begin{smallmatrix}
        \mathbf{B}_{\eta}\left(\Delta t, b\right)
      \end{smallmatrix} = \left[\begin{smallmatrix}
        0\\
        b\\
      \end{smallmatrix}\right],
    \end{split}
    \label{eq:model_heading}
  \end{equation}
  with $\Delta t$ being the sampling step of the estimator, and with $p_1$, $p_2$, $p_3$ being the 1$^{\mathrm{st}}$ order transfer parameters for the horizontal, vertical and heading subsystems respectively.
  This model has only three free parameters (assuming that both horizontal axes behave identically), which which simplifies its tuning and allows it to be reused between various \ac{UAV} platforms without changes.
  The decoupling of the system to (\ref{eq:model_translation}) and (\ref{eq:model_heading}) is used during implementation to speed up the computations thanks to operations with smaller matrices.

  \subsubsection{System input}

  System input $\mathbf{u}$ consists of the \emph{unbiased desired acceleration} $\mathbf{a}_d$.
  As discussed later in \refsec{sec:unbiased_desired_acceleration}, the controllers report on the desired acceleration caused by their control output.
  However, the controllers are required to supply the desired unbiased acceleration, i.e., without compensation for gravity acceleration, integrated \emph{body} and \emph{world} \added{force} disturbances, and the estimated \ac{UAV} mass difference.
  All the biases compensated by our controllers are subtracted from the desired acceleration, thanks to their physical dimension being convertible into acceleration.
  The heading subsystem is left without an input, since measurement corrections from embedded gyroscopes (see \refsec{sec:angular_rate_to_heading}) are more than sufficient to maintain a stable and quickly-converging estimate.

  \subsubsection{Sources of measurement}

  We often work with a very diverse set of onboard sensors and localization systems.
  Some systems directly provide us with 3D \ac{UAV} position and heading, e.g., 3D visual and laser \ac{SLAMs} \cite{qin2018vins, zhang2014loam}, which can be directly fused into the position and heading state of our filters.
  When a sensory system provides only a 2D (horizontal) position measurement, e.g., the \ac{GNSS} system or a 2D laser \ac{SLAM} \cite{kohlbrecher2011flexible}, we use an additional measurement of \ac{UAV} height above the ground provided by down-facing rangefinder.
  Some systems may provide us only with a velocity measurement, e.g., an optic flow algorithm\footnote{\url{http://github.com/ctu-mrs/mrs_optic_flow}}.
  Optic flow measurements can be used for an odometry estimate of the position and heading when coupled with a height sensor.
  3D \ac{LiDAR} SLAM might also provide us with odometry velocity measurements (e.g., from a scan-matching algorithm), and absolute position measurements.
  Heading estimation fuses measured angular velocity $\bm{\omega}$ supplied by an \ac{IMU}.
  The magnetometer is fused when flying with the use of a \ac{GNSS} localization system.

  \subsection{Linear Kalman Filter}

  The dynamic model is estimated by a recursive discrete \ac{LKF}.
  This estimator, coupled with the model \eqref{eq:linear_model}, exhibits stable and fast tracking of the states of the translational dynamics, under the condition that the reference controller is capable of calculating and compensating for biases such as external force disturbance or internal input offset (see \refsec{sec:reference_controller}).
  Under these conditions, the use of more complex nonlinear filters, such as the \ac{EKF} or the \ac{UKF}, would not provide us with the desired computational-cost benefit.

  In our experience, it is simpler and more practical to utilize controllers with this property (potentially using any source of the UAV state) rather than to build complex (nonlinear) estimators, which would estimate the biases themselves.
  With this approach, any source of the UAV state can be used, even substituting the proposed estimator.
  The sources of measurements and the estimators that are used may change from platform to platform within a laboratory, even to the extent of not using an onboard estimator at all (e.g., when using the external motion capture system).
  For such situations, \added{force} disturbance estimation is provided by our control pipeline.
  If the proposed platform were to have relied on an estimator capable of estimating \added{external force} disturbances, this substitution would not have been possible.
  Thus, this choice keeps the platform more universal.



  \subsection{Updating the UAV orientation with the estimated heading}

  Since we estimate the \ac{UAV} heading separately, the original \ac{UAV} rotation $\mathbf{R}_o = \left[\mathbf{\hat{b}}_{1o}, \mathbf{\hat{b}}_{2o}, \mathbf{\hat{b}}_{3o}\right]$ supplied by the embedded flight controller is modified by generating new body frame vectors $\mathbf{\hat{b}}_1$, $\mathbf{\hat{b}}_2$, and $\mathbf{\hat{b}}_3$.
  For that, the original heading $\eta_o$ is firstly calculated as
  \begin{equation}
    \eta_o = \mathrm{atan2}\left(\mathbf{\hat{b}}_{1o}^\intercal\mathbf{\hat{e}}_2, \mathbf{\hat{b}}_{1o}^\intercal\mathbf{\hat{e}}_1\right).
  \end{equation}
  Then the difference between the original heading and the estimated heading is calculated:
  \begin{equation}
    \Delta\eta = \eta - \eta_o,
  \end{equation}
  and the original orientation is rotated around $\mathbf{\hat{e}}_3$ by the heading difference:
  \begin{equation}
    \mathbf{R} = \left[\begin{smallmatrix}
      \mathrm{cos}\Delta\eta & \mathrm{-sin}\Delta\eta & 0 \\
      \mathrm{sin}\Delta\eta & \mathrm{cos}\Delta\eta & 0 \\
      0 & 0 & 1
    \end{smallmatrix}\right] \mathbf{R}_o.
  \end{equation}



  \subsection{Fusing angular rates into the heading rate state}
  \label{sec:angular_rate_to_heading}

  Although $\omega_3$ (the yaw rate) is often treated as the heading rate, it is applicable only as an approximation under small tilts: $\angle (\mathbf{\hat{b}}_3, \mathbf{\hat{e}}_3) \lessapprox 10^{\circ}$.
  In general, all the components of an arbitrary angular speed $\bm{\omega}$ contribute to the resulting heading rate.
  To obtain the heading rate, we first apply \eqref{eq:model_uav_rotation} to obtain $\dot{\mathbf{R}}$, the first time derivative of the rotational matrix.
  Components $\dot{\mathbf{R}}_{(1,1)}$, $\dot{\mathbf{R}}_{(2,1)}$, which represent the rate of change of $\mathbf{\hat{b}}_1$ along $\mathbf{\hat{e}}_1$, $\mathbf{\hat{e}}_2$, respectively, are extracted and are used to evaluate the total differential of $\mathrm{atan2}()$ in the current orientation $\mathbf{R}$ to obtain the heading rate:
  \begin{equation}
    \dot{\eta} = \frac{-\mathbf{R}_{(2,1)}}{\mathbf{R}_{(1,1)}^2 + \mathbf{R}_{(2,1)}^2}\dot{\mathbf{R}}_{(1,1)} + \frac{\mathbf{R}_{(1,1)}}{\mathbf{R}_{(1,1)}^2 + \mathbf{R}_{(2,1)}^2}\dot{\mathbf{R}}_{(2,1)}.
  \end{equation}
  As with most heading-related operations, this operation is only feasible if $|\mathbf{\hat{e}}_3^\intercal\mathbf{\hat{b}}_1| < 1$.



  \subsection{Bank of filters for multiple hypotheses}

  With the individual filter structure defined, we now establish a bank of Kalman filters $\mathcal{K}=\{K_1, K_2,..., K_n\}$.
  The bank of filters allows for simultaneous estimation of the \ac{UAV} state from various combinations of onboard sensors, without necessarily combining all the measurements ($\mathbf{z} \in \real^m$, where $m$ is the number of measured states) into a single hypothesis.
  This type of separation is beneficial for many applications, e.g., for transitions from one sensory system to another (e.g., \ac{GNSS} $\rightarrow$ indoor SLAM), for running multiple instances of one filter with different parameters, or for maintaining a backup estimator to facilitate emergency landing.
  Each filter maintains its hypothesis $\mathbf{x}_n$, covariance $\mathbf{\Sigma}_n$, and is corrected by a different set of measurements $\mathbf{z}_n \subseteq \mathbf{z}$.

\begin{figure}
    \centering
    \resizebox{1.0\columnwidth}{!}{
      \input{fig/tikz/bank_of_filters.tex}
    }
    \caption[Bank of filters]{
      The bank of filters $\mathcal{K}=\{K_1,K_2,...,K_n\}$.
      The filters simultaneously estimate $\mathbf{x}_1,\mathbf{x}_2,...,\mathbf{x}_n$.
      The output hypothesis is chosen by the \emph{arbiter}.
    }
    \label{fig:bank_of_filters}
  \end{figure}
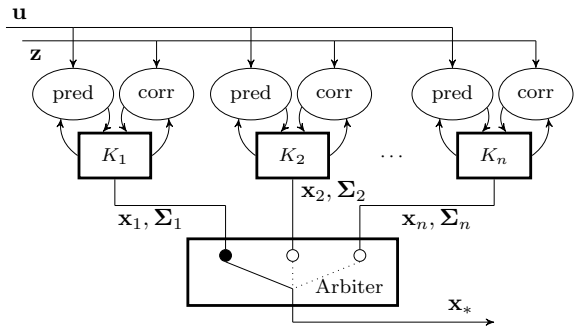

  Multiple hypotheses $\mathbf{x}_1, \mathbf{x}_2, ..., \mathbf{x}_n$ with covariances $\m{\Sigma}_1, \m{\Sigma}_2, ..., \m{\Sigma}_n$ are estimated by the respective filters $K_1,K_2,...,K_n$ as depicted in \reffig{fig:bank_of_filters}.
  An \emph{arbiter} chooses one of the available hypotheses that is being outputted as the current state estimate.
  The \emph{arbiter} selects/changes the current filter and its corresponding hypothesis $\mathbf{x}_{\ast}$ with covariance $\Sigma_{\ast}$ by one of several criteria:
  \begin{itemize}
    \item a request for a particular filter by the \emph{Mission \& navigation} part of the pipeline,
    \item the current filter becomes unreliable,
    \item $\mathbf{x}_{\ast} = \mathbf{x}_k = \operatorname*{argmin}_{\mathbf{x}} \mathrm{trace}\left(\bm{\Sigma}_k\right)$ otherwise.
  \end{itemize}
  Whenever the \emph{arbiter} switches the output, the coordinates of the \ac{UAV} in the world frame change, although the physical manifestation of the \ac{UAV} has not moved.
  This switch is treated by the rest of the control pipeline as a sudden change between frames of reference; the change in numerical values can be arbitrary.
  Any internal states of trackers and controllers are recalculated to the new frame of reference.
  Therefore, the switch is not apparent to an outside observer as the transition is perfectly smooth.

  The use of multiple hypotheses instead of fusing all measurements in a single filter provided the motivation for the bank of filters approach.
  Let us explain the problem with a practical example: fusing two \ac{GNSS} signal sources --- a classical GPS, and a differential \ac{RTK} GPS.
  Both sources of data localize the UAV within the same coordinate system.
  However, each source has a different level of accuracy (the measurements can differ by several meters), and the \ac{RTK} system may not be available all the time due to the physical limitations of the system.
  Fusing both systems into a single hypothesis creates a problem.
  For example, when the precise \ac{RTK} data starts to be fused (possibly after some time of inaccessibility, or because the \ac{RTK} system has just been activated during the flight), the hypothesis starts getting corrected.
  The correction step may introduce an innovation in the order of several meters, which produces state changes within the hypothesis that do not follow the model and do not respect any state constraints.
  More importantly, the motion of the hypothesis does not correspond to any real motion of the \acp{UAV}.
  This state convergence towards newly-fused measurements subsequently creates motion of the UAV due to the increased control error.
  However, this UAV motion is unplanned (and undesired), since it is not governed by feedforward action.
  In extreme cases, the sudden control error may even saturate the feedback controller and could endanger the UAV, as shown experimentally in \reffig{fig:plot_odometry_jump}.
  Any physical motion of the UAV should be produced by a desired and planned action, not by a state estimator suddenly shifting a hypothesis.

  In contrast, the same situation is handled here by a bank of filters.
  We would consider two separate estimators, one fusing \ac{GPS} and the other fusing both \ac{GPS} and \ac{RTK} \ac{GPS}.
  The control pipeline can be switched on demand to use any independent hypothesis, to recalculate all its inner states from one to another, and to treat both hypotheses as independent coordinate systems.
  Thus when the \emph{active} estimator is switched, the UAV does not move in a physical world, but its coordinates (and the coordinates of a control reference) will jump.
  It is then for the \emph{Mission \& navigation} part of the pipeline to decide whether the new coordinates within the new coordinate should be adjusted by generating a new control command.
  However, this multi-hypothesis structure requires the presence of an arbiter.
  The arbiter needs to switch the system automatically from the \ac{RTK} \ac{GPS} estimator to the \ac{GPS} estimator when the \ac{RTK} corrections become unavailable.

  The multiple hypothesis system also handles scenarios where sensors do not appear within the same frame of reference, e.g., an onboard visual-based SLAM and a \ac{GPS} localization system.
  Moreover, maintaining transformations between all the frames of reference allows us to close the feedback loop using the best estimator for control performance while generating references in other frames of reference.

  \section{Feedforward tracking and reference generation}
  \label{sec:reference_tracker}

  A \emph{Reference tracker} provides a feedforward control command and a reference to a \emph{Feedback controller} within the pipeline (see \reffig{fig:system_architecture}).
  An input to the \emph{Reference tracker} might be a 3D position and a heading reference $\left(\mathbf{r}_d, \eta_d\right)$, or a reference trajectory $\left\{\left(\mathbf{r}_d, \eta_d\right)_1, \left(\mathbf{r}_d, \eta_d\right)_2, \hdots ,\left(\mathbf{r}_d, \eta_d\right)_k\right\}$ from the \emph{Mission \& navigation} block.

  \subsection{MPC Tracker for normal flight}

  Feedforward trajectory tracking serves a crucial role in supplying a smooth and feasible reference for feedback controllers.
  The control reference consists of states of the differentially-flat translational dynamics (position, velocity, acceleration, jerk) as well as the heading and the heading rate:
  \begin{equation}
    \bm{\chi} = \left[x, \dot{x}, \ddot{x}, \dot{\ddot{x}}, y, \dot{y}, \ddot{y}, \dot{\ddot{y}}, z, \dot{z}, \ddot{z}, \dot{\ddot{z}}, \eta, \dot{\eta}\right]^\intercal.
    \label{eq:state_vector_reference}
  \end{equation}
  Our trajectory tracking approach, originally published in \cite{baca2018model}, utilizes linear \ac{MPC} for controlling a virtual \ac{UAV} model in real time.
  The linear \ac{MPC} controls the states of the virtual model (which behaves ideally).
  States of the virtual model are then sampled on demand, and are given to the \emph{feedback controller} as a reference.
  The linear \ac{MPC} produces optimal state transients in real time while satisfying state constraints.
  The \emph{\ac{MPC} tracker} creates a full-state reference $\bm{\chi}$ at \SI{100}{\hertz} either from a single 3D reference $\left(\mathbf{r}_d, \eta_d\right)$ or from a time-parametrized reference trajectory $\left\{\left(\mathbf{r}_d, \eta_d\right)_1, \left(\mathbf{r}_d, \eta_d\right)_2, \hdots ,\left(\mathbf{r}_d, \eta_d\right)_k\right\}$ sampled at arbitrary sampling rate.

  \subsection{Take-off and landing}
  \label{sec:trajectory_tracking_landoff}

  Take-off and landing can generally be solved by the same tracker as other situations during a routine flight.
  However, we separate the trajectory generation for take-off and for landing in order to increase safety.
  Safety concerns arise in the take-off phase, since the UAV can get entangled in ground foliage when taking off outdoors.
  When such a situation occurs, significant control errors can arise quickly, forcing the feedback controller into aggressive actions.
  We solve this with an admittance tracking mechanism, which saturates the movement of the control reference $\bm{\chi}$ beyond a set distance from the current UAV state $\mathbf{x}$.
  Automatic landing is performed by setting the altitude coordinate of the control reference below the estimated altitude of the UAV, which serves the same purpose and allows landing even when the altitude of the ground level is unknown.

  \subsection{Speed tracking for aerial swarming}
  \label{sec:speed_tracker}

  Aerial swarming, which will be briefly introduced in \refsec{sec:uav_swarms_and_formations}, imposes special requirements on control reference generation.
  2D swarming approaches often require classical tracking of the desired altitude and heading, but the horizontal motion may be dictated by the desired velocity or by the desired acceleration.
  We provide a specialized tracking approach that allows us to bypass desired states within the control reference $\bm{\chi}$, and to specify only the states that the swarming mechanism requires.
  The controllers, which will be described in the following section, use only the specified portion of the control reference to calculate the feedback.



  \section{Feedback Control}
  \label{sec:reference_controller}

  The \emph{Feedback controller} is a crucial component within the pipeline (see \reffig{fig:system_architecture}) for controlling the flight dynamics around an unstable equilibrium point of the UAV system.
  The task of the controller is to minimize the control error around the desired control reference $\bm{\chi}$ (provided by the \emph{Reference tracker} block) and to supply \updated{low-level control reference to the embedded} \emph{Attitude rate controller}.
  The \added{low-level control reference} produced by a controller within our pipeline consists of the desired intrinsic angular velocities of the \ac{UAV} body $\bm{\omega}_d \in \mathbb{R}^3$ and the desired collective motor speed $T_d~\in~\left[0, 1\right]$.
  This section focuses on the development of two feedback control approaches.
  Each of the approaches serves a particular purpose in various of field experimentation scenarios.
  The first purpose is an extension of the \emph{SE(3) geometric state feedback} \cite{lee2010geometric}.
  This controller is well-suited for fast and agile maneuvers, as well as for precise control.
  However, both the \ac{UAV} state estimate and the reference need to be continuous, smooth, and are assumed to follow the model.
  The second controller that we propose is a combination of a linear \ac{MPC} with a nonlinear \updated{\emph{SO(3)}} force tracking feedback.
  This controller is designed to provide stable feedback even when the \ac{UAV} state estimate is noisy or unreliable, or when state constraints need to be imposed on the control level.

  As shown in \reffig{fig:system_architecture}, our architecture is a cascade-based control loop.
  Cascade-based control architectures are based on the singular perturbation theory \cite{roussel2019nonlinear}, commonly known as the principle of time-scale separation.
  This approach assumes that the inner loop (in our case the attitude control) is exponentially stable and that the inner loop bandwidth is greater than the dynamics of the outer loop.
  So the controller of the outer loop can be designed without considering the dynamics of the inner loop.
  This assumption holds, since the attitude rate control loop within the PX4 firmware is executed at the utmost rate, with all new data from the embedded \ac{IMU}.


  \subsection{\updated{\emph{SO(3)}} geometric force tracking}
  \label{sec:force_tracking}

  We base our work on the geometric tracking controller proposed in \cite{lee2010geometric}.
  Specifically, we utilize the force tracking part of their approach.
  Given a desired force $\mathbf{f}_d$ to be acting on the UAV, and a desired heading vector
  \begin{equation}
    \mathbf{\hat{h}}_d =  \left[\mathrm{cos}\,\eta_d, \mathrm{sin}\,\eta_d, 0\right]^\intercal,
  \end{equation}
  we define a desired orientation matrix $\mathbf{R}_d$.
  The originally published way of constructing $\mathbf{R}_d$ is feasible; however, it does not maintain heading $\eta$ during maneuvers.
  We therefore, we also propose a different approach explicitly designed to facilitate heading angle control.

  \subsubsection{Original structure of desired orientation}

  The matrix
  \begin{equation}
    \mathbf{R}_d^o = \left[\mathbf{\hat{p}}_{1d}, \mathbf{\hat{p}}_{2d}, \mathbf{\hat{b}}_{3d}\right],
  \end{equation}
  which is composed of vectors
  \begin{equation}
    \begin{aligned}[]
      \mathbf{\hat{b}}_{3d} &= \frac{\mathbf{f_d}}{\|\mathbf{f_d}\|}, & \mathbf{\hat{p}}_{2d} &= \frac{\mathbf{\hat{b}}_{3d}\times \mathbf{\hat{h}}_d}{\|\mathbf{\hat{b}}_{3d}\times \mathbf{\hat{h}}_d\|}, & \mathbf{\hat{p}}_{1d} &= \mathbf{\hat{p}}_{2d} \times \mathbf{\hat{b}}_{3d},
    \end{aligned}
  \end{equation}
  maintains the desired force vector as the direction of the $\mathbf{\hat{b}}_{3d}$ axis, and finds $\mathbf{\hat{p}}_{1d}$ as the orthogonal projection of $\mathbf{\hat{h}}_d$ to the subspace
  \begin{equation}
    \mathcal{L}_d = span\left(\mathbf{\hat{b}}_{3d}\right)^{\perp}.
  \end{equation}
  However, the heading is not preserved in this case (the azimuth of the $\mathbf{\hat{p}}_{1d}$ axis is not generally equal to the azimuth of $\mathbf{\hat{h}}_d$).

  \subsubsection{Heading-compliant desired orientation}

  We tackle the heading control by constructing the desired orientation matrix as
  \begin{equation}
    \mathbf{R}_d = \left[\mathbf{\hat{b}}_{1d}, \mathbf{\hat{b}}_{2d}, \mathbf{\hat{b}}_{3d}\right],
  \end{equation}
  by finding $\mathbf{\hat{b}}_{1d}$ as an oblique projection of $\mathbf{\hat{h}}_d$ in the direction of $\mathbf{\hat{e}}_{3}$ to the subspace $\mathcal{L}_d$.
  This projection is constructed as
  \begin{equation}
    \mathbf{b}_{1d} = \mathbf{O}(\mathbf{P}^\intercal\mathbf{O})^{-1}\mathbf{P}^\intercal,\ \mathbf{\hat{b}}_{1d} = \frac{\mathbf{b}_{1d}}{\|\mathbf{b}_{1d}\|},
  \end{equation}
  where $\mathbf{O} \in \mathbb{R}^{3 \times 2}$ is the orthogonal projector to $\mathcal{L}_d$ (constructed, e.g., as the first two columns of $\mathbf{I} - \mathbf{\hat{b}}_{3d}\mathbf{\hat{b}}_{3d}^\intercal$),
  and $\mathbf{P} = \left[\mathbf{\hat{e}}_1, \mathbf{\hat{e}}_2\right]$ is the orthogonal basis of the world xy-plane.
  The $\hat{\mathbf{b}}_{2d}$ axis, is at last constructed as
  \begin{equation}
    \hat{\mathbf{b}}_{2d} = \mathbf{\hat{b}}_{3d}\times \mathbf{\hat{b}}_{1d}.
  \end{equation}
  Both $\mathbf{R}_d$ and $\mathbf{R}_d^o$ can be obtained under the assumption that $\mathbf{\hat{b}}_{3d} \nparallel \mathbf{\hat{h}}_d$.
  Both options are valid, and their selection should be carefully considered.
  We prefer the $\mathbf{R}_d^o$ option, due to the consistency of the resulting reference with the heading feedforward control described further in \refsec{sec:heading_rate_as_yaw_rate}, and \ref{sec:parasitic_heading_rate}.

  Given $\mathbf{R}_d$, we express the rotation error according to \cite{lee2010geometric} as
  \begin{equation}
    \mathbf{E}_R = \frac{1}{2}\left(\mathbf{R}_d^\intercal\mathbf{R} - \mathbf{R}^\intercal\mathbf{R}_d\right), \mathbf{e}_R = \left[\mathbf{E}_{(3,2)}, \mathbf{E}_{(1,3)}, \mathbf{E}_{(2,1)}\right]^\intercal.
  \end{equation}
  Finally, the desired angular rate is obtained as
  \begin{equation}
    \bm{\omega}_d = -\mathbf{k}_R\circ \mathbf{e}_R + \bm{\omega}_j - \bm{\omega}_c,
    \label{eq:desired_angular_rate}
  \end{equation}
  where $\mathbf{k}_R$ are the rotation control gains, $\bm{\omega}_j$ is the feedforward attitude rate caused by the desired jerk $\dot{\ddot{\mathbf{r}}}_d$, and $\bm{\omega}_c$ is the \emph{parasitic} heading rate compensation described further in \refsec{sec:parasitic_heading_rate}.
  The feedforward attitude rate is constructed as
  \begin{equation}
    \bm{\omega}_j = \frac{\|\mathbf{f}_d\|}{m_e}\left[\mathbf{\hat{e}}_3\right]_{\times}^\intercal\mathbf{R}_d^\intercal\dot{\ddot{\mathbf{r}}}_d,
  \end{equation}
  where $\|\mathbf{f}_d\|/m_e\,\left[\si{\meter\per\second\squared}\right]$ is the effective thrust, $m_e\,\left[\si{\kilogram}\right]$ is the estimated mass of the vehicle, $\left[\mathbf{\hat{e}}_3\right]_{\times}$ is the cross-product matrix satisfying the condition $\left[\mathbf{\hat{e}}_3\right]_{\times}\mathbf{v} = \mathbf{\hat{e}}_3 \times \mathbf{v}, \forall \mathbf{v} \in \mathbb{R}^3$.
  The final control output is the desired attitude rate $\bm{\omega}_d$ and the desired thrust force $f_d = \mathbf{f}_d^\intercal\mathbf{\hat{b}}_3$.



  \subsection{Applying the reference heading rate as the feedforward yaw rate}
  \label{sec:heading_rate_as_yaw_rate}

  As mentioned in \refsec{sec:reference_tracker}, the reference trackers output the heading $\eta$ and its derivative $\dot{\eta}$ as a reference.
  Using the heading rate for feedforward in \eqref{eq:desired_angular_rate} requires converting it to the desired yaw rate $\omega_{3d}$ (the yaw rate is the 4$^{\mathrm{th}}$ independently-controllable intrinsic \ac{DOF}, which does not influence the translational dynamics).
  First, we define the derivative of heading vector $\mathbf{h}$ as
  \begin{equation}
    \dot{\mathbf{h}} = \left[0, 0, \dot{\eta}\right]^\intercal \times \mathbf{h}.
  \end{equation}
  Then we define the orthogonal projector $\mathbf{P}$ on the linear subspace spanned by $\dot{\mathbf{h}}$.
  However, it is vital to define $\mathbf{P}$ even when $\dot{\mathbf{h}} = \mathbf{0}$.
  One option is:
  \begin{equation}
    \mathbf{P} = \left(\mathbf{\hat{e}}_3 \times \mathbf{\hat{h}}\right)\left(\mathbf{\hat{e}}_3 \times \mathbf{\hat{h}}\right)^\intercal.\\
  \end{equation}
  Then we project the orthogonal basis of the subspace spanned by the derivative of $\mathbf{\hat{b}}_1$, which is consequently $\mathbf{\hat{b}}_2$, on the subspace spanned by $\dot{\mathbf{h}}$:
  \begin{equation}
    \mathbf{p} = \mathbf{P}\mathbf{\hat{b}}_2.
  \end{equation}
  Now we find a scaling factor $k$ between $\dot{\mathbf{h}}$ and $\mathbf{p}$
  \begin{equation}
    k = \text{sign} \left(\dot{\mathbf{h}}^\intercal\mathbf{p}\right)\frac{\|\dot{\mathbf{h}}\|}{\|\mathbf{p}\|},~\text{for}~\|\mathbf{p}\| \neq 0,
  \end{equation}
  which is applied to recreate the desired derivative of $\mathbf{\hat{b}}_1$ as $k\mathbf{\hat{b}}_2$.
  Thus, the angular velocity around $\mathbf{\hat{b}}_3$ is
  \begin{equation}
    \omega_{3d} = k.
  \end{equation}



  \subsection{Compensating for the parasitic heading rate}
  \label{sec:parasitic_heading_rate}

  The desired angular rate $\bm{\omega}_d$ obtained from the force tracking approach, can influence the resulting heading rate $\dot{\eta}$.
  This can easily be observed while flying a dynamic trajectory with a constant desired heading.
  The control law \eqref{eq:desired_angular_rate} inevitably creates angular velocities around $\mathbf{\hat{b}}_1$ and $\mathbf{\hat{b}}_2$ that are being reflected in $\dot{\eta}$.
  These \emph{disturbances} will be counteracted by the feedback.
  However, feedback corrections are made after a control error has occurred, and this makes them appear too late during aggressive maneuvers.
  We compensate for them in advance by calculating the \emph{parasitic} heading rate created by the $\mathbf{\hat{b}}_1$ and $\mathbf{\hat{b}}_2$ rotations, similarly as in \refsec{sec:angular_rate_to_heading}.
  In addition, as in \refsec{sec:heading_rate_as_yaw_rate}, the heading rate is converted to the intrinsic yaw rate, $\omega_{3c}$, (the angular velocity around $\mathbf{\hat{b}}_3$), which is then added back to \eqref{eq:desired_angular_rate} as $\bm{\omega}_c = \left[0, 0, \omega_{3c}\right]^\intercal$.



  \subsection{Converting the desired thrust force to thrust}

  The motor speed is often controlled by dedicated modules, i.e. by \acp{ESC}.
  The input to an \ac{ESC} is typically a desired motor speed scaled linearly between $\left[0, 1\right]$, which represents the range from the minimum speed to the maximum speed.
  The desired thrust force $f_d = \|\mathbf{f}_d\|$ therefore needs to be converted to the output collective thrust $T_d \in \left[0, 1\right]$.
  The simplest but still effective thrust model relies on the approximate relationship between the produced force $f$ and the angular rate $\omega$ of a motor: $f \propto \omega^2$.
  We therefore model the thrust as
  \begin{equation}
    T_d = a_t\sqrt{f_d} + b_t,
    \label{eq:thrust_curve}
  \end{equation}
  where $a_t$ and $b_t$ are parameters of a quadratic thrust curve.
  The parameters are obtained empirically by the least-square fit on experimentally obtained data --- tuples of the thrust and the mass ($T_h$, $m$) --- using equation \eqref{eq:thrust_curve} as a hover thrust curve
  \begin{equation}
    T_h = a_t\sqrt{mg} + b_t.
    \label{eq:hover_thrust_curve}
  \end{equation}
  The accuracy of the thrust model is important for the correct calculation of the applied thrust, which has an influence on the overall control performance.
  The inversion of \eqref{eq:hover_thrust_curve}
  \begin{equation}
    m_a = \frac{1}{g}\left(\frac{T_d - b_t}{a_t}\right)^2
  \end{equation}
  is also used to deduce the current apparent mass $m_a$ based on the currently-used thrust output $T_d$, namely during landing for automatic touchdown detection.



  \subsection{Disturbance estimation}

  Various effects can cause a steady state control error in the position of the \ac{UAV}: $\mathbf{e}_p = \mathbf{r} - \mathbf{r}_d$.
  External disturbances that appear to be fixed within the world frame (e.g. wind) occur together with disturbances that are tied to the body frame of the \ac{UAV}, e.g. air drag.
  Also, a miscalibrated artificial horizon (accelerometer bias) will generally cause control errors, which can be observed and estimated as a steady-state body disturbance.
  \added{
    \ac{UAV} disturbance estimation is a well-established field with a prominent use of non-linear state estimators (often \ac{EKF} or \ac{UKF}) \cite{hentzen2019disturbnace, cayero2019optimal, benallegue2008high, aboudonia2018composite} with a linear estimator being also suitable, as shown in our prior work \cite{baca2016embedded}.
    It is common to account for the disturbances as dedicated observable states within the \ac{UAV} model and use an observer to estimate them.
    In this work, we strafe from the typical disturbance estimation process within the main state estimator of the \ac{UAV}.
    Estimating both the world and the body disturbances simultaneously would require a non-linear state estimator (to separate them within the model \cite{aboudonia2018composite}).
    Since we utilize the simpler \ac{LKF}, we propose the following disturbance estimation approach based on the body- and world-frame integrators.
    The disturbance estimator within our system is part of the controllers, rather than the \ac{UAV} state estimator, which opens the possibility to use any \ac{UAV} state estimator that does not support disturbance estimation or even to fly without a state estimator completely, e.g., using a precise motion capture system.
    Although our disturbance estimator is arguably simple, it does not require the \ac{UAV} model.
The estimator does not need to know the \ac{UAV} mass; even more, it estimates the mass disturbance.
It can be executed regardless of the type of used \ac{UAV} estimator.
  }

  We continually estimate the world \added{force} disturbance $\mathbf{d}_w$ \added{$[\si{\newton}]$} and the body \added{force} disturbance $\mathbf{d}_b$ \added{$[\si{\newton}]$} simultaneously during the flight as
  \begin{equation}
    \begin{aligned}[]
      \mathbf{d}_w &= \sum_{n=0}^N\mathbf{k}_{iw}\circ \mathbf{\mathbf{e}}_{p[n]}\,\Delta t_{[n]},\\
      \mathbf{d}_b &= \mathbf{H}_{[N]}\sum_{n=0}^N\mathbf{k}_{ib}\circ(\mathbf{H}^\intercal_{[n]}\,\mathbf{e}_{p[n]})\,\Delta t_{[n]},
    \end{aligned}
  \end{equation}
  where
  \begin{equation}
    \mathbf{H}_{[n]} = \left[\begin{smallmatrix}
      \mathrm{cos}\ \eta_{[n]} & \mathrm{-sin}\ \eta_{[n]} & 0 \\
      \mathrm{sin}\ \eta_{[n]} & \mathrm{cos}\ \eta_{[n]} & 0 \\
      0 & 0 & 1
    \end{smallmatrix}\right]
  \end{equation}
  is the heading rotation matrix at sample $n$, $\mathbf{k}_{iw}$ is the world integral gain, and $\mathbf{k}_{ib}$ is the body integral gain.
  Until the \ac{UAV} changes its heading, all estimated disturbances are equally split in both $\mathbf{d}_w$ and $\mathbf{d}_b$.
  The physical interpretation of the x-axis and y-axis components is the force \added{$[\si{\newton}]$} after we compensate for them by the feedback in the desired force $\mathbf{f}_d$, as described in \refsec{sec:se3_state_feedback} and \refsec{sec:mpc_controller}.

  Another undesired effect is the apparent change in the mass of the \ac{UAV} that can be deduced from the applied thrust.
  This can indeed be a change in the mass of the \ac{UAV}, e.g., due to deploying the payload or gathering objects, or it can be an apparent change caused by a discharge of the battery, contact of a horizontal surface, and real-time changes in the efficiency of the propulsion system.
  Either way, we estimated the apparent mass change by using the z-axis disturbance as a part of a total estimated mass of the \ac{UAV}
  \begin{equation}
    m_e = m + \left(\mathbf{d}_w + \mathbf{H}\mathbf{d}_b\right)^\intercal\mathbf{\hat{e}}_3,
  \end{equation}
  where $m$ stands for the nominal mass obtained by weighting the \ac{UAV}.
  The physical interpretation of \updated{$\left(\mathbf{d}_w + \mathbf{H}\mathbf{d}_b\right)^\intercal\mathbf{\hat{e}}_3$} is the mass difference \added{$[\si{\kilo\gram}]$} from the nominal take-off mass, thanks, again, to the total estimated mass $m_e$ being used in the \added{control} feedback loop.



  \subsection{\updated{\emph{SE(3)}} state feedback}
  \label{sec:se3_state_feedback}

  The first of our controller variants is the \emph{agile} controller option.
  It is based upon the \emph{SE(3) geometric tracking feedback} \cite{lee2010geometric} with the addition of disturbance compensation.
  To supplement the force tracking from section~\ref{sec:force_tracking}, we define the desired force as
  \begin{equation}
    \begin{split}
      \mathbf{f}_d = &\overbrace{-m_e\mathbf{k}_p\circ \mathbf{e}_p}^{\begin{tabular}{c}
        \tiny position\\
        \tiny feedback
      \end{tabular}} + \overbrace{-m_e\mathbf{k}_v\circ \mathbf{e}_v}^{\begin{tabular}{c}
        \tiny velocity\\
        \tiny feedback
      \end{tabular}} + \overbrace{m_e\ddot{\mathbf{r}}_d}^{\begin{tabular}{c}
        \tiny reference\\
        \tiny feedforward
      \end{tabular}} + \\
      & \underbrace{m_eg\mathbf{\hat{e}}_3}_{\begin{tabular}{c}
        \tiny gravity\\
        \tiny compensation
      \end{tabular}} + \underbrace{-\mathbf{d}_w\circ \left[\begin{smallmatrix}
        1\\
        1\\
        0
      \end{smallmatrix}\right]}_{\begin{tabular}{c}
        \tiny world disturbance\\
        \tiny compensation
      \end{tabular}} + \underbrace{-\mathbf{d}_b\circ \left[\begin{smallmatrix}
        1\\
        1\\
        0
      \end{smallmatrix}\right]}_{\begin{tabular}{c}
        \tiny body disturbance\\
        \tiny compensation
      \end{tabular}},
    \end{split}
  \end{equation}
  where $\mathbf{d}_w$, $\mathbf{d}_b$ $\left[\si{\meter\per\second\squared}\right]$ are the world and body disturbance force terms, $m_e$ $\left[\si{\kilogram}\right]$ is the estimated \ac{UAV} mass, $\ddot{\mathbf{r}}_d$ $\left[\mathrm{\si{\meter\per\second\squared}}\right]$ is the desired acceleration, $g$ $\left[\si{\meter\per\second\squared}\right]$ is the magnitude of the gravitational acceleration, $\mathbf{k}_p$ are the position gains, $\mathbf{k}_v$ are the velocity gains, and $\mathbf{e}_v = \dot{\mathbf{r}} - \dot{\mathbf{r}}_d$ is the velocity control error. The z-axis component of the disturbances is eliminated by the element-wise product, as it is already compensated for in the form of the estimated mass $m_e$.



  \subsection{Model Predictive Control Force Feedback}
  \label{sec:mpc_controller}

  This controller uses a linear \ac{MPC} approach to generate a desired acceleration $\mathbf{c}_d \in \real^3$.
  The acceleration is used while calculating the desired force, similarly to the previous case:
  \begin{equation}
    \begin{split}
      \mathbf{f}_d = &\overbrace{m_e\ddot{\mathbf{r}}_d}^{\begin{tabular}{c}
        \tiny reference \\
        \tiny feedforward
      \end{tabular}} + \overbrace{m_e\mathbf{c}_d}^{\begin{tabular}{c}
        \tiny MPC \\
        \tiny feedforward
      \end{tabular}} + \overbrace{m_eg\mathbf{\hat{e}}_3}^{\begin{tabular}{c}
        \tiny gravity \\
        \tiny compensation
      \end{tabular}} + \\
      &\underbrace{-\mathbf{d}_w\circ \left[\begin{smallmatrix}
        1\\
        1\\
        0
      \end{smallmatrix}\right]}_{\begin{tabular}{c}
        \tiny world disturbance\\
        \tiny compensation
      \end{tabular}} + \underbrace{-\mathbf{d}_b\circ \left[\begin{smallmatrix}
        1\\
        1\\
        0
      \end{smallmatrix}\right]}_{\begin{tabular}{c}
        \tiny body disturbance\\
        \tiny compensation
      \end{tabular}}.
    \end{split}
  \end{equation}
  Linear \ac{MPC} is a robust feedback method for a system with a known model.
  In this case, the \ac{MPC} controller is formulated such that the control input of its model is the acceleration of the point-mass translation dynamics.
  Thus, the control input is directly used as $\mathbf{c}_d$.
  Moreover, the \ac{MPC} approach naturally solves the control problem optimally subject to given state and input constraints.
  This ensures the feasibility and the smoothness of the acceleration command, bound to satisfy maximum velocity, acceleration, and jerk.


  \subsubsection{MPC Model}

  The \ac{MPC} controller operates with an \ac{LTI} model, similar to the model used for estimation.
  However, the heading is still controlled via the \updated{\emph{SO(3)}} feedback, so there is no need to include it here.
  For \ac{MPC} we consider the following state vector:
  \begin{equation}
    \mathbf{x}_m = \left[x, \dot{x}, \ddot{x}, y, \dot{y}, \ddot{y}, z, \dot{z}, \ddot{z}\right]^\intercal.
    \label{eq:state_vector_mpc}
  \end{equation}
  The model matrices are defined as
  \begin{align*}
    &\begin{smallmatrix}
      \mathbf{A}_m\left(\Delta t\right)
    \end{smallmatrix} = \left[\begin{smallmatrix}
      \mathbf{A}_r\left(\Delta t, 0\right) & \mathbf{0}                              & \mathbf{0}                              \\
      \mathbf{0}                              & \mathbf{A}_r\left(\Delta t, 0\right) & \mathbf{0}                              \\
      \mathbf{0}                              & \mathbf{0}                              & \mathbf{A}_r\left(\Delta t, 0\right) \\
    \end{smallmatrix}\right],\numberthis
    \label{eq:model_A_mpc}
  \end{align*}
  \begin{align*}
    &\begin{smallmatrix}
      \mathbf{B}_m\left(\Delta t\right)
    \end{smallmatrix} = \left[\begin{smallmatrix}
      \mathbf{B}_r\left(\Delta t, 1\right) & \mathbf{0}                           & \mathbf{0} \\
      \mathbf{0}                           & \mathbf{B}_r\left(\Delta t, 1\right) & \mathbf{0} \\
      \mathbf{0} & \mathbf{0} & \mathbf{B}_r\left(\Delta t, 1\right)
    \end{smallmatrix}\right], \numberthis
    \label{eq:model_B_mpc}
  \end{align*}
  where $\mathbf{A}_r$ and $\mathbf{B}_r$ are the same subsystem matrices (\ref{eq:model_translation}) as in the estimator model, with $\Delta t :=$ \SI{0.05}{\second}.
  However, this time we use the free parameters of the model to apply the system input directly without delay to the acceleration state ($p_1 = p_2 = 0$).



  \subsubsection{MPC controller}

  An \ac{MPC} control error is defined along a prediction horizon of length $n$ as
  \begin{equation}
    \mathbf{e}_{[i]}~=~\mathbf{x}_{m[i]} - \mathbf{x}_{md[i]}, \forall i \in \{1, \hdots, n\},
  \end{equation}
  where $\mathbf{x}_{m[i]}$ is a state vector and $\mathbf{x}_{md[i]}$ is a reference at sample $i$ of the prediction.
  The reference state takes the form of
  \begin{equation}
    \mathbf{x}_{md[i]} = \left[x_d, 0, 0, y_d, 0, 0, z_d, 0, 0\right]^{\intercal}, \forall i \in \{1, \hdots n\}.
  \end{equation}
  The initial condition $\mathbf{x}_{m[0]}$ is commonly set to values of the current state estimate.
  However, to make the system more stable even when estimated states violate dynamic constraints, position derivatives can be substituted with states of reference vector $\bm{\chi}$ from a feedforward tracker:
  \begin{equation}
    \mathbf{x}_{m[0]} = \begin{cases}
      \small \left[x, \dot{x}, \ddot{x}, y, \dot{y}, \ddot{y}, z, \dot{z}, \ddot{z}\right]^\intercal, \text{\scriptsize if constraints satisfied,}\\
      \small \left[x, \dot{x}_d, \ddot{x}_d, y, \dot{y}_d, \ddot{y}_d, z, \dot{z}_d, \ddot{z}_d\right]^\intercal, \text{\scriptsize if violated}.
    \end{cases}
  \end{equation}
  The \ac{MPC} is formulated as a quadratic programming problem
  \begin{align*}
    & \min_{\mathbf{u}_{[1:n]}} \label{eq:mpc_cost}\numberthis
    & \frac{1}{2}\sum_{i=1}^{n-1}\left(\mathbf{e}^\intercal_{[i]}\mathbf{Q}\mathbf{e}_{[i]}\right) + \mathbf{e}^\intercal_{[n]}\mathbf{S}\mathbf{e}_{[n]}
  \end{align*}\begin{align*}
    \text{s.t.}~ \mathbf{x}_{m[i]} &= \mathbf{A}_m\mathbf{x}_{m[i-1]} \plus \mathbf{B}_m\mathbf{u}_{[i]}, &\forall i &\in \{1, \hdots, n\}\numberthis\label{eq:mpc_model}\\
    \mathbf{x}_{m[i]} &\leq \mathbf{x}_{\mathrm{max}}, &\forall i &\in \{1, \hdots, n\}\label{eq:mpc_max}\numberthis\\
    \mathbf{x}_{m[i]} &\geq \minus\mathbf{x}_{\mathrm{max}}, &\forall i &\in \{1, \hdots, n\}\label{eq:mpc_min}\numberthis \\
    \mathbf{u}_{[i]} \minus \mathbf{u}_{[i\minus1]} &\leq \dot{\mathbf{u}}_{\mathrm{max}} \Delta t, &\forall i &\in \{2, \hdots, n\}\label{eq:mpc_slew_min}\numberthis\\
    \mathbf{u}_{[i]} \minus \mathbf{u}_{[i\minus1]} &\geq \minus\dot{\mathbf{u}}_{\mathrm{max}} \Delta t, &\forall i &\in \{2, \hdots, n\}\label{eq:mpc_slew_max}\numberthis
  \end{align*}
  where the minimized quadratic cost is the sum of the squares of the control errors over the prediction horizon of length $n \in \mathbb{Z}^{\\+}$.
  $\mathbf{Q} \succcurlyeq 0$ is the state error penalization matrix and $\mathbf{S} \succcurlyeq 0$ is the last state error penalization matrix.
  Constraint (\ref{eq:mpc_model}) forces the states to follow model (\ref{eq:model_A_mpc})--(\ref{eq:model_B_mpc}), while (\ref{eq:mpc_max})--(\ref{eq:mpc_min}) bound the states of the dynamical system, and \eqref{eq:mpc_slew_min}--\eqref{eq:mpc_slew_max} limit the input slew rate, i.e., the system jerk.
  Note that we do not penalize the input action within the cost function.
  No penalty is necessary, because the slew rate directly limits the jerk.
  With the \ac{MPC} problem solved in every iteration of the control loop (at \SI{100}{\hertz}), the acceleration reference $\mathbf{c}_d$ is extracted directly from $\mathbf{u}_{[1]}$, i.e., the first control input of the \ac{MPC}.

  The particular values of $\mathbf{Q}$ and $\mathbf{S}$ were found empirically as
  \begin{align*}
    \mathbf{Q} &= \mathrm{diag}(500, 100, 100, 500, 100, 100, 100, 10, 10),\\
    \mathbf{S} &= \mathrm{diag}(1000, 300, 300, 1000, 300, 300, 100, 10, 10)\label{eq:penalizations}\numberthis.
  \end{align*}
  These values were extensively tested on a variety of platforms, ranging from \SI{1.5}{\kilogram}, $\approx$ \SI{0.5}{\meter} DJI f450, to \SI{15}{\kilogram}, $\approx$ \SI{1.2}{\meter} Tarot t18.
  We rely on this controller even for solving \emph{emergency} situations when the \updated{\emph{SE(3)}} controller fails, since the \ac{MPC} is designed to be more stable with respect to sensor noise and is designed to intrinsically satisfy state constraints.
  The choice of the constraints $\mathbf{x}_{max}$ and $\mathbf{u}_{max}$ depends on the particular application scenario.
  Most of the time, we allow the controller to reach speeds up to \SI{2}{\si{\meter\per\second}} with acceleration of \SI{2}{\meter\per\second\squared} and jerk \SI{5}{\meter\per\second\cubed}.
  However, to make the flight safe, the controller also overrides state constraints of feedforward trackers, to ensure that they are at most half the value for the controller.




  \subsection{Unbiased desired acceleration}
  \label{sec:unbiased_desired_acceleration}

  The unbiased desired acceleration is created by subtracting the estimated disturbances from the desired force created by controllers, applied to the current body orientation:
  \begin{equation}
    \mathbf{a}_d = \frac{f_d\mathbf{\hat{b}}_3 - g\mathbf{\hat{e}}_3 + \mathbf{d}_w\circ \left[\begin{smallmatrix}
      1\\
      1\\
      0
    \end{smallmatrix}\right] + \mathbf{d}_b\circ \left[\begin{smallmatrix}
      1\\
      1\\
      0
    \end{smallmatrix}\right]}{m_e}.
  \end{equation}
  The acceleration $\mathbf{a}_d$ then has a zero DC component, although nonzero tilt is produced, e.g., in order to compensate for wind and for a mass disturbance.
  Both can be achieved by dividing the compensated force by the estimated mass $m_e$ in the denominator.



  \subsection{Take-off and landing}

  Take-off and landing can be executed using both of the proposed controllers without special modification, as in the case with the reference generation (see \refsec{sec:trajectory_tracking_landoff}).
  The \emph{SE(3) controller} is preferred when high control accuracy is required, but only if the localization system provides a smooth enough state estimate.
  As later shown experimentally in \refsec{sec:estimator_noise_supression}, the \emph{MPC controller} provides much better estimator noise suppression, which is desirable during take-off and landing.
  In general, the \emph{MPC controller} is the default choice within our pipeline.



  \subsection{Feedforward failsafe controller}
  \label{sec:feedforward_failsafe_controller}

  Position feedback control cannot be executed when a localization system is lost in mid-flight.
  If velocity odometry is present, e.g., in the form of an optical flow system, the active state estimator can be switched, and an emergency landing can be executed.
  However, the system cannot continue with flight without any position and velocity state estimate.
  In such a situation, we employ a feedforward failsafe landing, which relies on the attitude controller within the installed embedded flight controller (Pixhawk).
  The \emph{failsafe controller} outputs the desired orientation to keep the UAV leveled and to maintain the desired thrust to cause moderate uncontrolled descent.
  The initial thrust is calculated using the hover-thrust curve \eqref{eq:hover_thrust_curve} with the last known estimated mass $m_e$.
  Then the thrust is decreased by a fixed rate to cause the UAV to descend.
  This procedure stops the UAV from accelerating in any direction.
  When such an emergency occurs during aggressive maneuvers, it is up to a safety pilot to recognize that there is a problem and to regain control using a remote controller.
  However, if this type of situation occurs during a slow indoor flight at low altitude, the UAV typically safely reaches the ground before any damage can occur.




\section{\added{Trajectory generation}}
\label{sec:trajectory_generation}

\added{
Generating a reference trajectory from a set of waypoints in real time is an essential step for some \ac{UAV} applications.
A commonly-adopted approach relies on optimizing a parametrized polynomial function that is later sampled \cite{mellinger2011minimum} to obtain a trajectory $\left\{\left(\mathbf{r}_d, \eta_d\right)_1, \left(\mathbf{r}_d, \eta_d\right)_2, \hdots ,\left(\mathbf{r}_d, \eta_d\right)_k\right\}$.
The approach in \cite{mellinger2011minimum} was later extended in \cite{richter2016polynomial, burri2015robust} by solving a nonlinear optimization problem to solve both for the geometry of the path and time sampling within one problem.
We extend the approach in \cite{richter2016polynomial} even further.
Although the approach generates trajectories that do not violate state constraints, they also do not minimize the total flight time, sometimes resulting in prolonged flight.
We updated the constraint satisfaction mechanism in \cite{richter2016polynomial} by rescaling only the generated trajectory segments that directly violate the constraints instead of rescaling the whole trajectory.
We also provide lower-bound initial segment time estimates for each segment.
Again, this modification improves the overall flight time of the path since the constraint satisfaction mechanism only prolongs the segment times if necessary but does not shorten them if possible.
Furthermore, we provide an iterative sub-sectioning mechanism that automatically satisfies the given maximum distance of the generated trajectory from the provided waypoint path.
Our other improvements to the method and software provided by \cite{richter2016polynomial} can be found at our GitHub page\footnote{\href{https://github.com/ctu-mrs/mrs_uav_trajectory_generation}{github.com/ctu-mrs/mrs\_uav\_trajectory\_generation}} with the rest of our software.
}



  \section{Implementation}
  \label{sec:implementation}

\begin{figure}
    \centering
    \resizebox{1.0\columnwidth}{!}{
      \input{fig/tikz/implementation_diagram.tex}
    }
    \caption{
      An illustration of the implementation diagram of the proposed \ac{UAV} system.
      Onboard sensors and actuator modules are depicted as grey blocks.
      The sensor combination varies depending on the particular \ac{UAV} task.
      White blocks represent \ac{ROS} components responsible for managing sensors or for interacting with the actuators (Mavros).
      Green blocks stand for feedback controllers (see \refsec{sec:reference_controllers_impl}) and red blocks stand for reference trackers (see \refsec{sec:reference_trackers_impl}).
      Purple blocks represent high-level components that provide the controllers and trackers with up-to-date data and maintain the synchronicity of the events.
      These include controller, tracker and estimator switching, gain and constraint scheduling, and take-off and landing.
    }
    \label{fig:implementation_diagram}
  \end{figure}
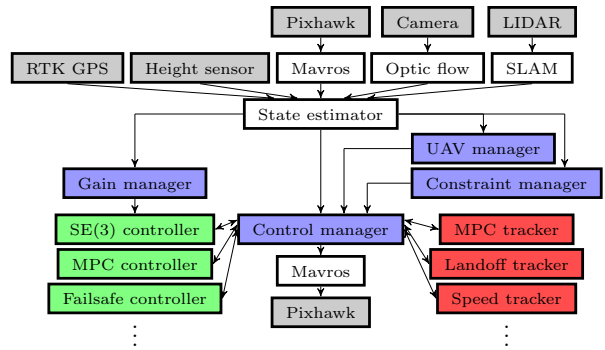

  Implementation aspects has not often been an integral part of published reports on control-oriented research.
  However, we argue that there is a need for a system-oriented manuscript that includes software and sources.
  In this section, we will discuss implementation and software design considerations of our system\footnote{\href{http://github.com/ctu-mrs/mrs_uav_system}{github.com/ctu-mrs/mrs\_uav\_system}} that have been shaped by the requirements of real-world deployment.
  Real-world deployment and verification of novel \ac{UAV} methods often require a specific platform configuration for the method being verified.
  The proposed system is designed to be extensively modular, allowing \emph{hot-swapping} of feedback controllers, trajectory trackers, state estimators, controller gains, and dynamic constraints.
  These can be changed in mid-flight so that new methods can benefit from existing and tested systems, for safely managing the initial take-off and landing, or for regaining control in the event of unwanted behavior of the tested methods.
  It is very useful to have the option to switch to a reliable backup system is when testing new real-time software.
  The proposed system is built on \ac{ROS}\footnote{Robot Operating System, \url{http://ros.org}}, and is available as open-source with all the components described within this section.
  We have striven to provide a well-documented system to allow researchers and students to flatten the initial learning curve and to focus on their particular research instead of developing yet another control pipeline.
  This approach has been shown to be effective, as demonstrated in \refsec{sec:pushing_the_frontiers} on various examples of real-world use and deployment of the platform.
  Figure~\ref{fig:implementation_diagram} is an implementation diagram of the system with its modules, which will be presented in the following sections.


  \subsection{State Estimator}

  The state estimator (see \refsec{sec:state_estimation}) was designed to provide multi-frame localization.
  Unlike a generally accepted approach to fuse all available sensory inputs into a single hypothesis, we execute a bank of estimators, each for a subset of inputs.
  If, for example, a \ac{UAV} is provided with data from a \ac{GPS} receiver (with a magnetometer), a 2D SLAM and an optic flow algorithm (velocity relative to the ground plane), we may consider executing the following estimators simultaneously: \ac{GPS}, 2D SLAM, 2D SLAM \& optic flow, optic flow.
  At any time, all hypotheses are available, and the \ac{UAV} is simultaneously localized within multiple independent world coordinate systems.
  One estimator (the coordinate system) is always selected as the \emph{primary estimator}, which is used for feedback control at the moment.
  The primary estimator can be switched in mid-flight on-demand or automatically when its hypothesis is deemed \emph{unreliable}.
  Transformations between the coordinate systems are maintained (using the \ac{ROS} Transformation library\footnote{ROS tf2, \url{http://wiki.ros.org/tf2}}), which allows a seamless definition of references in any of the existing frames of reference.
  This significantly increases the overall stability of the system.
  For example, the feedback loop can be closed using an optic flow odometry estimator when the \ac{GPS} signal is too inaccurate for feedback control.
  However, control references can still be given in the \ac{GPS} coordinate frame, without the need to change the \emph{mission \& navigation} software.
  Frequent switching of frames of references can occur, e.g., when manipulating with the environment using local sensor information.
  In the 2020 MBZIRC competition, we employed frequent switching between onboard visual servoing and global \ac{GNSS} localization.
  The UAV was attempting to grasp a brick autonomously while being localized relative to the object of interest and transitioning between pickup and place locations using \ac{GNSS} localization.



  \subsection{Control Manager}

  As demonstrated by the system architecture diagram in \reffig{fig:system_architecture}, the two most important parts of the control system are the feedforward \emph{reference tracker} and the feedback \emph{reference controllers}; for brevity \emph{trackers} and \emph{controllers}.
  Implementation-wise, various methods are used for both components, in addition to the methods presented in \refsec{sec:reference_tracker} and in \refsec{sec:reference_controller}.
  We employ multiple trackers to fulfill different roles during the flight, and being able to switch between each of these roles is a major software design factor within the system.
  Trackers and controllers are implemented as \ac{ROS} plugins (using the \emph{\ac{ROS} Pluginlib library}), which makes them follow an interface pre-defined by a central \emph{plugin manager}, called the \emph{Control Manager}.
  The controller and tracker interfaces were designed to keep tracker and controller implementation minimalistic, while the \emph{Control manager} is responsible for
  \begin{itemize}
    \item loading a defined set of trackers and controllers,
    \item gathering estimator data,
    \item synchronizing the active tracker and controller,
    \item providing all trackers and controllers with current dynamic constraints,
    \item providing a unified interface for setting desired trajectories and references,
    \item providing an \ac{API} to the common libraries used throughout the plugins,
    \item outputting the desired attitude rate and thrust command.
  \end{itemize}
  Moreover, all the incoming references and desired trajectories are transformed into the current control frame before being given to the active tracker.
  When the current control frame changes (when the active estimator is switched), all the loaded controllers and trackers are synchronously prompted to transform their internal state from the old frame to the new frame.
  When a controller or a tracker is switched, the newly activated plugin is given the last state and result of the previously-active plugin, making the transitions safe and imperceptible.
  Additionally, the \emph{Control manager} facilitates routines for
  \begin{itemize}
    \item handling excessive control errors using emergency and feedforward failsafe landing,
    \item the virtual allowed safety area with no-fly zones,
    \item the virtual reactive obstacle bumper,
    \item the control bindings to an RC controller (via Mavros\footnote{Mavros, a ROS interface to the Mavlink protocol and thus to the Pixhawk flight controller \href{http://github.com/mavlink/mavros}{github.com/mavlink/mavros}.}).
  \end{itemize}

  The system is designed with emphasis on simplifying the development and testing of new trackers/controllers and on developing new trackers and controllers for use in particular specialized applications.
  Thanks to the plugin architecture, a custom tracker and a controller can be deployed with no software changes to the proposed platform (except for customization of the \ac{ROS} \emph{launch} and \emph{config} files).
  This helps users to keep the core unchanged and therefore updated and it simplifies customization for a particular project and application, even when a single \ac{UAV} is shared by multiple users and projects.



  \subsection{Reference controllers}
  \label{sec:reference_controllers_impl}

  The feedback controllers, which are described in \refsec{sec:reference_controller}, form part of a bank of controllers loaded by the \emph{Control manager}.
  The \emph{SE(3) controller} (\refsec{sec:se3_state_feedback}) takes on the role of an agile and \emph{fast} controller that is capable of executing aggressive maneuvers with accelerations approaching \SI{10}{\meter\per\second\squared}.
  The \emph{\ac{MPC} controller} (\refsec{sec:mpc_controller}) is almost immune against estimation noise and disturbances, and also against reference infeasibilities.
  Furthermore, we utilize a \emph{Failsafe controller}, which provides feed-forward action in situations when feedback is not computable.



  \subsection{Reference trackers}
  \label{sec:reference_trackers_impl}

  The trackers are the reference generators for the controllers.
  Although we use the \emph{\ac{MPC} tracker} \cite{baca2018model} most of the time, there are scenarios where different approaches are required.
  We intentionally separated landing and take-off reference generation to another tracker, called the \emph{Landoff tracker}.
  Landing and take-off do not usually require fast maneuvers, agility, or tracking of complex trajectories.
  In contrast, admittance tracking is used to mitigate excessive control errors due to the \ac{UAV} being trapped on the ground by an unwanted mechanical attachment during take-off.
  In addition, research on \ac{UAV} swarming \cite{saska2020formation, saska2016formations} often requires more direct access to the desired states of the \ac{UAV}.
  For those situations, we provide the \emph{Speed tracker}, which allows direct control of the desired speed and/or acceleration of the \ac{UAV}, while maintaining the desired height and heading.
  In the \emph{Speed tracker}, we only constrain the first derivative of given references by a low-pass filter, which gives users more hands-on control over the hardware while still maintaining safety.



  \subsection{Gain \& Constraint Management}

  Dynamic constraints are supplied and managed globally for all trackers and controllers by the \emph{Constraint manager}.
  Pre-defined groups of constraints are loaded during each software startup, allowing users to switch between them in mid-flight.
  The following dynamic constraints are considered within one group: speed, acceleration, jerk, and snap for horizontal translation, and for vertical ascending and for vertical descending translation.
  For rotations, we consider heading speed, acceleration, jerk and snap, and the intrinsic roll, pitch, and yaw rates.
  The group can be designated with a name (e.g., \emph{slow}, \emph{medium}, \emph{fast}) and can be assigned to a matrix of allowed constraints for each type of estimator.
  A fallback option (a default constraint group) is also defined for each state estimator type.
  When the estimator type is switched during a flight, the fallback constraint group is switched automatically, if the group is missing within the allowed constraints.
  The \emph{Mission \& navigation} software can switch the constraint groups on demand, but only if they are within the list of allowed constraints.
  The \emph{Constraint manager} transfers the particular constraint values to the \emph{Control manager}, which distributes the values to all loaded trackers and controllers.

  A similar mechanism is employed to manage the \emph{SE(3) controller} gains, since the gains depend on the particular application and on the type of sensor fusion that is used.
  Again, groups of gains are defined (e.g., \emph{soft}, \emph{medium}, \emph{tight}) and are assigned to the estimator types.
  This mechanism is necessary, especially when the estimators that are used vary significantly in their noise parameters and therefore require different gains to make a flight possible.



  \subsection{UAV Manager}

  The \emph{UAV manager} implements essential high-level state machines for take-off and landing.
  Both take-off and landing routines can use a specified tracker and controller.
  The selected tracker and controller are also automatically activated after take-off.
  The user or the \emph{Mission \& navigation} software may issue an instruction to land immediately, or after returning to the last take-off location, or after flying to particular coordinates.



  \subsection{Mission \& navigation software}

  In a typical scenario, the UAV control pipeline is commanded by a user directly, using a remote terminal, or by onboard mission control software.
  Typically, both scenarios include supplying the control pipeline with desired references, trajectories, switching between constraints, trackers, and controllers.
  Although this element of the system is essential in practical applications, it is highly application-specific and it is independent of the core control pipeline.
  For examples of practical applications of the proposed control pipeline, including references to relevant perception, planning, and mission control algorithms, see Sec.~\ref{sec:pushing_the_frontiers}.

  \subsection{Simulation environment}

  The simulation software is a crucial tool for robotic research.
  For this purpose, we have developed our simulation environment, which is also made publicly available\footnote{Simulation, \url{http://github.com/ctu-mrs/simulation}}.
  It makes use of the open-source Gazebo simulator and it is set up for multiple different variants of our hardware \ac{UAV} platforms (DJI f450, DJI f550, Tarot 650 sport).
  It can also easily be extended to a new hardware setup.
  All UAV hardware elements, including the Pixhawk flight controller, the actuators, and various sensors are simulated with high fidelity, so there is only a minimal difference between simulated flight and real-world flight when using the proposed UAV system.
  This ensures a smooth transition between simulation and reality, which significantly accelerates the deployment of new robotic methods and algorithms.




  \section{Experimental evaluation}
  \label{sec:experimental_evaluation}

\begin{figure*}
    \includegraphics[width=1.0\textwidth]{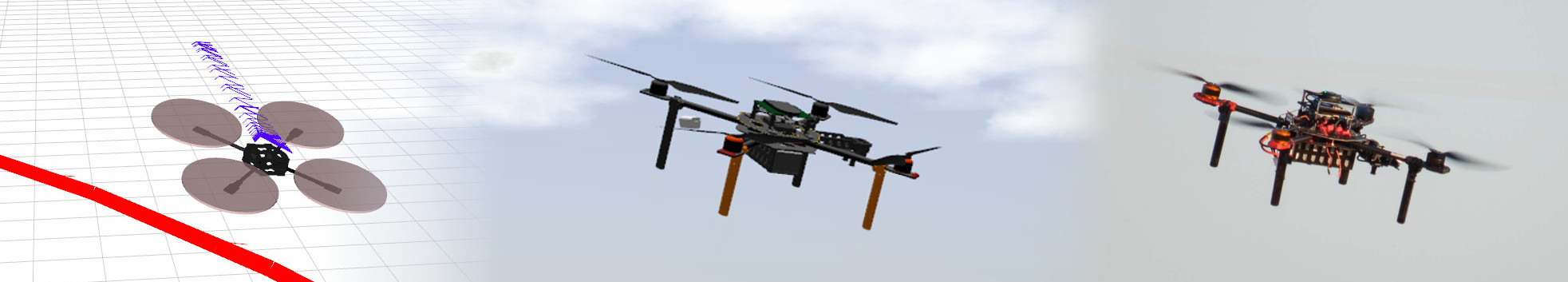}
    \caption{The Tarot 650 UAV platform is modeled with high fidelity within the simulation platform provided here.}
    \label{fig:drone_collage}
  \end{figure*}

  We performed a series of experiments to demonstrate the control and tracking performance of the proposed system in various conditions.
  All experiments were carried out in the real world as and also in the proposed simulator environment using the Tarot 650 platform (see \reffig{fig:drone_collage}).
  Figure~\ref{fig:goto_circle_collage} shows photos from the experiments, as described in the following sections.
  As shown in the comparative figures within this section, the dynamics system behaves almost identically in simulation as well as in real world.
  Importantly, the conducted maneuvers were near to the physical limits of the tested \ac{UAV}, particularly its maximum thrust.

  \subsection{Agile tracking of step position references}

\begin{figure*}
    \centering
    \includegraphics[width=1.0\textwidth]{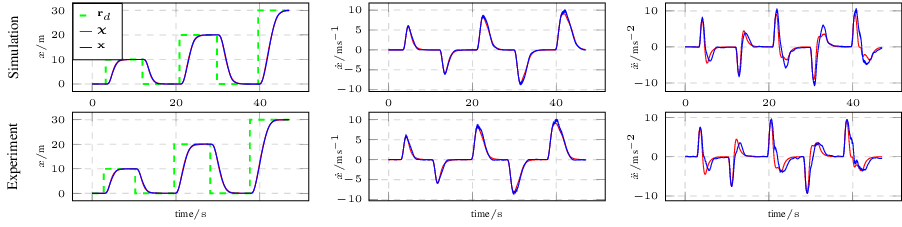}
    \caption{Comparison of the simulated response and the real response to a step position reference $\mathbf{r}_d$ in a single axis. The UAV was controlled using the \emph{SE(3) controller}. The graphs show the position, the velocity and the acceleration of the system, in terms of both the control reference $\bm{\chi}$ and the estimated state $\mathbf{x}$. The \emph{MPC tracker} operated with the following constraints: $\dot{x}_{max} = \SI{9}{\meter\per\second}$, $\ddot{x}_{max} = \SI{12}{\meter\per\second\squared}$, $\dot{\ddot{x}}_{max} = \SI{50}{\meter\per\second\cubed}$, $\ddot{\ddot{x}}_{max} = \SI{50}{\meter\per\second\tothe{4}}$.}
    \label{fig:plot_gotos_1d}
  \end{figure*}

\begin{figure*}
    \centering
    \includegraphics[width=1.0\textwidth]{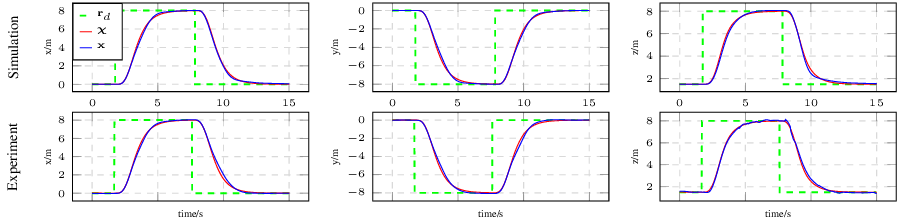}
    \caption{Comparison of the simulated response and the real response to a step position reference $\mathbf{r}_d$ in all three translation axes. The UAV was controlled using the \emph{SE(3) controller}. The graphs show the x, y and z position of the UAV, in terms of both the control reference $\bm{\chi}$ and the estimated state $\mathbf{x}$. The \emph{MPC tracker} operated with the following constraints: $\dot{x}_{max} = \SI{9}{\meter\per\second}$, $\ddot{x}_{max} = \SI{12}{\meter\per\second\squared}$, $\dot{\ddot{x}}_{max} = \SI{50}{\meter\per\second\cubed}$, $\ddot{\ddot{x}}_{max} = \SI{50}{\meter\per\second\tothe{4}}$.}
    \label{fig:plot_gotos_3d}
  \end{figure*}

  A step reference with increasing size in the desired position was supplied to the reference tracker.
  Figure~\ref{fig:plot_gotos_1d} shows the position, the velocity and the acceleration of the UAV under a series of step references in a single axis.
  Figure~\ref{fig:plot_gotos_3d} shows the position of the UAV when commanded with a 3D step reference.
  Both situations demonstrate precise and agile control near the limits of the physical capabilities of the UAV given the state constraints: $\dot{x}_{max} = \SI{9}{\meter\per\second}$, $\ddot{x}_{max} = \SI{12}{\meter\per\second\squared}$, $\dot{\ddot{x}}_{max} = \SI{50}{\meter\per\second\cubed}$, $\ddot{\ddot{x}}_{max} = \SI{50}{\meter\per\second\tothe{4}}$.


  \subsection{Circular trajectory}

\begin{figure*}
    \centering

    \includegraphics[width=1.0\textwidth]{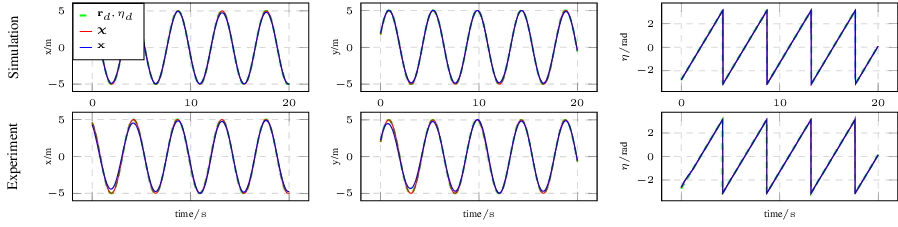}

    \caption{Comparison of the simulated tracking and the real tracking of a horizontal circular reference $\mathbf{r}_d$ with a \SI{5}{\meter} radius, \SI{7}{\meter\per\second} speed, constant height, and with the heading pointing towards the center of the circle. The UAV was controlled using the \emph{SE(3) controller}. The graphs show the x, y position of the UAV and heading $\eta$ in terms of both the control reference $\bm{\chi}$ and the estimated state $\mathbf{x}$.}
    \label{fig:plot_circles_center_heading}

  \end{figure*}

\begin{figure*}
    \centering

    \includegraphics[width=1.0\textwidth]{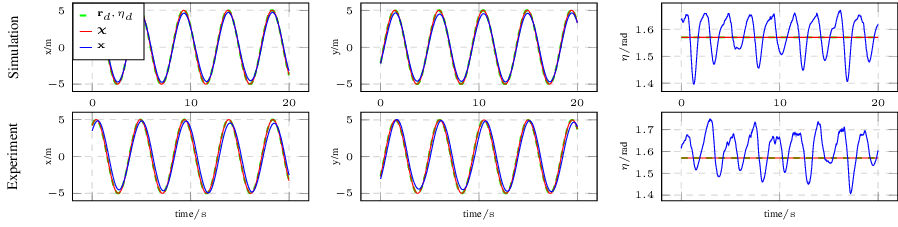}

    \caption{Comparison of the simulated tracking and the real tracking of a horizontal circular reference $\mathbf{r}_d$ with a \SI{5}{\meter} radius, \SI{7}{\meter\per\second} speed, constant height, and a constant heading. The UAV was controlled using the \emph{SE(3) controller}. The graphs show the x, y position of the UAV and heading $\eta$ in terms of both the control reference $\bm{\chi}$ and the estimated state $\mathbf{x}$.}
    \label{fig:plot_circles_constant_heading}

  \end{figure*}

\begin{figure*}
    \centering
    \subfloat {\begin{tikzpicture}
      \node[anchor=south west,inner sep=0] (a) at (0,0) {\includegraphics[width=0.32\textwidth]{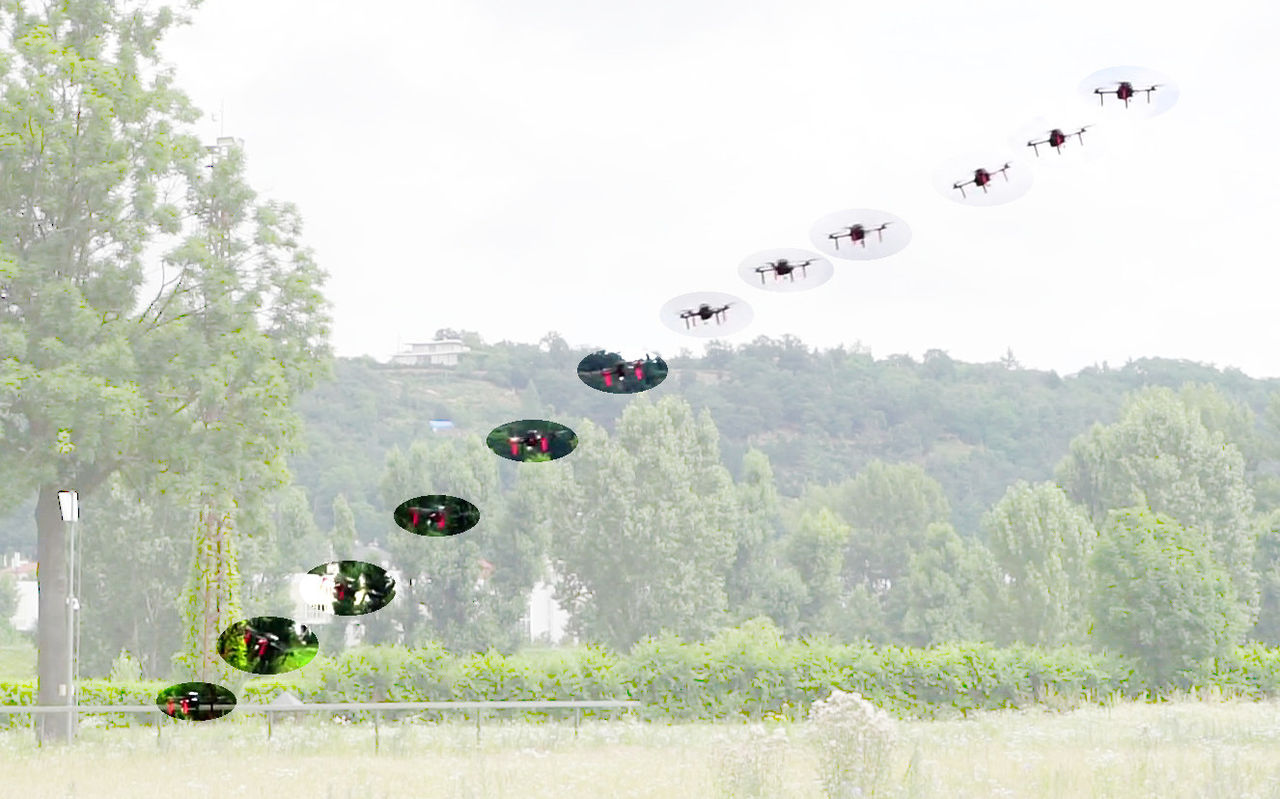}};
      \begin{scope}[x={(a.south east)},y={(a.north west)}]
        \fill[white] (0.001, 0.001) rectangle (0.09,0.13);
        \fill[draw=black, draw opacity=0.5, fill opacity=0] (0,0) rectangle (1, 1);
        \draw (0.045,0.06) node [text=black] {\small (a)};
      \end{scope}
    \end{tikzpicture}}
    \hfill%
    \subfloat {\begin{tikzpicture}
      \node[anchor=south west,inner sep=0] (a) at (0,0) {\includegraphics[width=0.32\textwidth]{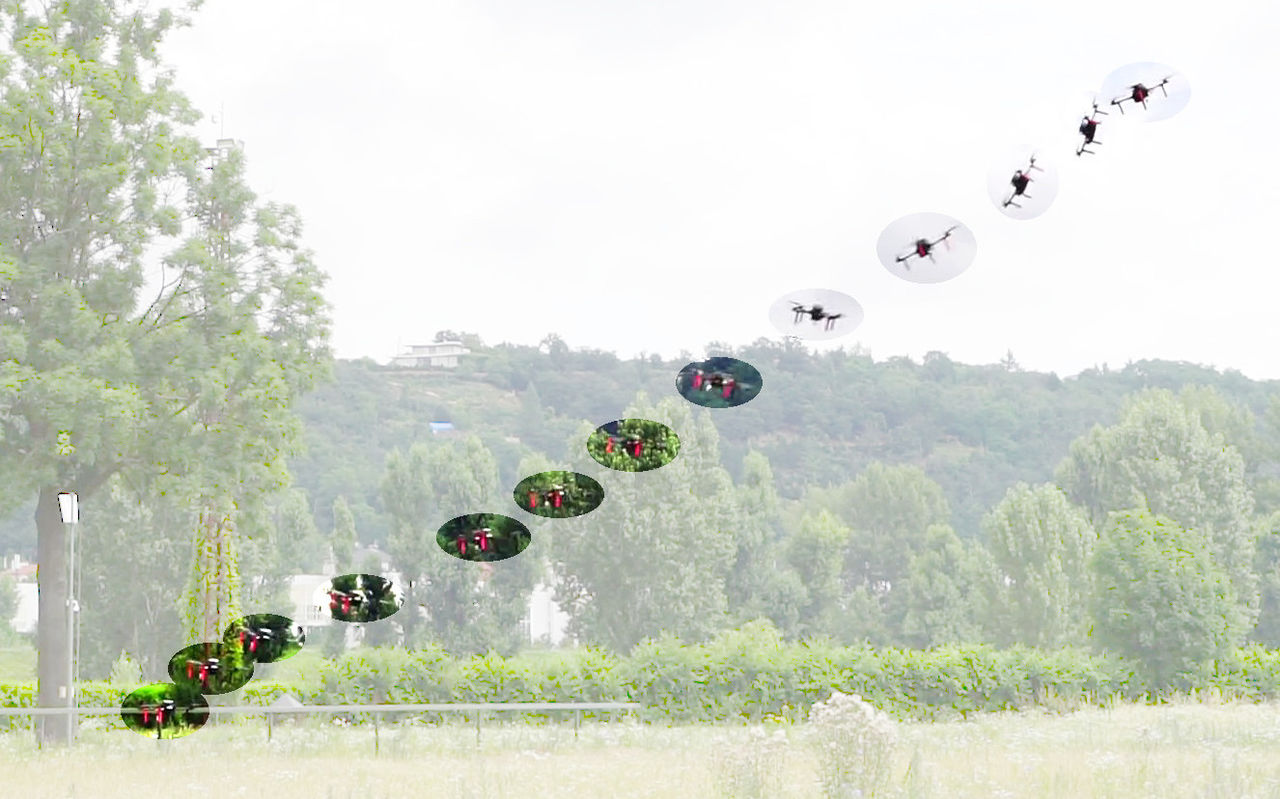}};
      \begin{scope}[x={(a.south east)},y={(a.north west)}]
        \fill[white] (0.001, 0.001) rectangle (0.09,0.13);
        \fill[draw=black, draw opacity=0.5, fill opacity=0] (0,0) rectangle (1, 1);
        \draw (0.045,0.06) node [text=black] {\small (b)};
      \end{scope}
    \end{tikzpicture}}
    \hfill%
    \subfloat {\begin{tikzpicture}
      \node[anchor=south west,inner sep=0] (a) at (0,0) {\includegraphics[width=0.32\textwidth]{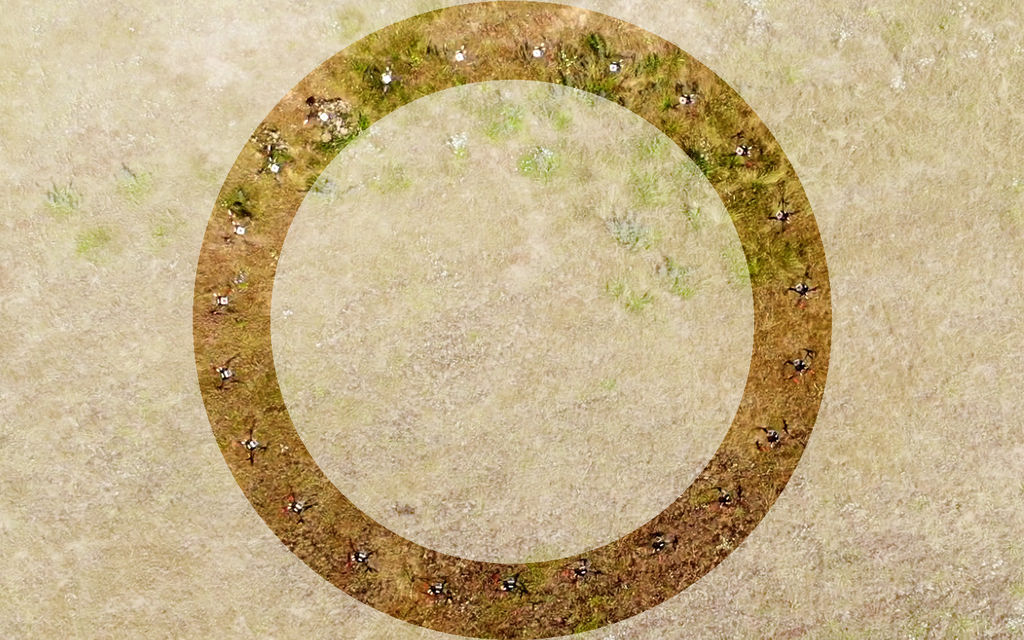}};
      \begin{scope}[x={(a.south east)},y={(a.north west)}]
        \fill[white] (0.001, 0.001) rectangle (0.09,0.13);
        \fill[draw=black, draw opacity=0.5, fill opacity=0] (0,0) rectangle (1, 1);
        \draw (0.045,0.06) node [text=black] {\small (c)};
      \end{scope}
    \end{tikzpicture}}
    \caption{Photo collage of a UAV performing aggressive testing maneuvers. Figures (a) and (b) depict the UAV performing the 3D step reference, as showcased in \reffig{fig:plot_gotos_3d}. Figure (c) shows a top-down view of the circular trajectory, showcased in \reffig{fig:plot_circles_constant_heading} and in \reffig{fig:plot_circles_center_heading}. Go to \protect\url{http://mrs.felk.cvut.cz/mrs-uav-system} for video material from the experiments.}
    \label{fig:goto_circle_collage}
  \end{figure*}

  Tracking a circular trajectory is a challenging task due to the ever-changing acceleration of the vehicle.
  Figures \ref{fig:plot_circles_center_heading} and \ref{fig:plot_circles_constant_heading} show the x, y position, and the heading $\eta$ of the UAV while tracking a horizontal trajectory with a radius of \SI{5}{\meter} and a speed of \SI{7}{\meter\per\second}.
  The UAV produced centripetal acceleration close to \SI{10}{\meter\per\second\squared} to maintain the circular motion.
  Figure~\ref{fig:plot_circles_center_heading} shows a trajectory with the heading pointing towards the center of the circle.
  This is the simplest scenario, for several reasons.
  The air drag acts on the vehicle from the same direction throughout the flight, enabling an estimate to be made using the proposed body disturbance estimator.
  In addition, this situation does not create any parasitic heading rate, and the desired heading rate is completely satisfied with just the $\omega_1$ and $\omega_2$ angular velocities.
  On the other hand, \reffig{fig:plot_circles_constant_heading} shows a circular trajectory with a constant heading in the world.
  This is a challenging trajectory to follow, since the air drag cannot be estimated using the proposed pipeline, and the motion requires continuous action using the angular velocity $\omega_3$ to produce the feedforward heading rate motion and to compensate the parasitic heading rate.
  However, despite these difficulties, the \emph{SE(3) controller} is able to track trajectories of this type with an average position error of \SI{0.5}{\meter}, and \SI{0.1}{\meter} for the first case.
  As with the step references, these circular trajectories are near the physical limits of the tested UAV.


  \subsection{Estimator noise suppression}
  \label{sec:estimator_noise_supression}

\begin{figure*}
    \centering

    \includegraphics[width=1.0\textwidth]{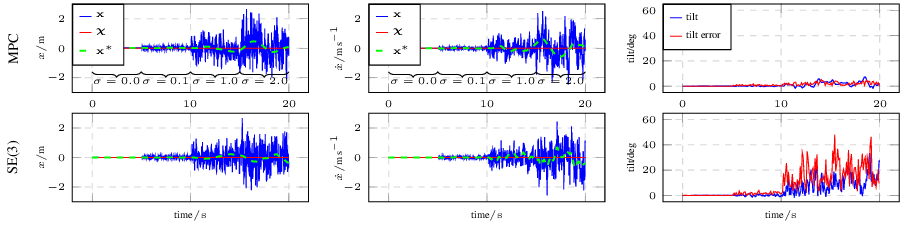}

    \caption{Comparison of simulated control performance under noisy UAV state estimate $\mathbf{x}$. A simulated ground truth is denoted by $\mathbf{x}^{\ast}$. The UAV state estimate is based upon noisy position and velocity measurements (e.g., visual odometry) with artificial noise drawn from the distribution $\mathcal{N}(\mu, \sigma^2)$, where $\mu=0.0$ and $\sigma \in \{0.0, 0.1, 1.0, 2.0\}$. The \emph{MPC controller} (first row) successfully stabilizes the UAV without producing excessive tilts $\angle (\mathbf{\hat{b}}_3, \mathbf{\hat{e}}_3)$ and tilt control errors. The \emph{SE(3) controller} (second row) handles the situation with more difficulty while producing excessive tilts and tilt control errors. The control action of the \emph{MPC controller} is thus fit for noisy localization systems. The \emph{SE(3) controller} may destabilize the UAV and disturb the onboard localization systems with excessive control actions.}
    \label{fig:plot_estimator_noise}

  \end{figure*}

  The proposed \emph{MPC controller} provides stabilization and control even with a noisy state estimate.
  It is vital to deploy this type of control scheme in scenarios where the localization system may produce noisy measurements.
  Tuning a state estimator to smooth out the noise in the estimated states is not always an option, as it can increase the transfer delay of the estimator to such an extent that the estimator can make the closed loop unstable.
  We therefore, we prefer to use a controller that is resistant to excessive noise in the estimated states.
  Figure~\ref{fig:plot_estimator_noise} shows a simulation of the stabilization properties of the \emph{MPC controller} and the \emph{SE(3) controller}, when the estimated position and velocity are increasingly noisy.
  The performance of the \emph{MPC controller} allows the flight to continue even after a significant noise is present, whereas the \emph{SE(3) controller} would possibly lead to a premature uncontrolled landing due to excessive control actions leading to a loss of onboard localization systems.


  \subsection{Estimator position jump handling}

\begin{figure*}
    \centering

    \includegraphics[width=1.0\textwidth]{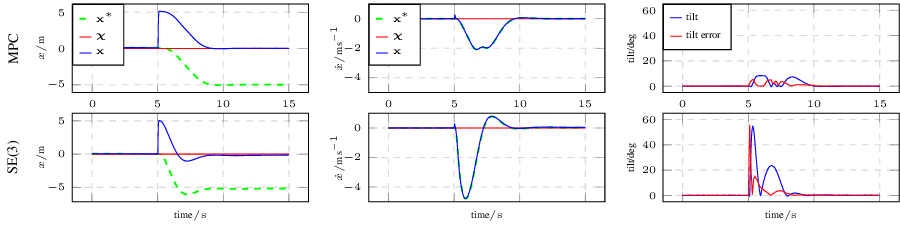}

    \caption{Simulated comparison of control reactions to a \SI{5}{\meter} jump in the estimated position of the UAV $\mathbf{x}$ (e.g., due to a malfunction of a localization system), $\mathbf{x}^{\ast}$ stands for the ground truth. The \emph{MPC controller} (first row) successfully minimizes the control error while satisfying its dynamics constraints (\SI{2}{\meter\per\second} speed, \SI{2}{\meter\per\second\squared} acceleration) and thus produces reasonably small tilts $\angle (\mathbf{\hat{b}}_3, \mathbf{\hat{e}}_3)$ and tilt control errors. The \emph{SE(3) controller} (second row) handles the situation with more difficulty while producing unbounded speed, acceleration, and therefore excessive tilts and tilt control errors, which may further disturb the onboard localization systems.}
    \label{fig:plot_odometry_jump}

  \end{figure*}

  As in the case of high estimator noise, the \emph{MPC controller} outperforms the \emph{SE(3) controller} in terms of resistance to state estimator infeasibilities.
  Jumps in the estimated positions are common problem with onboard \ac{SLAMs}.
  Similarly, jumps in the control reference may occur when developing and testing new trajectory tracking approaches.
  Figure~\ref{fig:plot_odometry_jump} shows a feedback reaction of both controllers to a \SI{5}{\meter} jump in the estimated position.
  The \emph{MPC controller} minimizes the control error smoothly while satisfying its internal state constraints (\SI{2}{\meter\per\second} speed, \SI{2}{\meter\per\second\squared} acceleration) and producing mild control actions.
  The \emph{SE(3) controller} also stabilizes the UAV.
  However, the controlled states are unbounded, leading to excessive tilts and again possibly to the loss of onboard localization systems.


  \section{Pushing the frontiers of UAV research}
  \label{sec:pushing_the_frontiers}

  The proposed \ac{UAV} system has been used extensively for evaluating of basic research outside laboratory conditions, in applied research, and during real-world verification of novel approaches within robotic challenges and competitions.
  The system has been evolving continuously over the years as we have faced the challenges of various scenarios described in this section.
  None of the previously published papers contains a complete and up-to-date description of the system, mainly due to their focus on high-level robotics tasks.
  This publication therefore focuses solely on the underlying UAV system, which has been shaped by the vast number of application scenarios that have required different onboard sensor configurations.
  One of the main contributions of this publication resulting from the diverse application requirements is the creation of a universal system.
  The following subsections will briefly discuss the major results achieved using the proposed architecture.

  \subsection{UAV mutual detection and localization}

\begin{figure}
    \centering
    \subfloat {\begin{tikzpicture}
      \node[anchor=south west,inner sep=0] (a) at (0,0) { \includegraphics[width=0.345\textwidth]{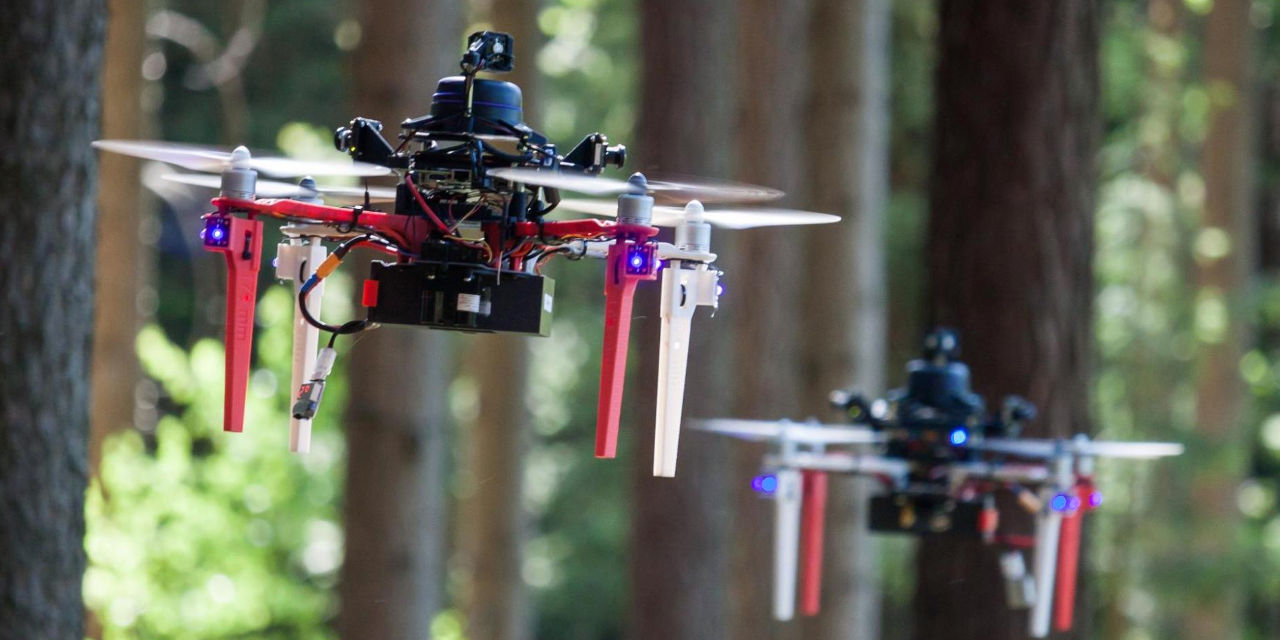}};
      \begin{scope}[x={(a.south east)},y={(a.north west)}]

        \draw[->, white, thick] (0.40, 0.20) -- (0.18, 0.59);
        \draw[->, white, thick] (0.40, 0.20) -- (0.48, 0.56);
        \draw[->, white, thick] (0.40, 0.20) -- (0.58, 0.24);
        \draw (0.40,0.15) node [text=white] {\small UV blinkers};

        \draw[->, white, thick] (0.80, 0.80) -- (0.80, 0.4);
        \draw[->, white, thick] (0.80, 0.80) -- (0.66, 0.4);
        \draw[->, white, thick] (0.80, 0.80) -- (0.50, 0.77);
        \draw (0.80,0.86) node [text=white] {\small UV cameras};

        \fill[white] (0.001, 0.001) rectangle (0.08,0.13);
        \fill[draw=black, draw opacity=0.5, fill opacity=0] (0,0) rectangle (1, 1);
        \draw (0.04,0.06) node [text=black] {\small (a)};
      \end{scope}
    \end{tikzpicture}}
    \hfill%
    \subfloat {\begin{tikzpicture}
      \node[anchor=south west,inner sep=0] (a) at (0,0) { \includegraphics[width=0.115\textwidth]{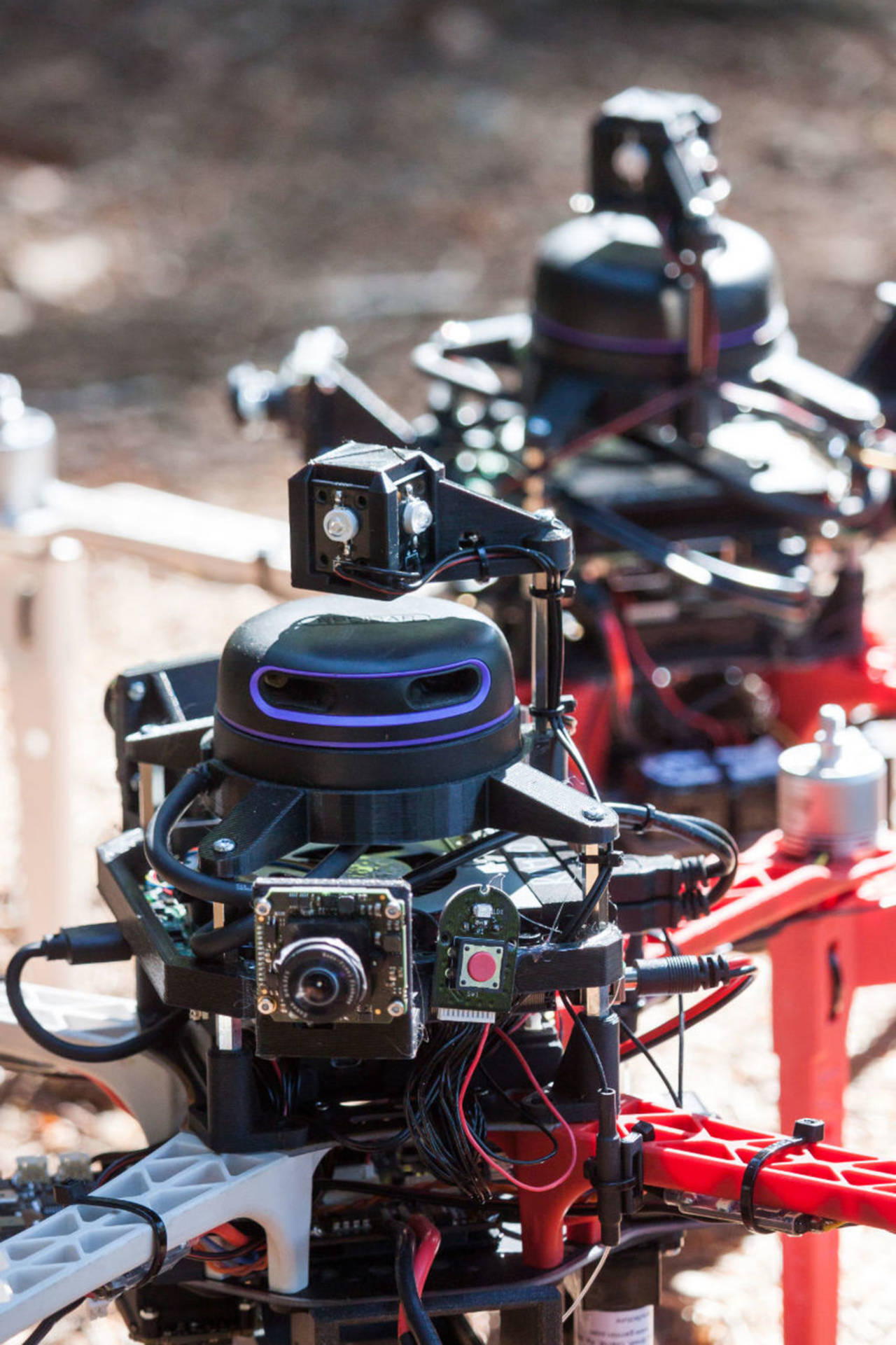}};
      \begin{scope}[x={(a.south east)},y={(a.north west)}]
        \fill[white] (0.001, 0.001) rectangle (0.20,0.13);
        \fill[draw=black, draw opacity=0.5, fill opacity=0] (0,0) rectangle (1, 1);
        \draw (0.10,0.06) node [text=black] {\small (b)};
      \end{scope}
    \end{tikzpicture}}
    \caption{Mutual localization of \acp{UAV} by the UVDAR system is provided by (a) \ac{UV} blinkers on the \ac{UAV} arms and top. The blinkers are observed by onboard cameras (b) equipped with \ac{UV} band pass filters.}
    \label{fig:uvdar}
  \end{figure}
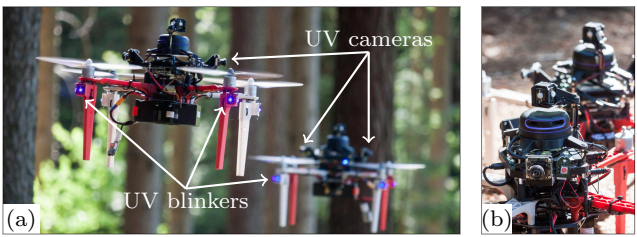

  The system played an integral role in the ongoing research on relatively-localized \ac{UAV} swarms and formations.
  Onboard marker-less \ac{UAV} detection and localization were studied in \cite{vrba2020markerless, vrba2019onboard}.
  Two approaches to \ac{UAV} detection were proposed, and were experimentally verified with the proposed system: a Convolutional Neural Network-based method, and a system for processing depth-camera images.
  Mutual localization of \acp{UAV} within swarms and formations was presented in \cite{walter2017selflocalization, walter2018fast, walter2018mutual, walter2019uvdar}.
  The system relies on modulated \ac{UV} \ac{LED} blinkers, which are detected using specialized onboard cameras (see \reffig{fig:uvdar}).
  This \ac{UVDAR} system is also available as open-source\footnote{UVDAR, \url{http://github.com/ctu-mrs/uvdar}}.

  \subsection{UAV motion planning}

  Basic research on optimal planning for data collection with \acp{UAV} was studied in \cite{penicka2019data, penicka2017dubins, penicka2017neighborhoods, penicka2017reactive, faigl2017onsolution}.
  The platform provided real-world verification and showed the feasibility of the proposed approaches.
  Coverage optimization for multi-\ac{UAV} cooperative surveillance was tackled in \cite{petrlik2019coverage, faigl2019unsupervised}.
  Complex maneuvers and cooperative load-carrying by multiple \acp{UAV} were reported on in \cite{spurny2019transport, spurny2016complex}.

  \subsection{Automatic control}

\begin{figure}
    \centering
    \subfloat {\begin{tikzpicture}
      \node[anchor=south west,inner sep=0] (a) at (0,0) { \includegraphics[width=0.235\textwidth]{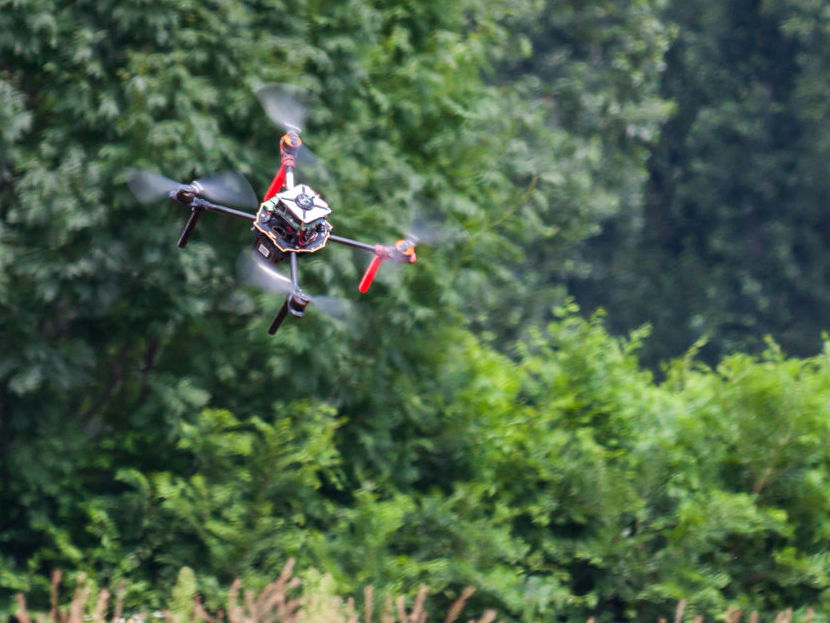}};
      \begin{scope}[x={(a.south east)},y={(a.north west)}]
        \fill[white] (0.001, 0.001) rectangle (0.12,0.13);
        \fill[draw=black, draw opacity=0.5, fill opacity=0] (0,0) rectangle (1, 1);
        \draw (0.06,0.06) node [text=black] {\small (a)};
      \end{scope}
    \end{tikzpicture}}
    \hfill%
    \subfloat {\begin{tikzpicture}
      \node[anchor=south west,inner sep=0] (a) at (0,0) { \includegraphics[width=0.235\textwidth]{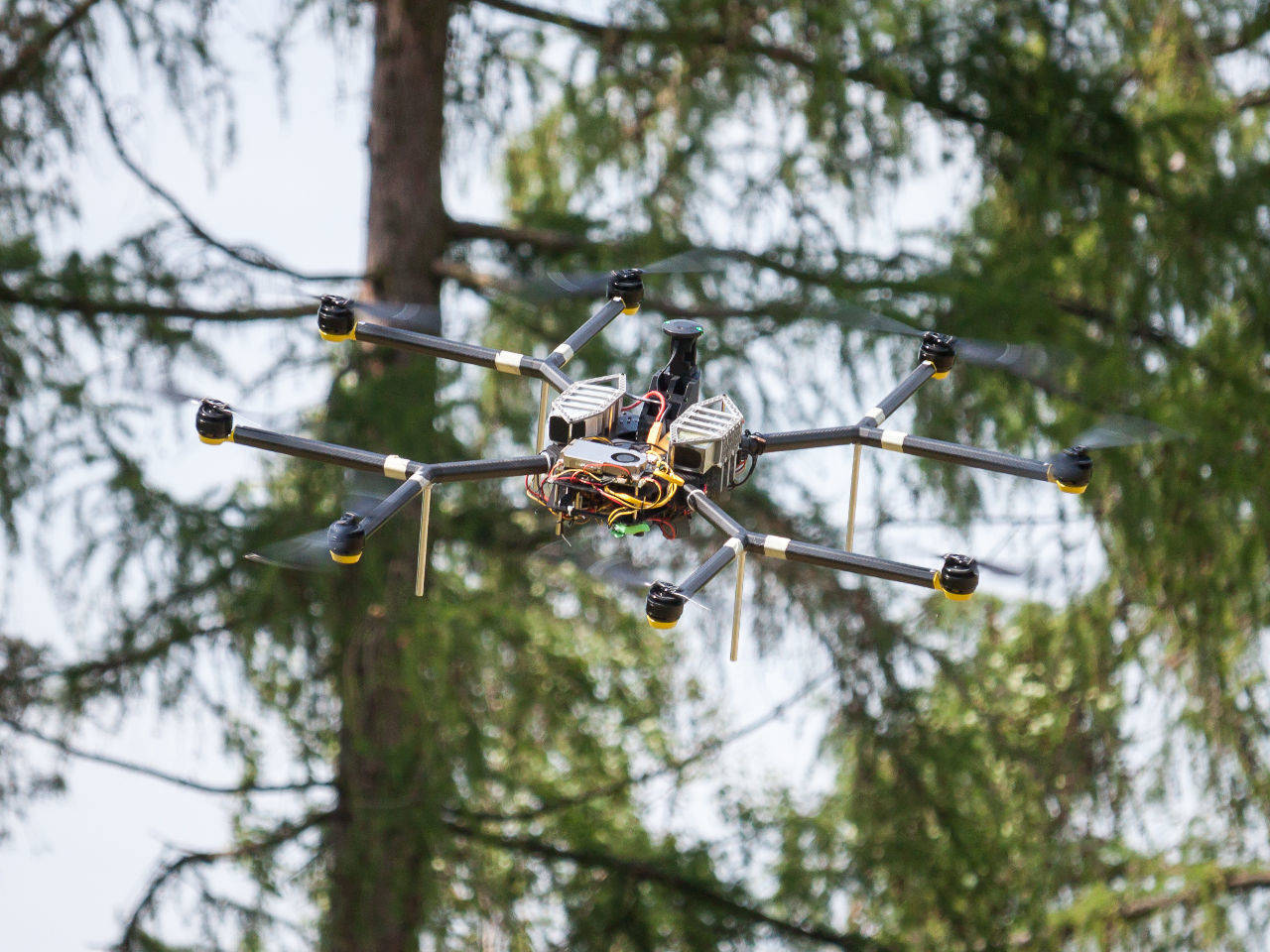}};
      \begin{scope}[x={(a.south east)},y={(a.north west)}]
        \fill[white] (0.001, 0.001) rectangle (0.12,0.13);
        \fill[draw=black, draw opacity=0.5, fill opacity=0] (0,0) rectangle (1, 1);
        \draw (0.06,0.06) node [text=black] {\small (b)};
      \end{scope}
    \end{tikzpicture}}
    \caption{Novel control approaches can be tested on a real hardware. Off-the-shelf platforms such as (a) Tarot 650, and also (b) custom-built airframes, can be equipped with the proposed system.}
    \label{fig:control}
  \end{figure}
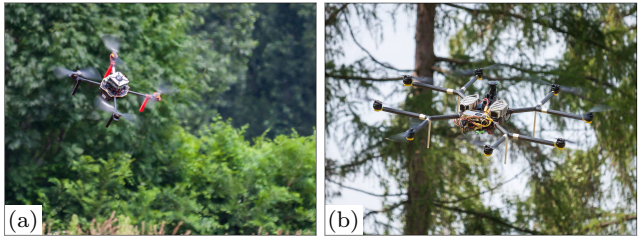

  A system for automatic gain tuning for the \emph{SE(3) controller} (see \refsec{sec:se3_state_feedback}) was published in \cite{giernacki2019realtime}.
  A novel optimal control design approach for automatic fire extinguishing is showcased in \cite{saikin2020wildfire}.
  The properties of the \updated{\emph{SE(3)}} geometric feedback proved crucial for verifying the feasibility of the almost-free-fall trajectories designed to dispatch water during extreme maneuvers (see \reffig{fig:control}).

  \subsection{Data gathering}

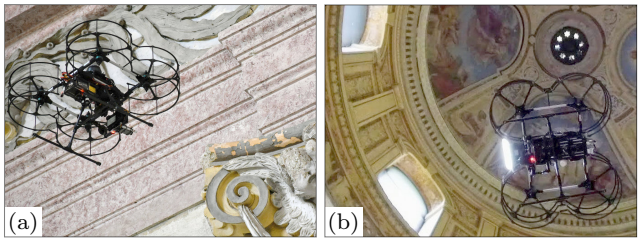
\begin{figure}
    \centering
    \subfloat {\begin{tikzpicture}
      \node[anchor=south west,inner sep=0] (a) at (0,0) { \includegraphics[width=0.235\textwidth]{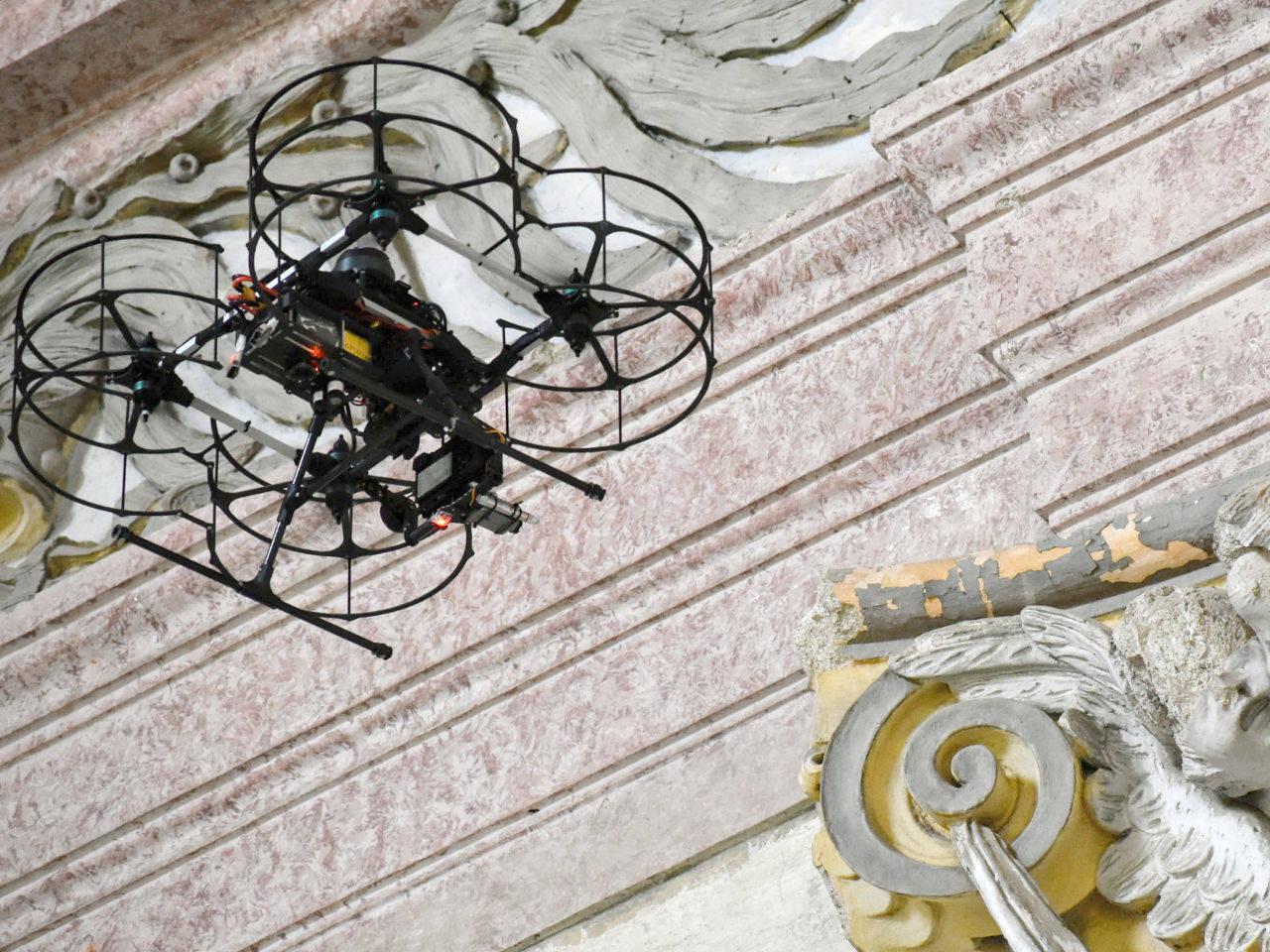}};
      \begin{scope}[x={(a.south east)},y={(a.north west)}]
        \fill[white] (0.001, 0.001) rectangle (0.12,0.13);
        \fill[draw=black, draw opacity=0.5, fill opacity=0] (0,0) rectangle (1, 1);
        \draw (0.06,0.06) node [text=black] {\small (a)};
      \end{scope}
    \end{tikzpicture}}
    \hfill%
    \subfloat {\begin{tikzpicture}
      \node[anchor=south west,inner sep=0] (a) at (0,0) { \includegraphics[width=0.235\textwidth]{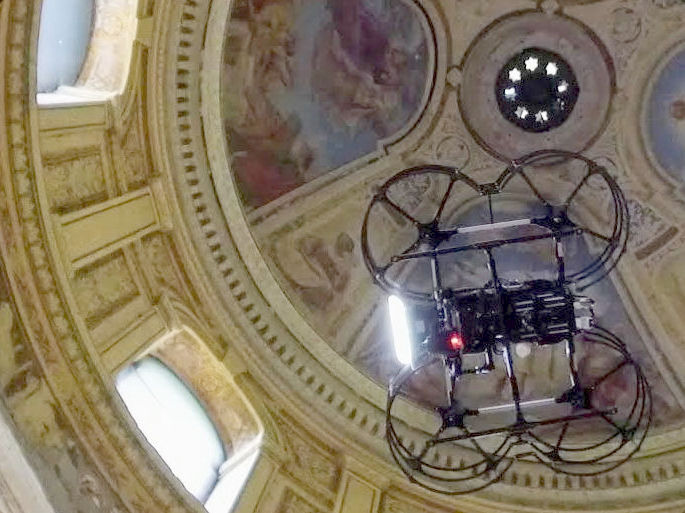}};
      \begin{scope}[x={(a.south east)},y={(a.north west)}]
        \fill[white] (0.001, 0.001) rectangle (0.12,0.13);
        \fill[draw=black, draw opacity=0.5, fill opacity=0] (0,0) rectangle (1, 1);
        \draw (0.06,0.06) node [text=black] {\small (b)};
      \end{scope}
    \end{tikzpicture}}
    \caption{An inspection of an indoor historical building is conducted (a) to monitor the state of frescoes, and (b) to assess the state of wall paintings.}
    \label{fig:dronument}
  \end{figure}

  The system is being used actively in a project working on indoor aerial inspection of historical buildings and monuments \cite{petracek2020dronument, saska2017documentation, kratky2020autonomous}.
  Within this scenario, a \ac{UAV} is equipped with a 3D \ac{LiDAR} sensor and is automatically guided through an indoor environment, where it captures detailed imagery of hard-to-reach points of interest (see \reffig{fig:dronument}).
  In another project\footnote{\url{http://mrs.felk.cvut.cz/radron}}, ionizing radiation mapping and localization is studied in \cite{baca2018rospix, baca2019timepix, stibinger2020localization}.
  Similarly, transmission radio sources were automatically localized in \cite{vrba2019realtime}.

  \subsection{UAV swarms and formations}
  \label{sec:uav_swarms_and_formations}

\begin{figure}
    \centering
    \subfloat {\begin{tikzpicture}
      \node[anchor=south west,inner sep=0] (a) at (0,0) { \includegraphics[width=0.235\textwidth]{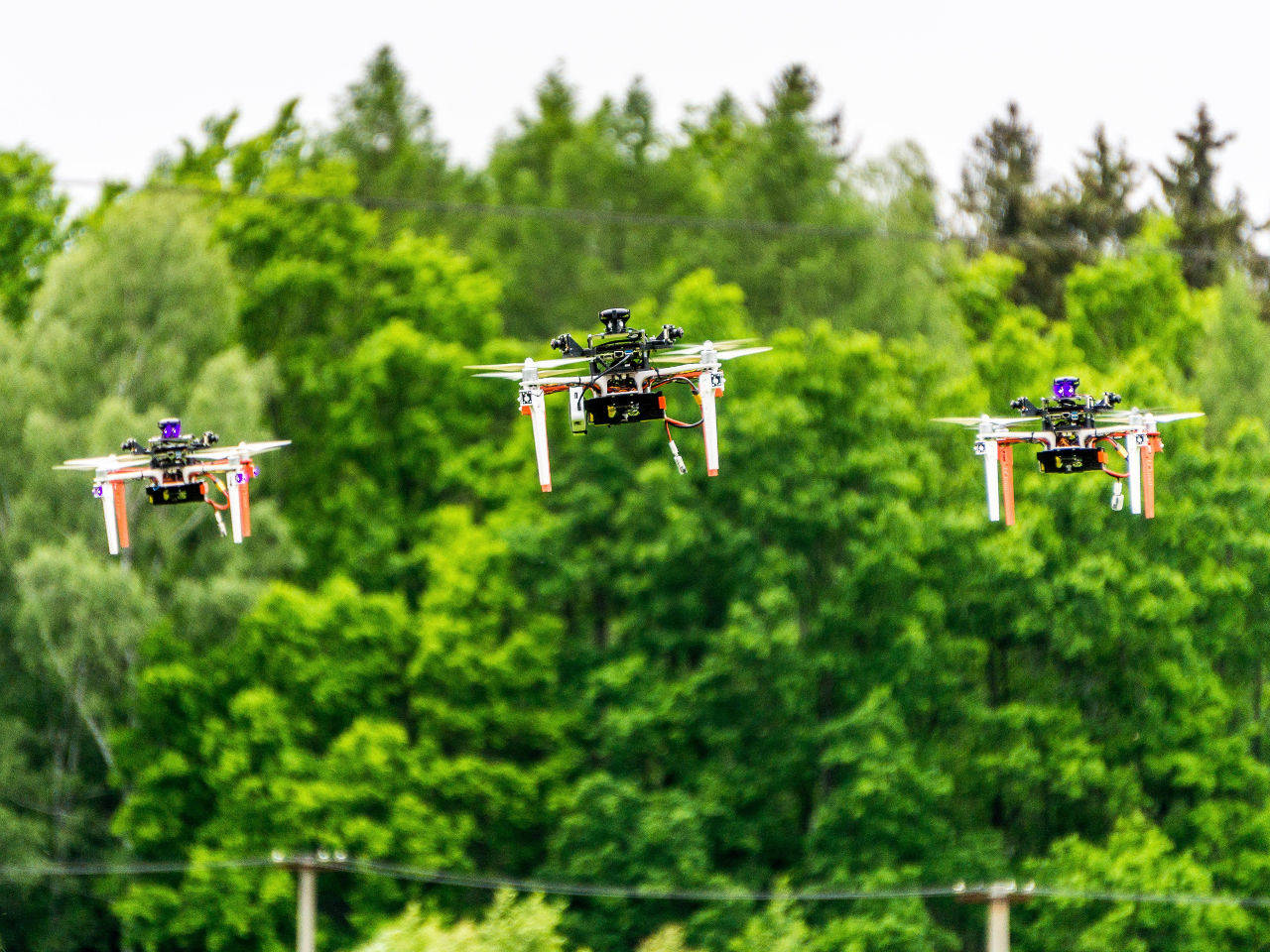}};
      \begin{scope}[x={(a.south east)},y={(a.north west)}]
        \fill[white] (0.001, 0.001) rectangle (0.12,0.13);
        \fill[draw=black, draw opacity=0.5, fill opacity=0] (0,0) rectangle (1, 1);
        \draw (0.06,0.06) node [text=black] {\small (a)};
      \end{scope}
    \end{tikzpicture}}
    \hfill
    \subfloat {\begin{tikzpicture}
      \node[anchor=south west,inner sep=0] (a) at (0,0) { \includegraphics[width=0.235\textwidth]{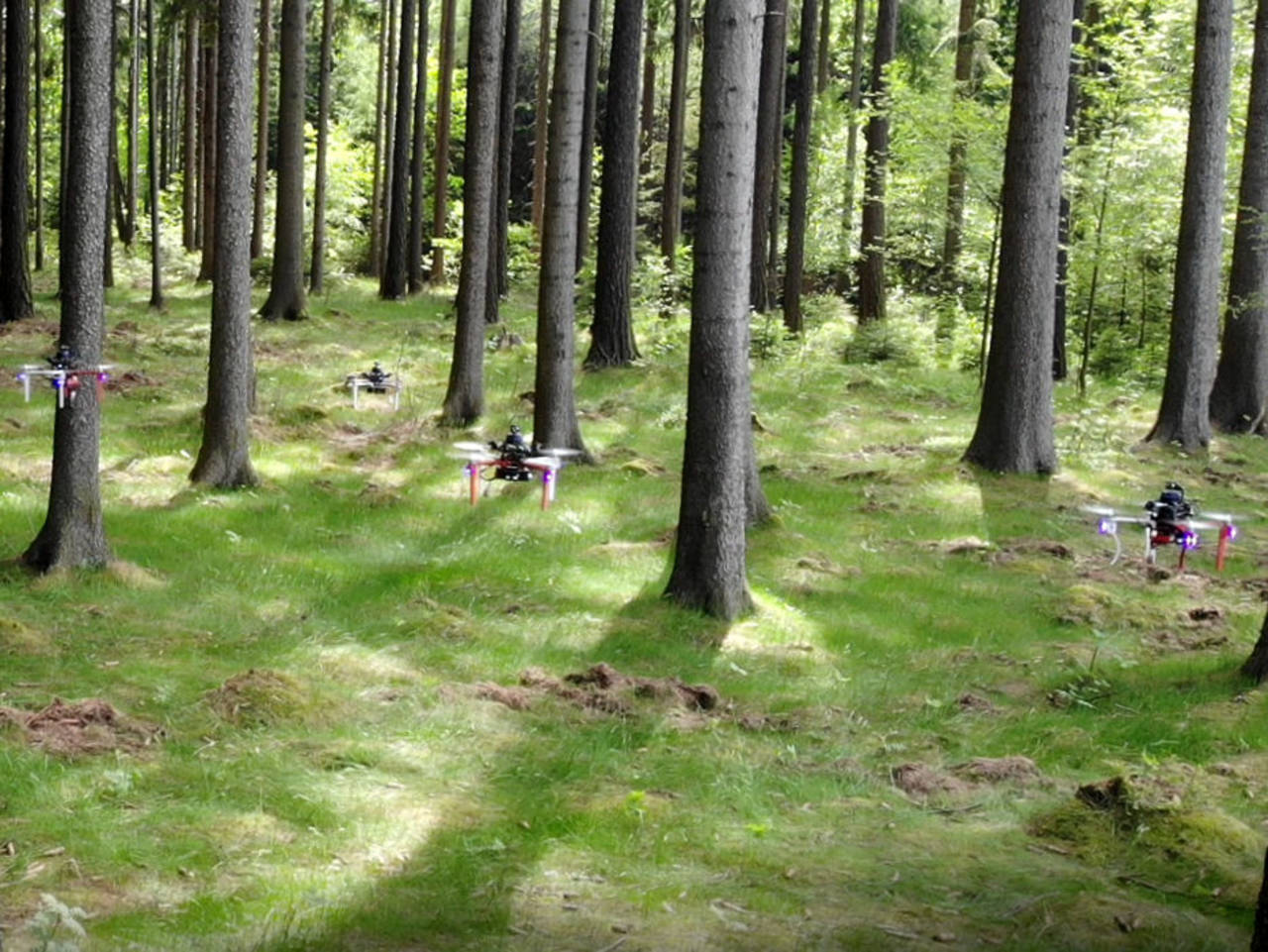}};
      \begin{scope}[x={(a.south east)},y={(a.north west)}]

        \draw[->, white, thick] (0.45, 0.20) -- (0.07, 0.57);
        \draw[->, white, thick] (0.45, 0.20) -- (0.30, 0.55);
        \draw[->, white, thick] (0.45, 0.20) -- (0.40, 0.45);
        \draw[->, white, thick] (0.45, 0.20) -- (0.85, 0.40);
        \draw (0.45,0.15) node [text=white] {\small UAVs};

        \fill[white] (0.001, 0.001) rectangle (0.12,0.13);
        \fill[draw=black, draw opacity=0.5, fill opacity=0] (0,0) rectangle (1, 1);
        \draw (0.06,0.06) node [text=black] {\small (b)};
      \end{scope}
    \end{tikzpicture}}
    \caption{Swarms of multirotor \acp{UAV} testing novel flocking algorithms while localized (a) by a \ac{GNSS} system, and (b) by onboard sensors only within a forest environment.}
    \label{fig:swarms}
  \end{figure}
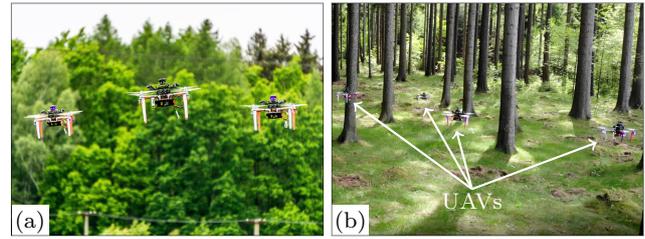

  Basic research in the area of \ac{UAV} swarming and formation flying was studied in \cite{saska2020formation, saska2016formations, saska2019large}.
  UAV swarm control is a relatively new field of research, and its applications are yet to be explored.
  One of many possibilities being explored by the authors is the use of \acp{UAV} for inspecting hard-to-access locations such as power line towers without putting personnel at risk\footnote{\url{https://aerial-core.eu}}.
  This type of application requires the swarm coordination to be flexible, and to move, while minimizing the observed object estimation error.
  Flocking capabilities are being explored within the framework of ongoing projects with real-world experiments in a field, and also within a forest environment (see \reffig{fig:swarms}).
  Interactions between \acp{UAV} are studied in order to overcome challenging situations such as GNSS-denied environment navigation.

  \subsection{MBZIRC 2017 competition}

\begin{figure}
    \centering
    \subfloat {\begin{tikzpicture}
      \node[anchor=south west,inner sep=0] (a) at (0,0) { \includegraphics[width=0.235\textwidth]{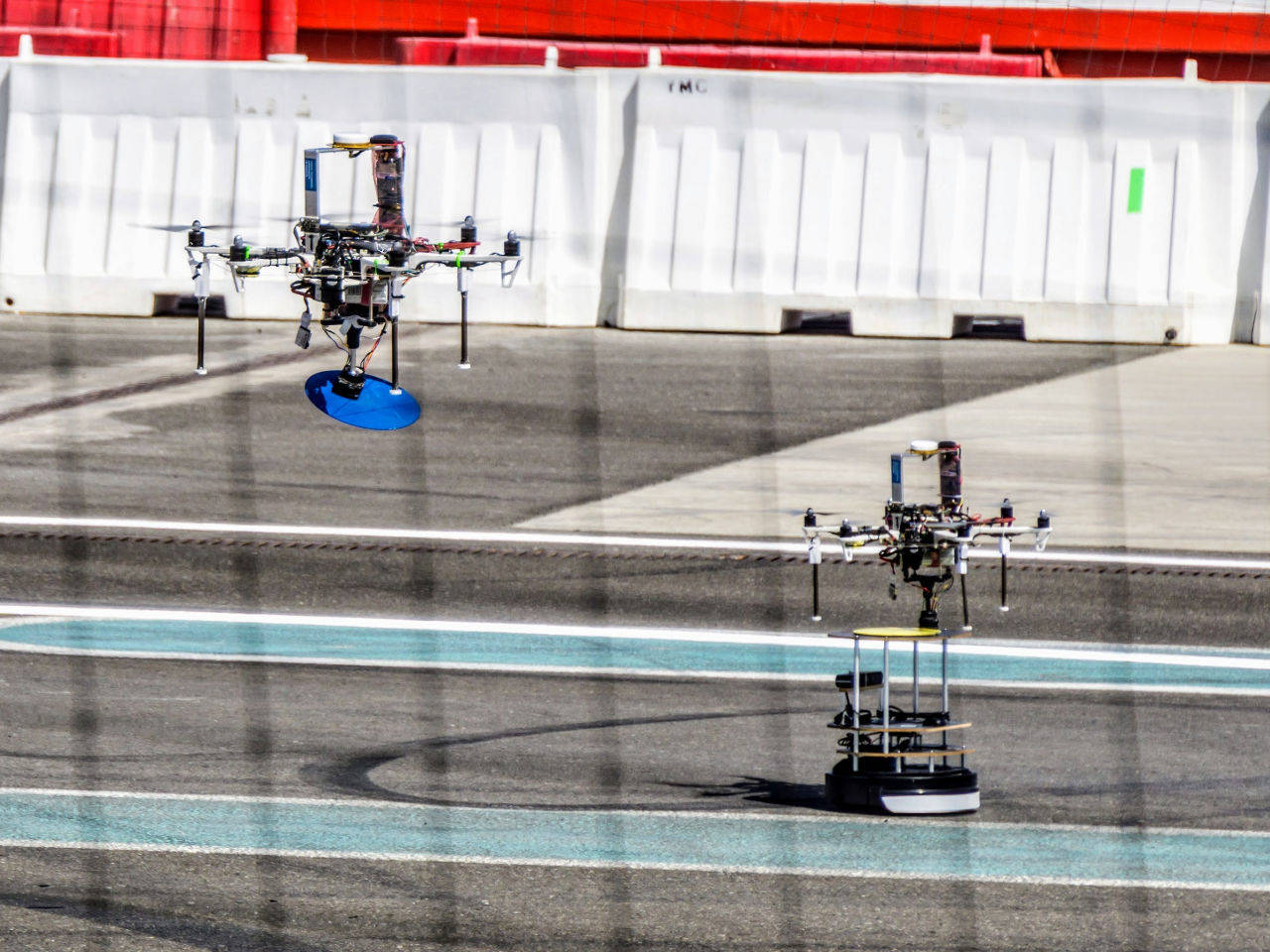}};
      \begin{scope}[x={(a.south east)},y={(a.north west)}]
        \fill[white] (0.001, 0.001) rectangle (0.12,0.13);
        \fill[draw=black, draw opacity=0.5, fill opacity=0] (0,0) rectangle (1, 1);
        \draw (0.06,0.06) node [text=black] {\small (a)};
      \end{scope}
    \end{tikzpicture}}
    \hfill%
    \subfloat {\begin{tikzpicture}
      \node[anchor=south west,inner sep=0] (a) at (0,0) { \includegraphics[width=0.235\textwidth]{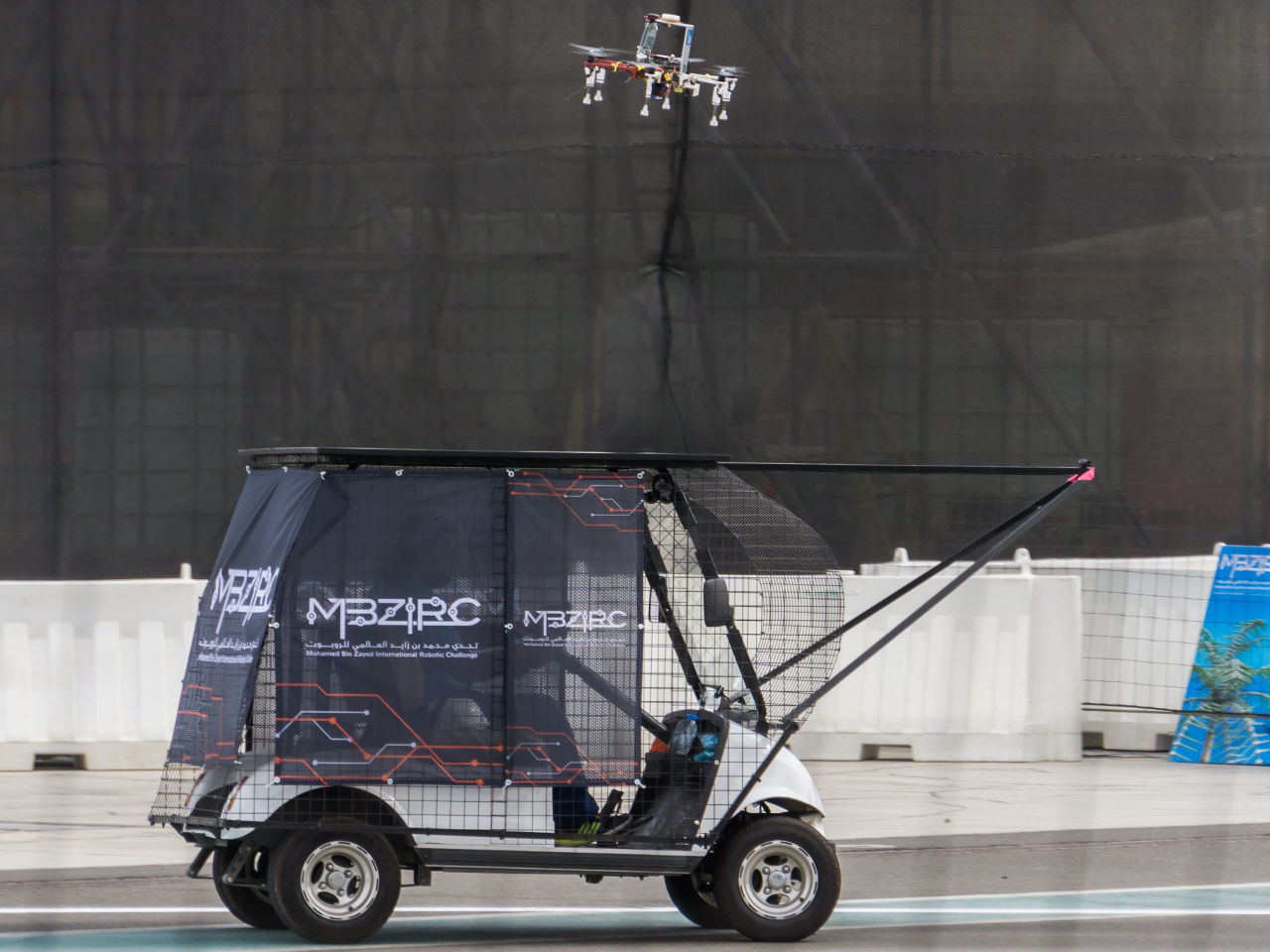}};
      \begin{scope}[x={(a.south east)},y={(a.north west)}]
        \fill[white] (0.001, 0.001) rectangle (0.12,0.13);
        \fill[draw=black, draw opacity=0.5, fill opacity=0] (0,0) rectangle (1, 1);
        \draw (0.06,0.06) node [text=black] {\small (b)};
      \end{scope}
    \end{tikzpicture}}
    \caption{The CTU-UPENN-UoL team during the MBZIRC 2017 competition. The photos show (a) two \acp{UAV} while delivering ferrous objects, and (b) a UAV during autonomous landing on a moving car.}
    \label{fig:mbzirc_2017}
  \end{figure}
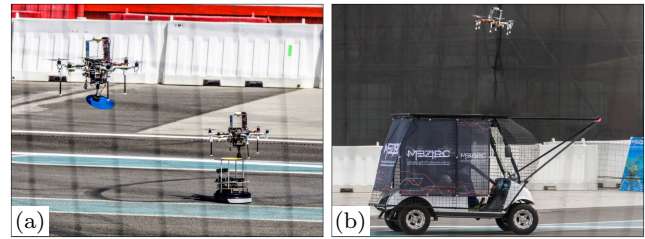

  The \ac{MBZIRC} 2017\footnote{MBZIRC 2017, \url{http://mbzirc.com/challenge/2017}} aimed at pushing the frontiers of field robotics.
  Two tasks out of the three challenges within the competition were focused solely on aerial manipulation and UAV control.
  The competition imposed real-world constraints in its tasks that forced the participating teams to show the current state of the art in robotics and to perform the tasks within a short time window and within specified time slots.
  The first task --- autonomous gathering of colored ferrous objects by a group of \acp{UAV} --- was successfully tackled by the CTU-UPENN-UoL\footnote{Collaboration of Czech Technical University in Prague, University of Pennsylvania, and the University of Lincoln.} team, using the proposed system \cite{spurny2019cooperative, faigl2019unsupervised, loianno2018localization} (see \reffig{fig:mbzirc_2017}).
  We won 1$^{\mathrm{st}}$ place among the best teams from all over the world.
  The second task of autonomous landing on a moving car was also tackled by the proposed system.
  We achieved the fastest autonomous landing among all the teams, and we took the 2$^{\text{nd}}$ place overall in the competition \cite{baca2019autonomous, stepan2019vision}.

  \subsection{The DARPA Subterranean (SubT) challenge}

\begin{figure}
    \centering
    \subfloat {\begin{tikzpicture}
      \node[anchor=south west,inner sep=0] (a) at (0,0) { \includegraphics[width=0.235\textwidth]{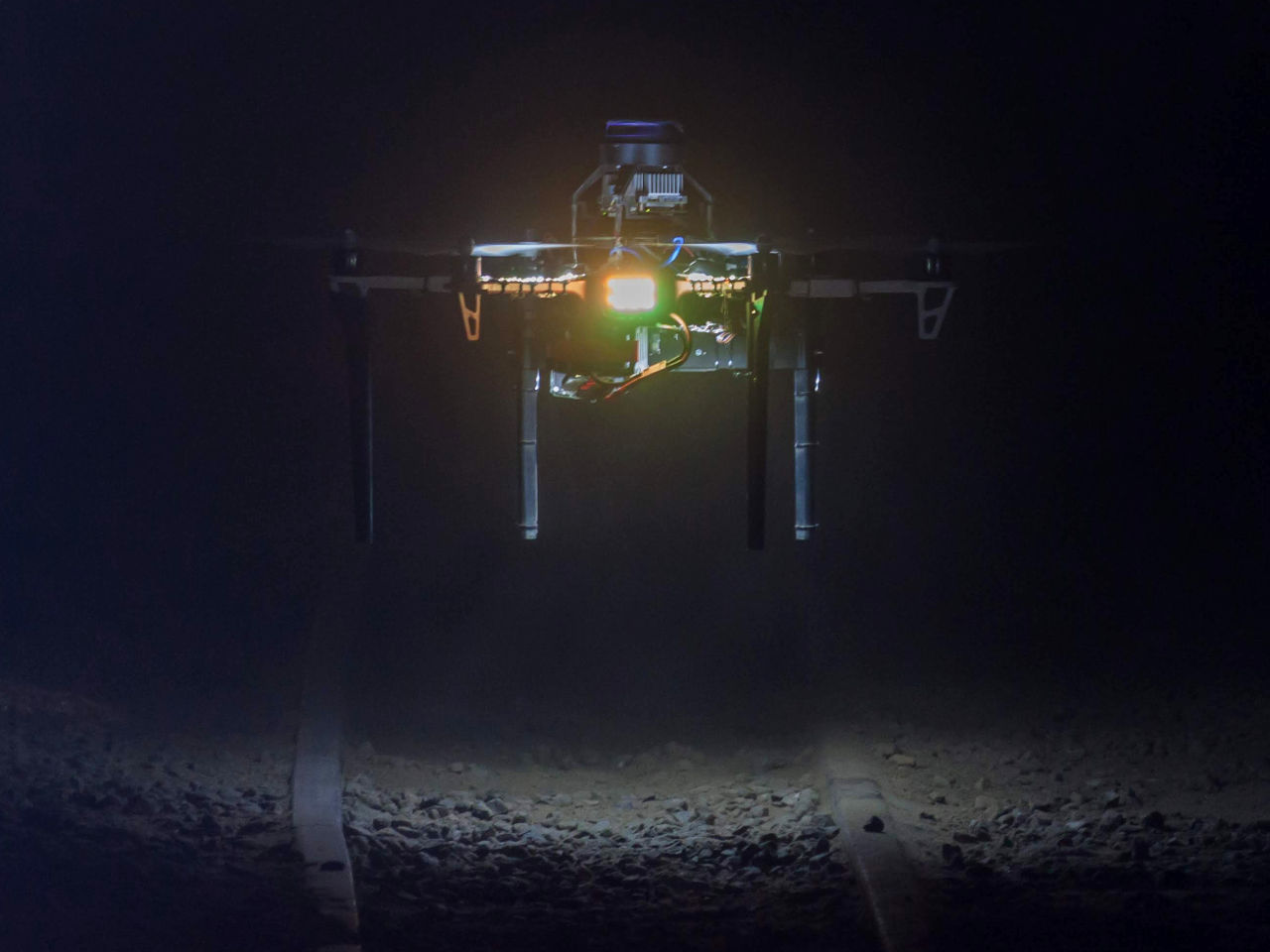}};
      \begin{scope}[x={(a.south east)},y={(a.north west)}]
        \fill[white] (0.001, 0.001) rectangle (0.12,0.13);
        \fill[draw=black, draw opacity=0.5, fill opacity=0] (0,0) rectangle (1, 1);
        \draw (0.06,0.06) node [text=black] {\small (a)};
      \end{scope}
    \end{tikzpicture}}
    \hfill%
    \subfloat {\begin{tikzpicture}
      \node[anchor=south west,inner sep=0] (a) at (0,0) { \includegraphics[width=0.235\textwidth]{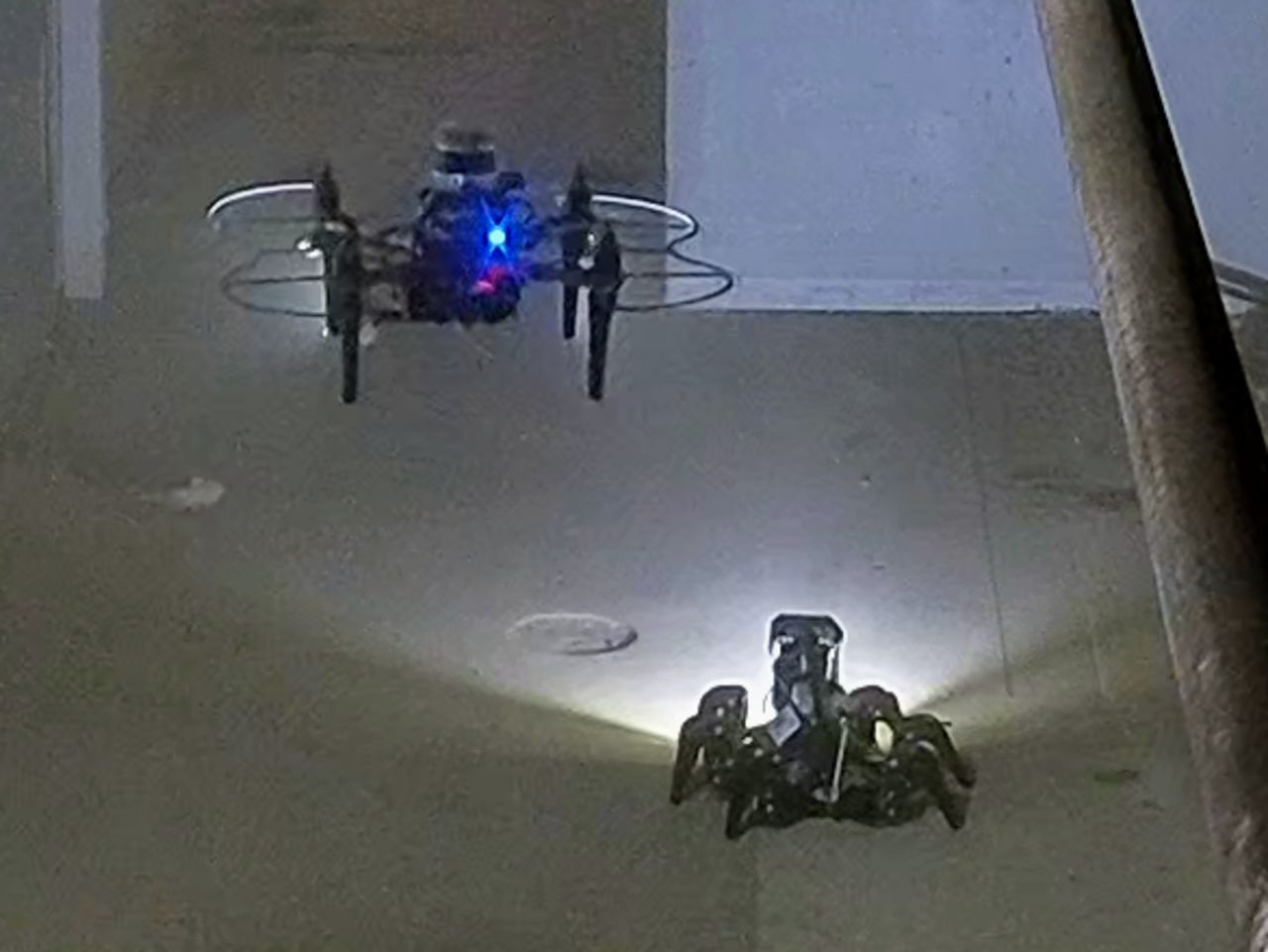}};
      \begin{scope}[x={(a.south east)},y={(a.north west)}]
        \fill[white] (0.001, 0.001) rectangle (0.12,0.13);
        \fill[draw=black, draw opacity=0.5, fill opacity=0] (0,0) rectangle (1, 1);
        \draw (0.06,0.06) node [text=black] {\small (b)};
      \end{scope}
    \end{tikzpicture}}
    \caption{Unmanned Aerial Vehicles during the \ac{DARPA} SubT challenge. The photos depict (a) a UAV exploring an underground mine, and (b) mapping an unfinished nuclear power plant.}
    \label{fig:darpa}
  \end{figure}
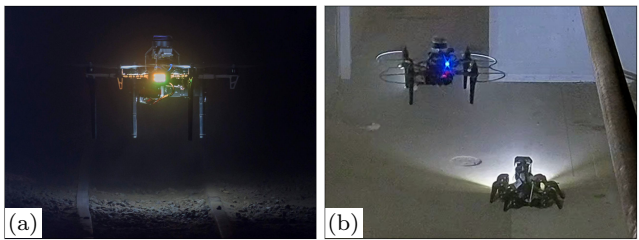

  The \ac{DARPA}, an agency of the United States Department of Defense, organizes series of challenges focused on automatic search \& rescue in an underground environment --- the \ac{DARPA} Subterranean challenge.
  In the \ac{DARPA} Tunnel Circuit, the first round of the challenge, we deployed autonomous \acp{UAV} and semi-autonomous ground robots to explore underground mine shafts \cite{petrlik2020robust, roucek2019darpa}.
  Our team deployed autonomous \acp{UAV} with the proposed system (see \reffig{fig:darpa}), which navigated the underground tunnels and returned safely to the entrance while autonomously localizing objects of interest.
  We won the 1$^{\mathrm{st}}$ prize among the self-funded teams and the 3$^{\mathrm{rd}}$ prize overall.
  To the best of our knowledge, our \acp{UAV} managed to explore a greater distance into the tunnels than any of the other teams.

  In the \ac{DARPA} Urban Circuit, the second round of the challenge, we deployed autonomous \acp{UAV} and semi-autonomous ground robots to explore the infrastructure of an unfinished nuclear power plant.
  Our \acp{UAV} managed to explore \SI{2867}{\meter\cubed} of one floor of the reactor building while automatically navigating up to \SI{100}{\meter} in just \SI{200}{\second} in a completely unknown environment.
  We again took 1$^{\text{st}}$ place among the self-funded teams, and 3$^{\text{rd}}$ place overall.
  Scientific publications on tasks within the Urban Circuit are under preparation.

  \subsection{MBZIRC 2020 competition}

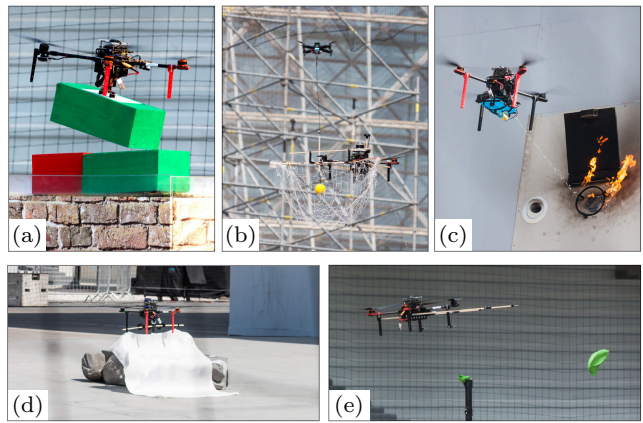
\begin{figure}
    \centering
    \subfloat {\begin{tikzpicture}
      \node[anchor=south west,inner sep=0] (a) at (0,0) { \includegraphics[width=0.155\textwidth]{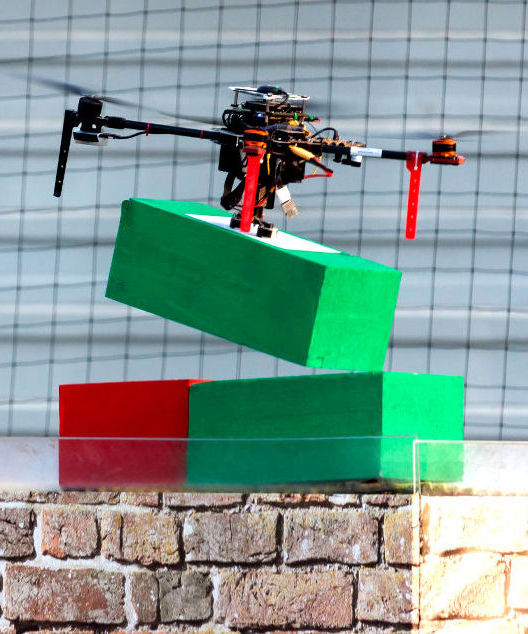}};
      \begin{scope}[x={(a.south east)},y={(a.north west)}]
        \fill[white] (0.001, 0.001) rectangle (0.18,0.13);
        \fill[draw=black, draw opacity=0.5, fill opacity=0] (0,0) rectangle (1, 1);
        \draw (0.09,0.06) node [text=black] {\small (a)};
      \end{scope}
    \end{tikzpicture}}
    \hfill%
    \subfloat {\begin{tikzpicture}
      \node[anchor=south west,inner sep=0] (a) at (0,0) { \includegraphics[width=0.155\textwidth]{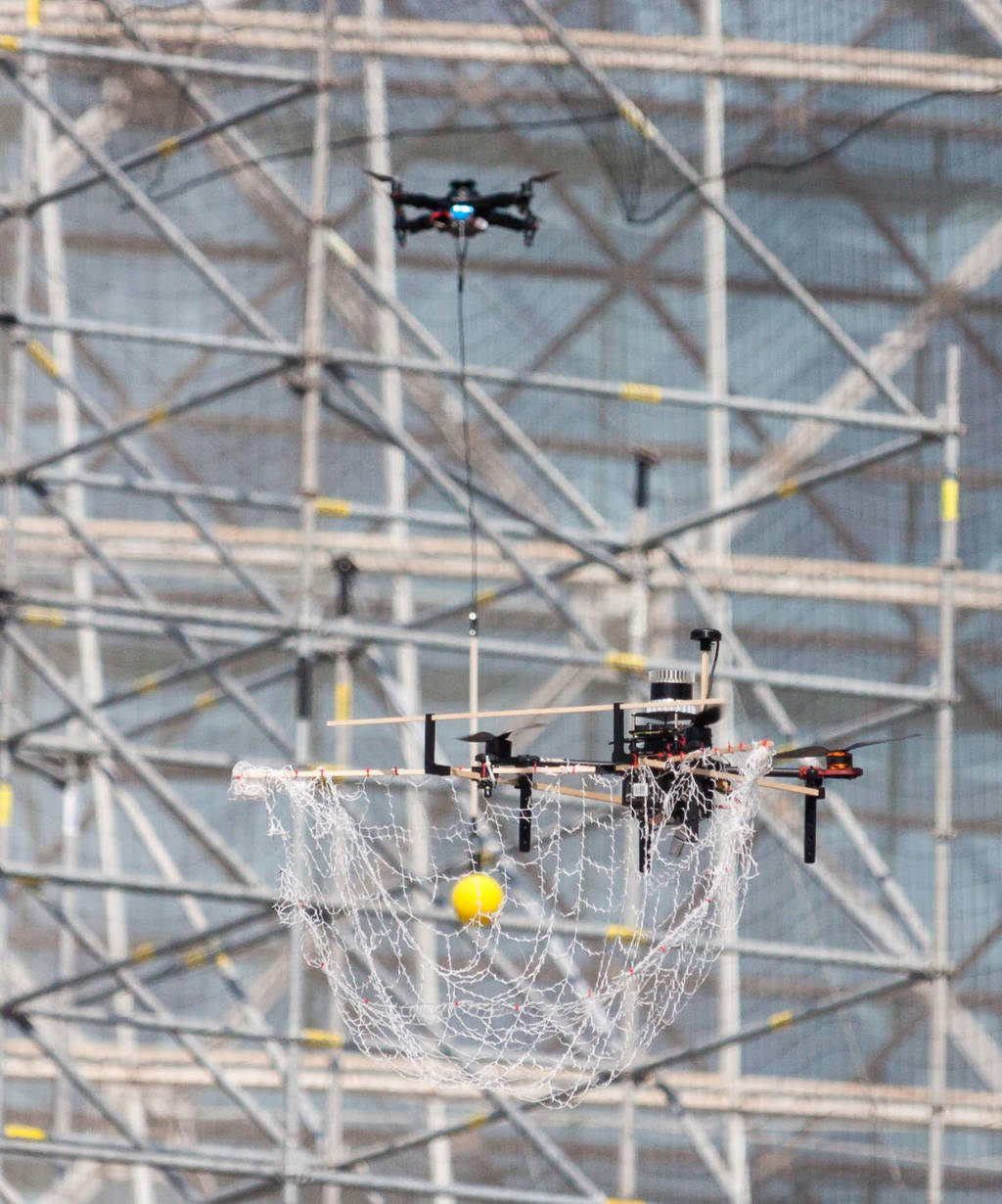}};
      \begin{scope}[x={(a.south east)},y={(a.north west)}]
        \fill[white] (0.001, 0.001) rectangle (0.18,0.13);
        \fill[draw=black, draw opacity=0.5, fill opacity=0] (0,0) rectangle (1, 1);
        \draw (0.09,0.06) node [text=black] {\small (b)};
      \end{scope}
    \end{tikzpicture}}
    \hfill%
    \subfloat {\begin{tikzpicture}
      \node[anchor=south west,inner sep=0] (a) at (0,0) { \includegraphics[width=0.155\textwidth]{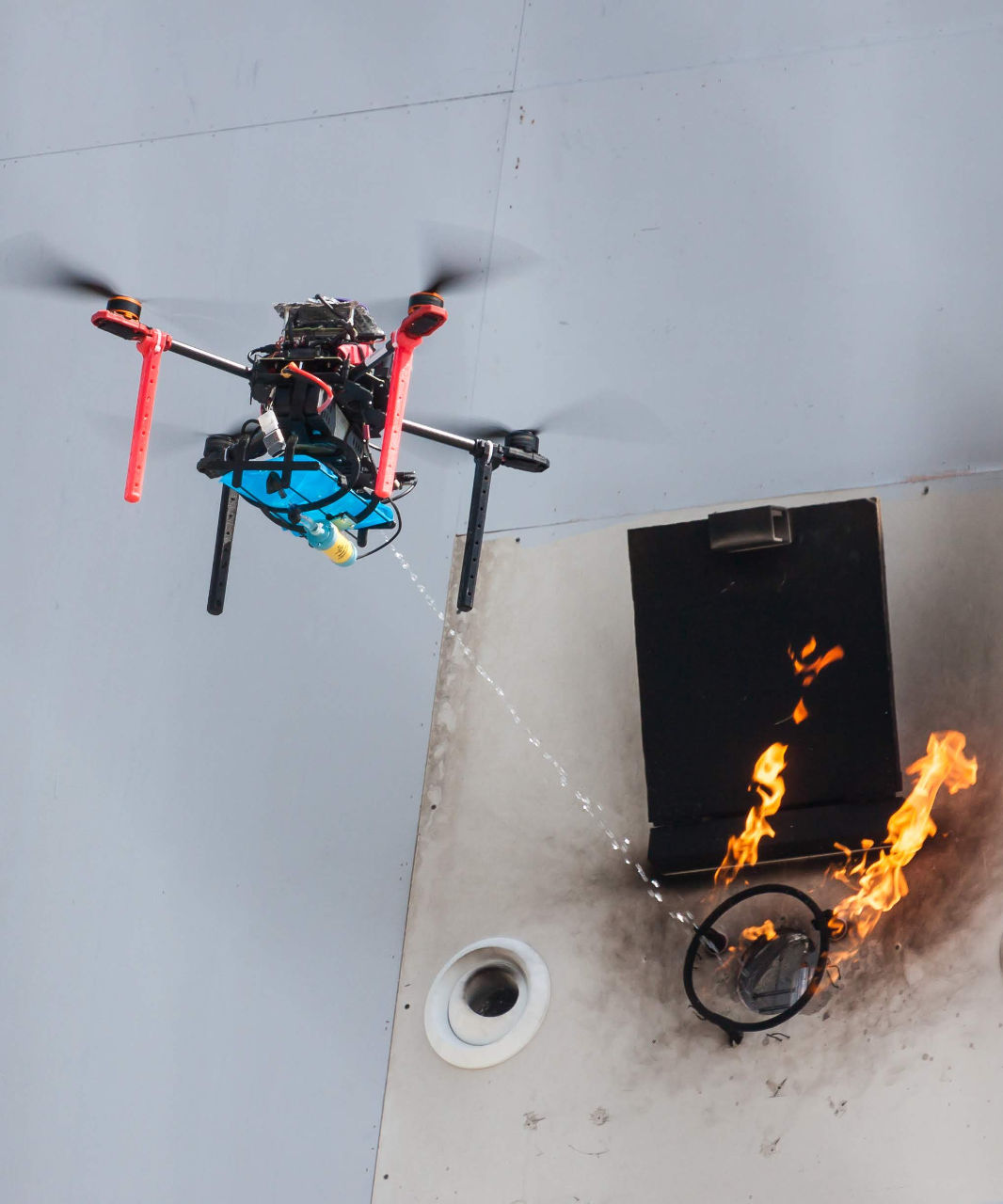}};
      \begin{scope}[x={(a.south east)},y={(a.north west)}]
        \fill[white] (0.001, 0.001) rectangle (0.18,0.13);
        \fill[draw=black, draw opacity=0.5, fill opacity=0] (0,0) rectangle (1, 1);
        \draw (0.09,0.06) node [text=black] {\small (c)};
      \end{scope}
    \end{tikzpicture}}\\
    \vspace{-0.8em}
    \subfloat {\begin{tikzpicture}
      \node[anchor=south west,inner sep=0] (a) at (0,0) { \includegraphics[width=0.235\textwidth]{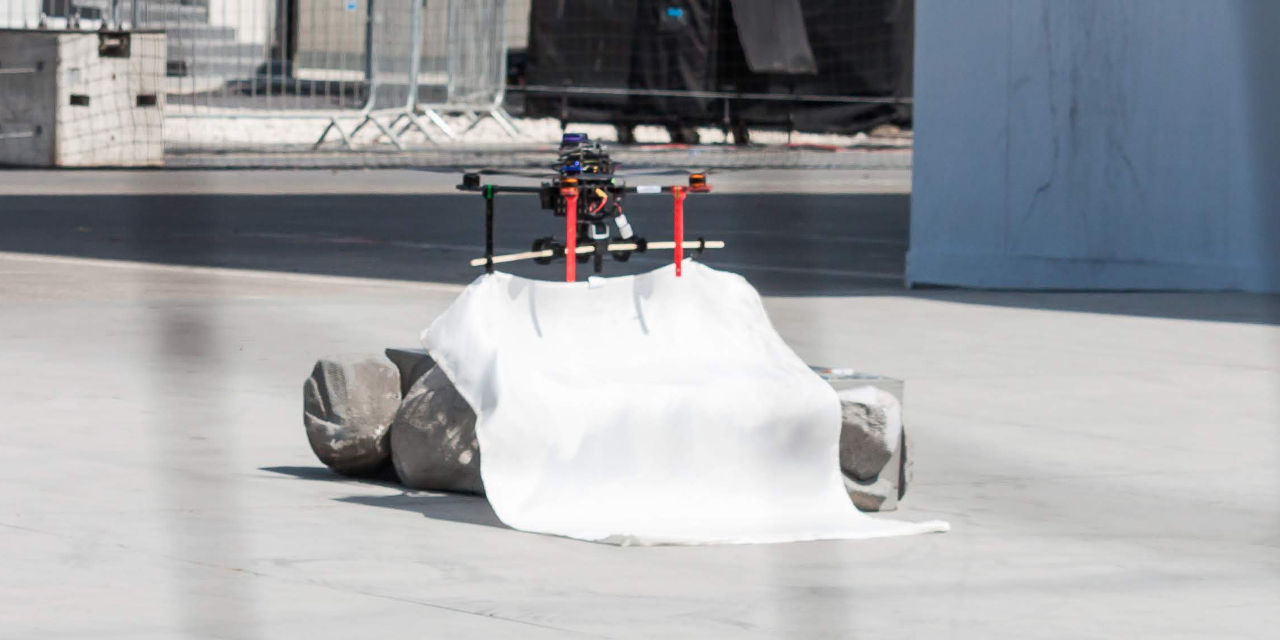}};
      \begin{scope}[x={(a.south east)},y={(a.north west)}]
        \fill[white] (0.001, 0.001) rectangle (0.13,0.20);
        \fill[draw=black, draw opacity=0.5, fill opacity=0] (0,0) rectangle (1, 1);
        \draw (0.065,0.090) node [text=black] {\small (d)};
      \end{scope}
    \end{tikzpicture}}
    \hfill%
    \subfloat {\begin{tikzpicture}
      \node[anchor=south west,inner sep=0] (a) at (0,0) { \includegraphics[width=0.235\textwidth]{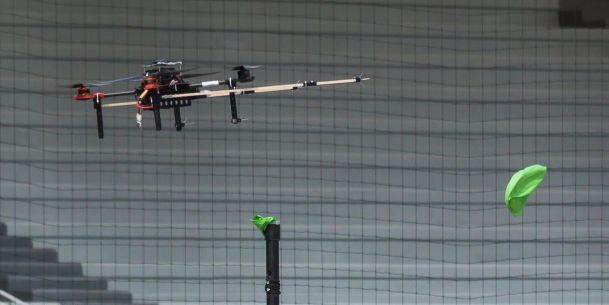}};
      \begin{scope}[x={(a.south east)},y={(a.north west)}]
        \fill[white] (0.001, 0.001) rectangle (0.13,0.20);
        \fill[draw=black, draw opacity=0.5, fill opacity=0] (0,0) rectangle (1, 1);
        \draw (0.065,0.090) node [text=black] {\small (e)};
      \end{scope}
    \end{tikzpicture}}
    \caption{The CTU-UPENN-NYU team during the MBZIRC 2020 competition. The photos depict (a) autonomous wall building, (b) autonomous ball catching, (c) autonomous fire extinguishing, (d) autonomous fire blanket deployment, and (e) autonomous balloon popping.}
    \label{fig:mbzirc_2020}
  \end{figure}

  The second round of the \ac{MBZIRC} competition was organized in 2020.
  It pushed the current state of the art in aerial robotics to its limits, with tasks such as organizing a group of \acp{UAV} and a \ac{UGV} to build a brick wall autonomously, autonomous indoor and outdoor firefighting with \acp{UAV}, and autonomously catching a ball carried by a \ac{UAV}, performed simultaneously with balloon popping by a group of \acp{UAV} (see \reffig{fig:mbzirc_2020}).
  All of the tasks were solved using the proposed UAV system, and our participation in the competition helped to consolidate many of the platform's functionalities.
  The CTU-UPENN-NYU\footnote{Collaboration between the Czech Technical University in Prague, the University of Pennsylvania, and the New York University.} team achieved the highest score of all the teams for building the brick wall autonomously.
  We also took $2^{\text{nd}}$ in the autonomous balloon popping and ball-catching task.
  We won the gold medal in the \emph{grand challenge} in which all the tasks were tested simultaneously.
  Scientific publications reporting on \ac{MBZIRC} 2020 are under preparation \added{\cite{baca2020autonomous}}.

  \subsection{IEEE RAS Summer School on Multi-robot Systems}

  The proposed system was used as an educational tool during the 2019 \ac{IEEE} \ac{RAS} summer school on multirobot systems\footnote{\url{http://mrs.felk.cvut.cz/summer-school-2019}}.
  More than 70 international students were challenged to solve a multi-UAV Dubins traveling salesman problem with neighborhoods during the summer school exercises.
  Student solutions were put to test during an outdoor experimental session.



  \section{CONCLUSIONS}
  \label{sec:conclusions}

  We have presented a multirotor \ac{UAV} control and estimation system created with emphasis on realistic simulations and real-world experiments.
  The system is a product of years of cutting-edge research on aerial systems and their use in various branches of autonomous robotics.
  The proposed architecture allows reliable deployment of \acp{UAV} outside laboratory conditions using only onboard sensors.
  The proposed control pipeline supports fast and agile maneuvers as well as safe flight even with noisy and unreliable sensors.
  We have provided a well-documented and open-source implementation, which is being actively used by many researchers in the field.
  The \ac{MRS} team at CTU in Prague has achieved outstanding results in robotics challenges and competitions using this system.
  The experience gained from the challenges helped to shape the proposed system into the presented form.




\section{Declarations}

\subsection{Ethical Approval}

All applicable institutional and national guidelines were followed.

\subsection{Consent to Participate}

Not applicable.

\subsection{Consent to Publish}

Informed consent was obtained from all the co-authors of this publication.

\subsection{Authors Contributions}

All authors contributed to the research, development, and testing of the proposed system. Tomas Baca conceptualized the proposed system and is the author of the proposed controller design and control architecture. He also leads the software development of the system. Matej Petrlik is the author of the multi-frame state estimator. Matous Vrba is responsible for maintaining low-level software libraries related to our efficient implementation of LKF, UKF, and ROS wrappers. Vojtech Spurny and Robert Penicka are responsible for maintaining the simulation pipeline and related subsystems of the software ecosystem. Daniel Hert is responsible for designing and maintaining the UAV hardware used for real-world experiments. Martin Saska is the head of the MRS group, CTU in Prague, and he provided us with the necessary guidance, funding and final proofreading. The first draft of the manuscript was written by Tomas Baca, and all authors commented on previous versions of the manuscript. All authors read and approved the final manuscript.

\subsection{Funding}

This work was supported by CTU grant no SGS20/174/OHK3/3T/13, by the Czech Science Foundation under research project No. 20-10280S, by Ministry of Education of the Czech Republic project CZ.02.1.01/0.0/0.0/16 019/0000765 ``Research Center for Informatics'', and by the European Union's Horizon 2020 research and innovation program under grant agreement No 871479.

\subsection{Competing Interests}

The authors declare that they have no conflict of interest.

\subsection{Availability of data and materials}

The presented software is provided as open-source at \url{https://github.com/ctu-mrs/mrs_uav_system}.
Additional multimedia materials are available at \url{http://mrs.felk.cvut.cz/mrs-uav-system}.





  \bibliographystyle{spmpsci}      
  \bibliography{main}


\end{document}

%% file: src/tikz.tex
\tikzset{
  state/.style={
    rectangle,
    draw=black, very thick,
    minimum height=1.0em,
    text centered,
  },
  smallstate/.style={
    rectangle,
    draw=black, very thick,
    minimum height=0.2em,
    text centered,
  },
  final_state/.style={
    rectangle,
    rounded corners,
    draw=black, very thick,
    minimum height=2em,
    text centered,
  },
  initial_state/.style={
    rectangle,
    double=white,
    double distance=1pt,
    inner sep=2pt,
    draw=black, very thick,
    minimum height=2em,
    text centered,
  },
  point/.style={
    circle,
    inner sep=0pt,
    minimum size=3pt,
    fill=red
  },
  adder/.style={
    circle,
    inner sep=2pt,
    minimum size=0.3in,
    draw=black, very thick,
    text centered
  },
  state_gray/.style={
    rectangle,
    draw=black, very thick,
    fill=gray!40,
    minimum height=1.0em,
    text centered,
    inner sep=0,
  },
  state_white/.style={
    rectangle,
    draw=black, very thick,
    fill=white,
    minimum height=1.0em,
    text centered,
    text=black,
    inner sep=0,
  },
  state_green/.style={
    rectangle,
    draw=black, very thick,
    fill=green!50,
    minimum height=1.0em,
    text centered,
    text=black,
    inner sep=0,
  },
  state_red/.style={
    rectangle,
    draw=black, very thick,
    fill=red!70,
    minimum height=1.0em,
    text centered,
    text=black,
    inner sep=0,
  },
  state_blue/.style={
    rectangle,
    draw=black, very thick,
    fill=blue!40,
    minimum height=1.0em,
    text centered,
    text=black,
    inner sep=0,
  },
  final_state/.style={
    rectangle,
    rounded corners,
    draw=black, very thick,
    minimum height=2em,
    text centered,
  },
  initial_state/.style={
    rectangle,
    double=white,
    double distance=1pt,
    inner sep=2pt,
    draw=black, very thick,
    minimum height=2em,
    text centered,
  },
  point/.style={
    circle,
    inner sep=0pt,
    minimum size=3pt,
    fill=red
  },
}

%% file: fig/tikz/pipeline_diagram.tex
\pgfdeclarelayer{foreground}
\pgfsetlayers{background,main,foreground}

\tikzset{radiation/.style={{decorate,decoration={expanding waves,angle=90,segment length=4pt}}}}

\begin{tikzpicture}[->,>=stealth', node distance=3.0cm,scale=1.0, every node/.style={scale=1.0}]


  \node[state, shift = {(0.0, 0.0)}] (navigation) {
      \begin{tabular}{c}
        \footnotesize Mission \&\\
        \footnotesize navigation
      \end{tabular}
    };


  \node[state, right of = navigation, shift = {(0.7, 0)}] (tracker) {
      \begin{tabular}{c}
        \footnotesize Reference \\
        \footnotesize tracker
      \end{tabular}
    };

  \node[state, right of = tracker, shift = {(0.1, 0)}] (controller) {
      \begin{tabular}{c}
        \footnotesize Reference \\
        \footnotesize controller
      \end{tabular}
    };

  \node[state, right of = controller, shift = {(0.8, -0)}] (attitude) {
      \begin{tabular}{c}
        \footnotesize Attitude rate\\
        \footnotesize controller
      \end{tabular}
    };

  \node[smallstate, below of = attitude, shift = {(-0.6, 2.1)}] (imu) {
      \footnotesize IMU
    };

  \node[state, right of = attitude, shift = {(0.7, -0)}] (actuators) {
      \begin{tabular}{c}
        \footnotesize UAV \\
        \footnotesize actuators
      \end{tabular}
    };

  \node[state, right of = actuators, shift = {(-0.8, -0)}] (sensors) {
      \begin{tabular}{c}
        \footnotesize Onboard \\
        \footnotesize sensors
      \end{tabular}
    };

  \node[state, below of = attitude, shift = {(0, 0.9)}] (estimator) {
      \begin{tabular}{c}
        \footnotesize State \\
        \footnotesize estimator
      \end{tabular}
    };

  \node[state, right of = estimator, shift = {(0.8, 0.0)}] (localization) {
      \begin{tabular}{c}
        \footnotesize Odometry \& \\
        \footnotesize localization
      \end{tabular}
    };



  \path[->] ($(navigation.east) + (0.0, 0)$) edge [] node[above, midway, shift = {(0.0, 0.05)}] {
      \begin{tabular}{c}
        \footnotesize desired reference\\
        \footnotesize $\mathbf{r}_d, \eta_d$\\
        \footnotesize \textit{on demand}
    \end{tabular}} ($(tracker.west) + (0.0, 0.00)$);


  \path[->] ($(tracker.east) + (0.0, 0)$) edge [] node[above, midway, shift = {(0.0, 0.05)}] {
      \begin{tabular}{c}
        \footnotesize full-state reference\\
        \footnotesize $\bm{\chi}_d$\\
        \footnotesize \SI{100}{\hertz}
    \end{tabular}} ($(controller.west) + (0.0, 0.00)$);

  \path[->] ($(tracker.south |- estimator.west) + (0.0, 0.0)$) edge [dotted] node[left, midway, shift = {(0.2, 0.00)}] {
      \begin{tabular}{r}
        \scriptsize initialization\\[-0.5em]
        \scriptsize only
    \end{tabular}} ($(tracker.south) + (0.0, 0.00)$);

  \path[->] ($(controller.east) + (0.0, 0)$) edge [] node[above, midway, shift = {(-0.2, 0.05)}] {
      \begin{tabular}{c}
        \footnotesize $\bm{\omega}_d$\\
        \footnotesize $T_d$ \\
        \footnotesize \SI{100}{\hertz}
    \end{tabular}} ($(attitude.west) + (0.0, 0.00)$);

  \draw[-] ($(controller.south)+(0.25,0)$) -- ($(controller.south |- estimator.west) + (0.25, 0.25)$) edge [->] node[above, near start, shift = {(-0.2, 0.05)}] {
      \begin{tabular}{c}
        \footnotesize $\mathbf{a}_d$
    \end{tabular}} ($(estimator.west) + (0, 0.25)$);

  \path[->] ($(attitude.east) + (0.0, 0)$) edge [] node[above, midway, shift = {(0.1, 0.05)}] {
      \begin{tabular}{c}
        \footnotesize $\bm{\tau}_d$ \\
        \footnotesize $\approx$\SI{1}{\kilo\hertz}
    \end{tabular}} ($(actuators.west) + (0.0, 0.00)$);

  \path[-] ($(estimator.west)+(0, 0)$) edge [] node[above, near start, shift = {(-1.0, 0.0)}] {
      \begin{tabular}{c}
        \footnotesize $\mathbf{x}$, $\mathbf{R}$, $\bm{\omega}$\\
        \footnotesize \SI{100}{\hertz}
    \end{tabular}} ($(navigation.south |- estimator.west)$) -- ($(navigation.south |- estimator.west)$) edge [->,] ($(navigation.south)+(0, 0)$);


  \path[->] ($(controller.south |- estimator.west)+(0, 0)$) edge [] ($(controller.south)$);

  \draw[-] ($(imu.east) + (0.0, 0.0)$) -- ($(estimator.north |- imu.east) + (0.3, 0)$) edge [->] node[right, midway, shift = {(-0.2, 0.3)}] {
      \begin{tabular}{c}
        \footnotesize $\mathbf{R}$, $\bm{\omega}$
    \end{tabular}} ($(estimator.north) + (0.3, 0.0)$);

  \draw[-] ($(sensors.south)+(0, 0)$) -- ($(sensors.south |- localization.east)$) edge [->] ($(localization.east)$);
  \draw[-] ($(sensors.south)+(0.1, 0)$) -- ($(sensors.south |- localization.east) + (0.1, -0.1)$) edge [->] ($(localization.east) + (0.0, -0.1)$);
  \draw[-] ($(sensors.south)+(-0.1, 0)$) -- ($(sensors.south |- localization.east) + (-0.1, 0.1)$) edge [->]  ($(localization.east) + (0.0, 0.1)$);

  \draw[->] ($(localization.west)+(0, 0)$) -- ($(estimator.east)$);
  \draw[->] ($(localization.west)+(0, 0.1)$) -- ($(estimator.east) + (0, 0.1)$);
  \draw[->] ($(localization.west)+(0, -0.1)$) -- ($(estimator.east) + (0, -0.1)$);




  \begin{pgfonlayer}{background}
    \path (attitude.west |- attitude.north)+(-0.45,0.8) node (a) {};
    \path (imu.south -| sensors.east)+(+0.25,-0.20) node (b) {};
    \path[fill=gray!3,rounded corners, draw=black!70, densely dotted]
      (a) rectangle (b);
  \end{pgfonlayer}
  \node [rectangle, above of=actuators, shift={(-0.6,0.55)}, node distance=1.7em] (autopilot) {\footnotesize UAV plant};

  \begin{pgfonlayer}{background}
    \path (attitude.west |- attitude.north)+(-0.25,0.47) node (a) {};
    \path (imu.south -| attitude.east)+(+0.25,-0.10) node (b) {};
    \path[fill=gray!3,rounded corners, draw=black!70, densely dotted]
      (a) rectangle (b);
  \end{pgfonlayer}
  \node [rectangle, above of=attitude, shift={(0,0.2)}, node distance=1.7em] (autopilot) {\footnotesize Embedded autopilot};


\end{tikzpicture}

%% file: fig/tikz/bank_of_filters.tex
\usetikzlibrary{shapes.geometric,backgrounds,calc,arrows}
\pgfdeclarelayer{background}
\pgfdeclarelayer{foreground}
\pgfsetlayers{background,main,foreground}

\tikzset{radiation/.style={{decorate,decoration={expanding waves,angle=90,segment length=4pt}}}}

\begin{tikzpicture}[->,>=stealth', node distance=3.0cm]


  \def\filtx{10pt}
  \def\dotoff{1.5}

  \node[state, shift = {(0, -0)}] (k1) {
      \begin{tabular}{c}
        \small $K_1$
      \end{tabular}
    };

  \node[state, right of = k1, shift = {(-\filtx, -0)}] (k2) {
      \begin{tabular}{c}
        \small $K_2$
      \end{tabular}
    };

  \node[right of = k2, shift = {(-\dotoff, -0)}] (kdots) {
      \begin{tabular}{c}
        \small $\cdots$
      \end{tabular}
    };

  \node[state, right of = kdots, shift = {(-\dotoff, -0)}] (kn) {
      \begin{tabular}{c}
        \small $K_n$
      \end{tabular}
    };



  \def\eloffx{1.5}
  \def\eloffy{1.2}
  \def\ellx{20pt}
  \def\elly{22pt}
  \node[ellipse, minimum height=\elly, minimum width=\ellx, draw, above left of = k1, shift = {(\eloffx, -\eloffy)}] (pred1) {
      \small pred
    };

  \path[->] (k1.west) [bend left]  edge node {} (pred1);
  \path[->] (pred1) [bend left]  edge node {} ([xshift=-5pt]k1.north);

  \node[ellipse, minimum height=\elly, minimum width=\ellx,  draw, above right of = k1, shift = {(-\eloffx, -\eloffy)}] (corr1) {
    \small corr
  };

  \path[->] (k1.east) [bend right]  edge node {} (corr1);
  \path[->] (corr1) [bend right]  edge node {} ([xshift=5pt]k1.north);

  \node[ellipse, minimum height=\elly, minimum width=\ellx,  draw, above left of = k2, shift = {(\eloffx, -\eloffy)}] (pred2) {
    \small pred
  };

  \path[->] (k2.west) [bend left]  edge node {} (pred2);
  \path[->] (pred2) [bend left]  edge node {} ([xshift=-5pt]k2.north);

  \node[ellipse, minimum height=\elly, minimum width=\ellx,  draw, above right of = k2, shift = {(-\eloffx, -\eloffy)}] (corr2) {
    \small corr
  };

  \path[->] (k2.east) [bend right]  edge node {} (corr2);
  \path[->] (corr2) [bend right]  edge node {} ([xshift=5pt]k2.north);

  \node[ellipse, minimum height=\elly, minimum width=\ellx,  draw, above left of = kn, shift = {(\eloffx, -\eloffy)}] (predn) {
    \small pred
  };

  \path[->] (kn.west) [bend left]  edge node {} (predn);
  \path[->] (predn) [bend left]  edge node {} ([xshift=-5pt]kn.north);

  \node[ellipse, minimum height=\elly, minimum width=\ellx,  draw, above right of = kn, shift = {(-\eloffx, -\eloffy)}] (corrn) {
    \small corr
  };

  \path[->] (kn.east) [bend right]  edge node {} (corrn);
  \path[->] (corrn) [bend right]  edge node {} ([xshift=5pt]kn.north);


  
    \def\apredy{-2}
  
    \node[above of = pred1, shift = {(0, \apredy)}] (apred1) {
    };
  
    \node[above of = pred2, shift = {(0, \apredy)}] (apred2) {
    };
  
    \node[above of = predn, shift = {(0, \apredy)}] (apredn) {
    };
  
    \node[left of = apred1, shift = {(2, 0)}] (input) {
    };
  
    \path[-] (input.center) edge node [above, shift = {(-0.3,0)}] {$\mathbf{u}$} (apred1.center);
    \path[-] (apred1.center) edge node {} (apred2.center);
    \path[-] (apred2.center) edge node {} (apredn.center);
  
    \path[->] (apred1.center) edge node {} (pred1.north);
    \path[->] (apred2.center) edge node {} (pred2.north);
    \path[->] (apredn.center) edge node {} (predn.north);
  

  
    \def\apredy{-2.2}
  
    \node[above of = corr1, shift = {(0, \apredy)}] (acorr1) {
    };
  
    \node[above of = corr2, shift = {(0, \apredy)}] (acorr2) {
    };
  
    \node[above of = corrn, shift = {(0, \apredy)}] (acorrn) {
    };
  
    \node[left of = acorr1, shift = {(1, 0)}] (measurement) {
    };
  
    \path[-] (measurement.center) edge node [below, shift = {(-0.8,0)}] {$\mathbf{z}$} (acorr1.center);
    \path[-] (acorr1.center) edge node {} (acorr2.center);
    \path[-] (acorr2.center) edge node {} (acorrn.center);
  
    \path[->] (acorr1.center) edge node {} (corr1.north);
    \path[->] (acorr2.center) edge node {} (corr2.north);
    \path[->] (acorrn.center) edge node {} (corrn.north);
  


  \def\mararb{5pt}
  \def\cs{5pt}
  \def\offcx{2}
  \def\offcy{1.5}

  \node[circle, inner sep=0pt, minimum size=\cs,  draw, below of = k2, shift = {(0,\offcy)}] (contact2) {
  };

  \node[circle, inner sep=0pt, minimum size=\cs,  draw, right of = contact2, shift = {(-\offcx, 0)}] (contactn) {
  };

  \node[circle, fill, inner sep=0pt, minimum size=\cs,  draw, left of = contact2, shift = {(\offcx, 0)}] (contact1) {
  };

  \def\pbky{10pt}
  \node[below of = k1, shift = {(0, 1.5*\offcy)}] (pbk1) {
  };

  \node[below of = kn, shift = {(0, 1.5*\offcy)}] (pbkn) {
  };

  \node[above of = contact1, shift = {(0, -1.5*\offcy)}] (pacontact1) {
  };

  \node[above of = contactn, shift = {(0, -1.5*\offcy)}] (pacontactn) {
  };


  \node[below of = contact2, shift = {(0, 2.5)}] (contactb) {
  };

  \node[below of = contactb, shift = {(0, 2.5)}] (barbiter) {
  };

  \node[right of = barbiter, shift = {(0, 0)}] (estimate) {
  };

  \path[-] (k1.south) edge node {} (pbk1.center);
  \path[-] (pbk1.center) edge node {} (pacontact1.center);
\path[-] (pacontact1.center) edge node [below left, shift ={(-0.5,0.4)}] {$\mathbf{x}_1, \mathbf{\Sigma}_1$} (contact1.north);

\path[-] (k2.south) edge node [right, shift = {(0,0.34)}] {$\mathbf{x}_2, \mathbf{\Sigma}_2$} (contact2.north);

  \path[-] (kn.south) edge node {} (pbkn.center);
  \path[-] (pbkn.center) edge node {} (pacontactn.center);
\path[-] (pacontactn.center) edge node [below right, shift = {(0.5,0.4)}] {$\mathbf{x}_n, \mathbf{\Sigma}_n$} (contactn.north);

  \path[-] (contact1.south) edge node {} (contactb.center);
  \path[-,dotted] (contact2.south) edge node {} (contactb.center);
  \path[-,dotted] (contactn.south) edge node {} (contactb.center);

  \path[-] (contactb.center) edge node {} (barbiter.center);


  \node[state, inner sep = 0cm, minimum width=2.8cm, minimum height = 1cm, below of = k2, shift = {(-0, \offcy-0.25)}] (arbiter) {
    \begin{tabular}{ccccr}
      \\
      & & & & \small Arbiter
    \end{tabular}
  };


  
    \node[right of = barbiter, shift = {(0, 0)}] (output) {
    };
  
    \path[->] (barbiter.center) edge node [above, shift = {(1,0)}] {$\mathbf{x}_{\ast }$} (output.center);


\end{tikzpicture}

%% file: fig/tikz/implementation_diagram.tex
\pgfdeclarelayer{foreground}
\pgfsetlayers{background,main,foreground}

\tikzset{radiation/.style={{decorate,decoration={expanding waves,angle=90,segment length=0.1em}}}}

\begin{tikzpicture}[->,>=stealth', node distance=6.0em]


  \node[state_blue, shift = {(0.0em, 0.0em)}] (control_manager) {
    \begin{tabular}{c}
      \scriptsize Control manager
    \end{tabular}
  };

  \node[state_red, right of = control_manager, shift = {(2.0em, 0.0em)}] (mpc_tracker) {
    \begin{tabular}{c}
      \scriptsize MPC tracker
    \end{tabular}
  };

  \node[state_red, below of = mpc_tracker, shift = {(0.0em, 4.5em)}] (landoff_tracker) {
    \begin{tabular}{c}
      \scriptsize Landoff tracker
    \end{tabular}
  };

  \node[state_blue, above of = mpc_tracker, shift = {(0.0em, -4.0em)}] (constraint_manager) {
    \begin{tabular}{c}
      \scriptsize Constraint manager
    \end{tabular}
  };

  \node[state_blue, above of = constraint_manager,  shift = {(-1.0em, -4.5em)}] (uav_manager) {
    \begin{tabular}{c}
      \scriptsize UAV manager
    \end{tabular}
  };

  \node[state_red, below of = landoff_tracker, shift = {(0.0em, 4.5em)}] (speed_tracker) {
    \begin{tabular}{c}
      \scriptsize Speed tracker
    \end{tabular}
  };

  \node[below of = speed_tracker, shift = {(0.0em, 4.5em)}] (dots_tracker) {
    \begin{tabular}{c}
      $\vdots$
    \end{tabular}
  };

  \node[state_green, left of = control_manager, shift = {(-2.0em, 0.0em)}] (se3_controller) {
    \begin{tabular}{c}
      \scriptsize SE(3) controller
    \end{tabular}
  };

  \node[state_green, below of = se3_controller, shift = {(0.0em, 4.5em)}] (mpc_controller) {
    \begin{tabular}{c}
      \scriptsize MPC controller
    \end{tabular}
  };

  \node[state_blue, above of = se3_controller, shift = {(0.0em, -4.0em)}] (gain_manager) {
    \begin{tabular}{c}
      \scriptsize Gain manager
    \end{tabular}
  };

  \node[state_green, below of = mpc_controller, shift = {(0.0em, 4.5em)}] (failsafe_controller) {
    \begin{tabular}{c}
      \scriptsize Failsafe controller
    \end{tabular}
  };

  \node[below of = failsafe_controller, shift = {(0.0em, 4.5em)}] (dots_controller) {
    \begin{tabular}{c}
      $\vdots$
    \end{tabular}
  };

  \node[state_white, below of = control_manager, shift = {(0.0em, 4.2em)}] (mavros2) {
    \begin{tabular}{c}
      \scriptsize Mavros
    \end{tabular}
  };


  \node[state_gray, below of = mavros2, shift = {(0.0em, 4.2em)}] (pixhawk2) {
    \begin{tabular}{c}
      \scriptsize Pixhawk
    \end{tabular}
  };

  \node[state_white, above of = control_manager, shift = {(0.0em, -1.0em)}] (estimator) {
    \begin{tabular}{c}
      \scriptsize State estimator
    \end{tabular}
  };

  \node[state_white, above of = estimator, shift = {(0.0em, -4.0em)}] (mavros) {
    \begin{tabular}{c}
      \scriptsize Mavros
    \end{tabular}
  };

  \node[state_gray, above of = mavros, shift = {(0.0em, -4.0em)}] (pixhawk) {
    \begin{tabular}{c}
      \scriptsize Pixhawk
    \end{tabular}
  };

  \node[state_white, right of = mavros, shift = {(-1.4em, 0.0em)}] (optic_flow) {
    \begin{tabular}{c}
      \scriptsize Optic flow
    \end{tabular}
  };

  \node[state_gray, above of = optic_flow, shift = {(0.0em, -4.0em)}] (camera) {
    \begin{tabular}{c}
      \scriptsize Camera
    \end{tabular}
  };

  \node[state_gray, left of = mavros, shift = {(0.8em, 0.0em)}] (height) {
    \begin{tabular}{c}
      \scriptsize Height sensor
    \end{tabular}
  };

  \node[state_gray, left of = height, shift = {(0.3em, 0.0em)}] (rtk) {
    \begin{tabular}{c}
      \scriptsize RTK GPS
    \end{tabular}
  };

  \node[state_white, right of = optic_flow, shift = {(-1.5em, 0.0em)}] (slam) {
    \begin{tabular}{c}
      \scriptsize SLAM
    \end{tabular}
  };

  \node[state_gray, above of = slam, shift = {(0.0em, -4.0em)}] (lidar) {
    \begin{tabular}{c}
      \scriptsize LIDAR
    \end{tabular}
  };



  \path[-] ($(rtk.south)$) edge [->] ($(estimator.north) + (-2em, 0)$);
  \path[-] ($(height.south)$) edge [->] ($(estimator.north) + (-1.0em, 0)$);
  \path[-] ($(optic_flow.south)$) edge [->] ($(estimator.north) + (1.0em, 0)$);
  \path[-] ($(slam.south)$) edge [->] ($(estimator.north) + (2em, 0)$);
  \path[-] ($(mavros.south)$) edge [->] ($(estimator.north) + (0em, 0)$);

  \path[] ($(camera.south)$) edge [->] ($(optic_flow.north)$);

  \path[-] ($(slam.north |- lidar.south)$) edge [->] ($(slam.north)$);

  \path ($(se3_controller.east)$) edge [<->] ($(control_manager.west) + (0, 0.4em)$);
  \path ($(mpc_controller.east)$) edge [<->] ($(control_manager.west) + (0, 0.20em)$);
  \path ($(failsafe_controller.east)$) edge [<->] ($(control_manager.west) + (0, 0.0em)$);

  \path ($(speed_tracker.west)$) edge [<->] ($(control_manager.east) + (0, 0.00em)$);
  \path ($(landoff_tracker.west)$) edge [<->] ($(control_manager.east) + (0, 0.20em)$);
  \path ($(mpc_tracker.west)$) edge [<->] ($(control_manager.east) + (0, 0.40em)$);

  \path[-] ($(gain_manager.south)$) edge [->] ($(se3_controller.north)$);

  \draw[-] ($(constraint_manager.west)$) -- ($(control_manager.north |- constraint_manager.west) + (2em, 0)$) edge [->] ($(control_manager.north) + (2em, 0)$);

  \draw[-] ($(uav_manager.west)$) -- ($(control_manager.north |- uav_manager.west) + (1em, 0)$) edge [->] ($(control_manager.north) + (1em, 0)$);

  \path ($(estimator.south) + (0.0em, 0)$) edge [->] ($(control_manager.north) + (0.0em, 0)$);

  \draw[-] ($(estimator.east)$) -- ($(uav_manager.north |- estimator.east) + (0em, 0)$) edge [->] ($(uav_manager.north) + (0em, 0)$);

  \draw[-] ($(estimator.west)$) -- ($(gain_manager.north |- estimator.west) + (0em, 0)$) edge [->] ($(gain_manager.north) + (0em, 0)$);

  \draw[-] ($(estimator.east)$) -- ($(constraint_manager.north |- estimator.east) + (2.5em, 0)$) edge [->] ($(constraint_manager.north) + (2.5em, 0)$);

  \path ($(pixhawk.south)$) edge [->] ($(mavros.north)$);

  \path ($(mavros2.north |- control_manager.south)$) edge [->] ($(mavros2.north)$);
  \path ($(pixhawk2.north)$) edge [<-] ($(mavros2.south)$);







\end{tikzpicture}

%% file: main.bbl
\begin{thebibliography}{10}
\providecommand{\url}[1]{{#1}}
\providecommand{\urlprefix}{URL }
\expandafter\ifx\csname urlstyle\endcsname\relax
  \providecommand{\doi}[1]{DOI~\discretionary{}{}{}#1}\else
  \providecommand{\doi}{DOI~\discretionary{}{}{}\begingroup
  \urlstyle{rm}\Url}\fi

\bibitem{abeywardena2015design}
Abeywardena, D., Pounds, P., Hunt, D., Dissanayake, G.: Design and development
  of recopter: An open source ros-based multi-rotor platform for research.
\newblock In: ACRA (2015)

\bibitem{aboudonia2018composite}
Aboudonia, A., Rashad, R., El-Badawy, A.: {Composite hierarchical
  anti-disturbance control of a quadrotor UAV in the presence of matched and
  mismatched disturbances}.
\newblock Journal of Intelligent \& Robotic Systems \textbf{90}(1-2), 201--216
  (2018)

\bibitem{baca2018model}
Baca, T., Hert, D., Loianno, G., Saska, M., Kumar, V.: {Model Predictive
  Trajectory Tracking and Collision Avoidance for Reliable Outdoor Deployment
  of Unmanned Aerial Vehicles}.
\newblock In: 2018 IEEE/RSJ International Conference on Intelligent Robots and
  Systems, pp. 1--8. IEEE (2018)

\bibitem{baca2019timepix}
Baca, T., Jilek, M., et~al.: {Timepix Radiation Detector for Autonomous
  Radiation Localization and Mapping by Micro Unmanned Vehicles}.
\newblock In: 2019 IEEE/RSJ International Conference on Intelligent Robots and
  Systems, pp. 1--8. IEEE (2019)

\bibitem{baca2016embedded}
Baca, T., Loianno, G., Saska, M.: {Embedded Model Predictive Control of
  Unmanned Micro Aerial Vehicles}.
\newblock In: {IEEE MMAR}, pp. 992--997. IEEE (2016)

\bibitem{baca2020autonomous}
Baca, T., Penicka Robert~Stepan, P., Petrlik, M., Spurny, V., Hert, D., Saska,
  M.: {Autonomous Cooperative Wall Building by a Team of Unmanned Aerial
  Vehicles in the MBZIRC 2020 Competition}.
\newblock {submitted to Robotics and Autonomous System}  (2020)

\bibitem{baca2019autonomous}
Baca, T., Stepan, P., Spurny, B., Hert, D., Penicka, R., Saska, M., Thomas, J.,
  Loianno, G., Kumar, V.: {Autonomous Landing on a Moving Vehicle with an
  Unmanned Aerial Vehicle}.
\newblock {Journal of Field Robotics} \textbf{36}, 874--891 (2019)

\bibitem{baca2018rospix}
Baca, T., Turecek, D., McEntaffer, R., Filgas, R.: {Rospix: modular software
  tool for automated data acquisitions of Timepix detectors on Robot Operating
  System}.
\newblock Journal of Instrumentation \textbf{13}(11), C11008 (2018)

\bibitem{benallegue2008high}
Benallegue, A., Mokhtari, A., Fridman, L.: {High-order sliding-mode observer
  for a quadrotor UAV}.
\newblock International Journal of Robust and Nonlinear Control:
  IFAC-Affiliated Journal \textbf{18}(4-5), 427--440 (2008)

\bibitem{burri2015robust}
Burri, M., Datwiler, M., Achtelik, M.W., Siegwart, R.: {Robust state estimation
  for micro aerial vehicles based on system dynamics}.
\newblock In: 2015 IEEE international conference on robotics and automation,
  pp. 5278--5283. IEEE (2015)

\bibitem{cayero2019optimal}
Cayero, J., Rotondo, D., Morcego, B., Puig, V.: {Optimal state observation
  using quadratic boundedness: Application to UAV disturbance estimation}.
\newblock International Journal of Applied Mathematics and Computer Science
  \textbf{29}(1), 99--109 (2019)

\bibitem{ceballos2011genom}
Ceballos, A., De~Silva, L., Herrb, M., Ingrand, F., Mallet, A., Medina, A.,
  Prieto, M.: {GenoM as a robotics framework for planetary rover surface
  operations}.
\newblock In: ASTRA, pp. 12--14 (2011)

\bibitem{rodriguez2020explicit}
Rodriguez~de Cos, C., Acosta, J.A.: {Explicit Aerodynamic Model
  Characterization of a Multirotor Unmanned Aerial Vehicle in Quasi-Steady
  Flight}.
\newblock Journal of Computational and Nonlinear Dynamics \textbf{15}(8) (2020)

\bibitem{delmerico2019current}
Delmerico, J., et~al.: The current state and future outlook of rescue robotics.
\newblock Journal of Field Robotics \textbf{36}(7), 1171--1191 (2019)

\bibitem{diebel2006representing}
Diebel, J.: {Representing attitude: Euler angles, unit quaternions, and
  rotation vectors}.
\newblock Matrix \textbf{58}(15-16), 1--35 (2006)

\bibitem{ebeid2018survey}
Ebeid, E., Skriver, M., Terkildsen, K.H., Jensen, K., Schultz, U.P.: A survey
  of open-source uav flight controllers and flight simulators.
\newblock Microprocessors and Microsystems \textbf{61}, 11--20 (2018)

\bibitem{elkady2012robotics}
Elkady, A., et~al.: {Robotics middleware: A comprehensive literature survey and
  attribute-based bibliography}.
\newblock Journal of Robotics \textbf{2012} (2012)

\bibitem{faigl2019unsupervised}
Faigl, J., Vana, P., Penicka, R., Saska, M.: {Unsupervised Learning based
  Flexible Framework for Surveillance Planning with Aerial Vehicles}.
\newblock Journal of Field Robotics \textbf{36}(1), 270--301 (2019)

\bibitem{faigl2017onsolution}
Faigl, J., Vana, P., Saska, M., Baca, T., Spurny, V.: {On solution of the
  Dubins touring problem}.
\newblock In: {IEEE ECMR}, pp. 1--6 (2017)

\bibitem{furrer2016rotors}
Furrer, F., Burri, M., Achtelik, M., Siegwart, R.: {RotorS --- A modular gazebo
  MAV simulator framework}.
\newblock In: Robot Operating System (ROS), pp. 595--625. Springer (2016)

\bibitem{gao2018online}
Gao, F., Wu, W., Lin, Y., Shen, S.: Online safe trajectory generation for
  quadrotors using fast marching method and bernstein basis polynomial.
\newblock In: 2018 IEEE International Conference on Robotics and Automation,
  pp. 344--351. IEEE (2018)

\bibitem{giernacki2019realtime}
Giernacki, W., Horla, D., Baca, T., Saska, M.: {Real-time model-free
  minimum-seeking autotuning method for unmanned aerial vehicle controllers
  based on fibonacci-search algorithm}.
\newblock Sensors \textbf{19}(2), 312 (2019)

\bibitem{hentout2016survey}
Hentout, A., Maoudj, A., Bouzouia, B.: A survey of development frameworks for
  robotics.
\newblock In: IEEE ICMIC, pp. 67--72. IEEE (2016)

\bibitem{hentzen2019disturbnace}
Hentzen, D., Stastny, T., Siegwart, R., Brockers, R.: {Disturbance Estimation
  and Rejection for High-Precision Multirotor Position Control}.
\newblock In: IEEE/RSJ IROS, pp. 2797--2804. IEEE (2019)

\bibitem{inigo2012robotics}
Inigo-Blasco, P., et~al.: {Robotics software frameworks for multi-agent robotic
  systems development}.
\newblock Robotics and Autonomous Systems \textbf{60}(6), 803--821 (2012)

\bibitem{jasim2020robust}
Jasim, O.A., Veres, S.M.: A robust controller for multi rotor uavs.
\newblock Aerospace Science and Technology \textbf{105}, 106010 (2020)

\bibitem{kamel2017robust}
Kamel, M., Alonso-Mora, J., Siegwart, R., Nieto, J.: {Robust Collision
  Avoidance for Multiple Micro Aerial Vehicles Using Nonlinear Model Predictive
  Control}.
\newblock In: 2017 IEEE/RSJ International Conference on Intelligent Robots and
  Systems, pp. 236--243. IEEE (2017)

\bibitem{kohlbrecher2011flexible}
Kohlbrecher, S., Meyer, J., von Stryk, O., Klingauf, U.: {A Flexible and
  Scalable SLAM System with Full 3D Motion Estimation}.
\newblock In: IEEE SSRR. IEEE (2011)

\bibitem{kratky2020autonomous}
Kratky, V., Petracek, P., Spurny, V., Saska, M.: Autonomous reflectance
  transformation imaging by a team of unmanned aerial vehicles.
\newblock IEEE Robotics and Automation Letters \textbf{5}(2), 2302--2309 (2020)

\bibitem{labbadi2019robust}
Labbadi, M., Cherkaoui, M.: {Robust adaptive backstepping fast terminal sliding
  mode controller for uncertain quadrotor UAV}.
\newblock Aerospace Science and Technology \textbf{93}, 105306 (2019)

\bibitem{lafflitto2018introduction}
L'Afflitto, A., Anderson, R.B., Mohammadi, K.: {An Introduction to Nonlinear
  Robust Control for Unmanned Quadrotor Aircraft: How to Design Control
  Algorithms for Quadrotors Using Sliding Mode Control and Adaptive Control
  Techniques [Focus on Education]}.
\newblock IEEE Control Systems Magazine \textbf{38}(3), 102--121 (2018)

\bibitem{lee2010geometric}
Lee, T., et~al.: {Geometric tracking control of a quadrotor UAV on SE(3)}.
\newblock In: {2010 IEEE Conference on Decision and Control}, pp. 5420--5425.
  IEEE (2010)

\bibitem{lee2013nonlinear}
Lee, T., et~al.: {Nonlinear robust tracking control of a quadrotor UAV on
  SE(3)}.
\newblock Asian Journal of Control \textbf{15}(2), 391--408 (2013)

\bibitem{loianno2018localization}
Loianno, G., Spurny, V., Thomas, J., Baca, T., Thakur, D., et~al.:
  {Localization, Grasping, and Transportation of Magnetic Objects by a team of
  MAVs in Challenging Desert like Environments}.
\newblock IEEE Robotics and Automation Letters \textbf{3}(3), 1576--1583 (2018)

\bibitem{meier2015px4}
Meier, L., Honegger, D., Pollefeys, M.: {PX4: A node-based multithreaded open
  source robotics framework for deeply embedded platforms}.
\newblock In: 2015 IEEE International Conference on Robotics and Automation,
  pp. 6235--6240. IEEE (2015)

\bibitem{mellado2013mavwork}
Mellado-Bataller, I., Pestana, J., Olivares-Mendez, M.A., Campoy, P., Mejias,
  L.: {MAVwork: a framework for unified interfacing between micro aerial
  vehicles and visual controllers}.
\newblock In: Frontiers of Intelligent Autonomous Systems, pp. 165--179.
  Springer (2013)

\bibitem{mellinger2011minimum}
Mellinger, D., Kumar, V.: Minimum snap trajectory generation and control for
  quadrotors.
\newblock In: 2011 IEEE International Conference on Robotics and Automation,
  pp. 2520--2525. IEEE (2011)

\bibitem{metta2006yarp}
Metta, G., Fitzpatrick, P., Natale, L.: {YARP: yet another robot platform}.
\newblock International Journal of Advanced Robotic Systems \textbf{3}(1), 8
  (2006)

\bibitem{mohta2018fast}
Mohta, K., Watterson, M., Mulgaonkar, Y., Liu, S., Qu, C., Makineni, A.,
  Saulnier, K., Sun, K., Zhu, A., Delmerico, J., et~al.: {Fast, autonomous
  flight in GPS-denied and cluttered environments}.
\newblock {Journal of Field Robotics} \textbf{35}(1), 101--120 (2018)

\bibitem{mueller2015computationally}
Mueller, M.W., Hehn, M., D'Andrea, R.: A computationally efficient motion
  primitive for quadrocopter trajectory generation.
\newblock IEEE Transactions on Robotics \textbf{31}(6), 1294--1310 (2015)

\bibitem{nascimento2019nmpc}
Nascimento, I.B., Ferramosca, A., Piment, L.C., Raffo, G.V.: {NMPC Strategy for
  a Quadrotor UAV in a 3D Unknown Environment}.
\newblock In: ICAR, pp. 179--184. IEEE (2019)

\bibitem{nascimento2019position}
Nascimento, T.P., Saska, M.: {Position and attitude control of multi-rotor
  aerial vehicles: A survey}.
\newblock Annual Reviews in Control \textbf{48}, 129--146 (2019)

\bibitem{penicka2019data}
Penicka, R., Faigl, J., Saska, M., Vana, P.: Data collection planning with
  non-zero sensing distance for a budget and curvature constrained unmanned
  aerial vehicle.
\newblock Autonomous Robots  (2019)

\bibitem{penicka2017dubins}
Penicka, R., Faigl, J., Vana, P., Saska, M.: {Dubins orienteering problem}.
\newblock {IEEE Robotics and Automation Letters} \textbf{2}(2), 1210--1217
  (2017)

\bibitem{penicka2017neighborhoods}
Penicka, R., Faigl, J., Vana, P., Saska, M.: Dubins orienteering problem with
  neighborhoods.
\newblock In: 2017 IEEE International Conference on Unmanned Aircraft Systems,
  pp. 1555--1562 (2017)

\bibitem{penicka2017reactive}
Penicka, R., Saska, M., Reymann, C., Lacroix, S.: {Reactive dubins traveling
  salesman problem for replanning of information gathering by UAVs}.
\newblock In: {IEEE ECMR}. IEEE (2017)

\bibitem{pereira2019nonlinear}
Pereira, J.C., Leite, V.J., Raffo, G.V.: {Nonlinear Model Predictive Control on
  SE(3) for Quadrotor Trajectory Tracking and Obstacle Avoidance}.
\newblock In: ICAR, pp. 155--160. IEEE (2019)

\bibitem{petracek2020dronument}
Petracek, P., Kratky, V., Saska, M.: Dronument: System for reliable deployment
  of micro aerial vehicles in dark areas of large historical monuments.
\newblock IEEE Robotics and Automation Letters \textbf{5}(2), 2078--2085 (2020)

\bibitem{petrlik2020robust}
Petrlik, M., Baca, T., Hert, D., Vrba, M., Krajnik, T., Saska, M.: {A Robust
  UAV System for Operations in a Constrained Environment}.
\newblock {IEEE Robotics and Automation Letters} \textbf{5} (2020)

\bibitem{petrlik2019coverage}
Petrlik, M., Vonasek, V., Saska, M.: Coverage optimization in the cooperative
  surveillance task using multiple micro aerial vehicles.
\newblock In: IEEE SMC (2019)

\bibitem{poultney2018robust}
Poultney, A., Kennedy, C., Clayton, G., Ashrafiuon, H.: {{Robust tracking
  control of quadrotors based on differential flatness: Simulations and
  experiments}}.
\newblock IEEE/ASME Transactions on Mechatronics \textbf{23}(3), 1126--1137
  (2018)

\bibitem{qin2018vins}
Qin, T., Li, P., Shen, S.: {Vins-mono: A robust and versatile monocular
  visual-inertial state estimator}.
\newblock IEEE Transactions on Robotics \textbf{34}(4), 1004--1020 (2018)

\bibitem{quigley2009ros}
Quigley, M., et~al.: Ros: an open-source robot operating system.
\newblock In: IEEE ICRA workshop on open source software, vol.~3, p.~5. Kobe,
  Japan (2009)

\bibitem{ramirez2020fuzzy}
Ramirez-Mendoza, A., Covarrubias-Fabela, J., Amezquita-Brooks, L.,
  Garcia~Salazar, O., Ramirez-Mendoza, W.: {Fuzzy Adaptive Neurons Applied to
  the Identification of Parameters and Trajectory Tracking Control of a
  Multi-Rotor Unmanned Aerial Vehicle Based on Experimental Aerodynamic Data}.
\newblock Journal of Intelligent \& Robotic Systems pp. 380--393 (2020)

\bibitem{richter2016polynomial}
Richter, C., Bry, A., Roy, N.: Polynomial trajectory planning for aggressive
  quadrotor flight in dense indoor environments.
\newblock In: Robotics Research, pp. 649--666. Springer (2016)

\bibitem{roucek2019darpa}
Roucek, T., et~al.: {DARPA Subterranean Challenge: Multi-robotic exploration of
  underground environments}.
\newblock In: IEEE MESAS, vol. 11995, pp. 274--290 (2019)

\bibitem{roussel2019nonlinear}
Roussel, M.R.: {Nonlinear Dynamics: A hands-on introductory survey}.
\newblock Morgan \& Claypool Publishers (2019)

\bibitem{saikin2020wildfire}
Saikin, D.A., Baca, T., Gurtner, M., Saska, M.: {Wildfire Fighting by Unmanned
  Aerial System Exploiting Its Time-Varying Mass}.
\newblock IEEE Robotics and Automation Letters \textbf{5}(2), 2674--2681 (2020)

\bibitem{sanchez2016aerostack}
Sanchez-Lopez, J.L., Suarez~Fernandez, R.A., Bavle, H., Sampedro, C., Molina,
  M., Pestana, J., Campoy, P.: {AEROSTACK: An architecture and open-source
  software framework for aerial robotics}.
\newblock In: 2016 IEEE International Conference on Unmanned Aircraft Systems,
  pp. 332--341 (2016)

\bibitem{sanchez2016reliable}
Sanchez-Lopez, J.L., et~al.: {A reliable open-source system architecture for
  the fast designing and prototyping of autonomous multi-UAV systems:
  Simulation and experimentation}.
\newblock Journal of Intelligent \& Robotic Systems \textbf{84}(1-4), 779--797
  (2016)

\bibitem{sanchez2020trajectory}
Sanchez-Lopez, J.L., et~al.: {Trajectory Tracking for Aerial Robots: an
  Optimization-Based Planning and Control Approach}.
\newblock Journal of Intelligent \& Robotic Systems pp. 1--44 (2020)

\bibitem{saska2019large}
Saska, M.: {Large Sensors with Adaptive Shape Realised by Self-stabilised
  Compact Groups of Micro Aerial Vehicles}.
\newblock In: Robotics Research, pp. 101--107. Springer International
  Publishing (2020)

\bibitem{saska2020formation}
Saska, M., Hert, D., Baca, T., Kratky, V., Nascimento, T.: {Formation Control
  of Unmanned Micro Aerial Vehicles for Straitened Environments}.
\newblock Autonomous Robots pp. 1573--7527 (2020)

\bibitem{saska2017documentation}
Saska, M., Kratky, V., Spurny, V., Baca, T.: {Documentation of dark areas of
  large historical buildings by a formation of unmanned aerial vehicles using
  model predictive control}.
\newblock In: {IEEE ETFA}. IEEE (2017)

\bibitem{saska2016formations}
Saska, M., et~al.: Formations of unmanned micro aerial vehicles led by
  migrating virtual leader.
\newblock In: {IEEE ICARCV}. IEEE (2016)

\bibitem{schmittle2018openuav}
Schmittle, M., et~al.: {OpenUAV: a UAV testbed for the CPS and robotics
  community}.
\newblock In: ACM/IEEE ICCPS, pp. 130--139. IEEE (2018)

\bibitem{spurny2016complex}
Spurny, V., Baca, T., Saska, M.: {Complex manoeuvres of heterogeneous MAV-UGV
  formations using a model predictive control}.
\newblock In: {IEEE MMAR}, pp. 998--1003 (2016)

\bibitem{spurny2019cooperative}
Spurny, V., Baca, T., Saska, M., Penicka, R., Krajnik, T., et~al.: {Cooperative
  Autonomous Search, Grasping and Delivering in a Treasure Hunt Scenario by a
  Team of UAVs}.
\newblock {Journal of Field Robotics} \textbf{{36}}({1}), {125--148} (2019)

\bibitem{spurny2019transport}
Spurny, V., Petrlik, M., Vonasek, V., Saska, M.: Cooperative transport of large
  objects by a pair of unmanned aerial systems using sampling-based motion
  planning.
\newblock In: IEEE ETFA, pp. 955--962 (2019)

\bibitem{stepan2019vision}
Stepan, P., Krajnik, T., Petrlik, M., Saska, M.: {Vision techniques for
  on-board detection, following and mapping of moving targets}.
\newblock Journal of Field Robotics \textbf{36}(1), 252--269 (2019)

\bibitem{stibinger2020localization}
Stibinger, P., Baca, T., Saska, M.: {Localization of Ionizing Radiation Sources
  by Cooperating Micro Aerial Vehicles With Pixel Detectors in Real-Time}.
\newblock IEEE Robotics and Automation Letters \textbf{5}, 3634--3641 (2020)

\bibitem{suarez2020skyeye}
Suarez~Fernandez, R., Rodriguez~Ramos, A., Alvarez, A., et~al.: {The Skyeye
  Team Participation in the 2020 Mohamed Bin Zayed International Robotics
  Challenge} (2020).
\newblock Online (accessed on 2020/09/05):
  \url{https://www.researchgate.net/publication/339725858_The_Skyeye_Team_Participation_in_the_2020_Mohamed_Bin_Zayed_International_Robotics_Challenge}

\bibitem{tsardoulias2017robotic}
Tsardoulias, E., Mitkas, P.: Robotic frameworks, architectures and middleware
  comparison.
\newblock arXiv preprint arXiv:1711.06842  (2017)

\bibitem{vrba2019onboard}
Vrba, M., Hert, D., Saska, M.: Onboard marker-less detection and localization
  of non-cooperating drones for their safe interception by an autonomous aerial
  system.
\newblock IEEE Robotics and Automation Letters \textbf{4}(4), 3402--3409 (2019)

\bibitem{vrba2019realtime}
Vrba, M., Pogran, J., Pritzl, V., Spurny, V., Saska, M.: Real-time localization
  of transmission sources using a formation of micro aerial vehicles.
\newblock In: IEEE RCAR, pp. 203--208 (2019)

\bibitem{vrba2020markerless}
Vrba, M., Saska, M.: Marker-less micro aerial vehicle detection and
  localization using convolutional neural networks.
\newblock IEEE Robotics and Automation Letters \textbf{5}(2), 2459--2466 (2020)

\bibitem{walter2017selflocalization}
Walter, V., Novak, T., Saska, M.: Self-localization of unmanned aerial vehicles
  using onboard sensors, with focus on optical flow from camera image.
\newblock In: MESAS (2017)

\bibitem{walter2018fast}
Walter, V., Saska, M., Franchi, A.: {Fast mutual relative localization of UAVs
  using ultraviolet LED markers}.
\newblock In: 2018 IEEE International Conference on Unmanned Aircraft Systems
  (2018)

\bibitem{walter2018mutual}
Walter, V., Staub, N., Saska, M., Franchi, A.: {Mutual Localization of UAVs
  based on Blinking Ultraviolet Markers and 3D Time-Position Hough Transform}.
\newblock In: IEEE CASE (2018)

\bibitem{walter2019uvdar}
Walter, W., Staub, N., Franchi, A., Saska, M.: {UVDAR System for Visual
  Relative Localization With Application to Leader-Follower Formations of
  Multirotor UAVs}.
\newblock IEEE Robotics and Automation Letters \textbf{4}(3), 2637--2644 (2019)

\bibitem{xiao2020xtdrone}
Xiao, K., Tan, S., Wang, G., An, X., Wang, X., Wang, X.: {XTDrone: A
  Customizable Multi-Rotor UAVs Simulation Platform}.
\newblock arXiv preprint arXiv:2003.09700  (2020)

\bibitem{zhang2019robust}
Zhang, J., Gu, D., Deng, C., Wen, B.: Robust and adaptive backstepping control
  for hexacopter uavs.
\newblock IEEE Access \textbf{7}, 163502--163514 (2019)

\bibitem{zhang2014loam}
Zhang, J., Singh, S.: {LOAM: Lidar Odometry and Mapping in Real-time}.
\newblock In: Robotics: Science and Systems, vol.~2 (2014)

\end{thebibliography}
